\documentclass[a4paper,fleqn]{cas-dc}

\usepackage[authoryear]{natbib}

\def\tsc#1{\csdef{#1}{\textsc{\lowercase{#1}}\xspace}}
\tsc{WGM}
\tsc{QE}
\tsc{EP}
\tsc{PMS}
\tsc{BEC}
\tsc{DE}

\usepackage{hyperref}
\usepackage{url}

\usepackage{microtype}
\usepackage{graphicx}
\usepackage{graphics}
\usepackage[caption=false,font=normalsize,labelfont=sf,textfont=sf]{subfig}
\usepackage{booktabs} 
\usepackage[utf8]{inputenc}
\usepackage{csquotes}
\usepackage{amsmath, amsfonts, amssymb}
\usepackage{dsfont}	
\usepackage{tikz}
\usepackage{adjustbox}
\usepackage{xcolor}
\usepackage{float}
\usepackage{csquotes}

\usepackage{algorithmic}
\usepackage{array}
\usepackage{textcomp}
\usepackage{stfloats}
\usepackage{verbatim}

\usepackage{multirow}
\usepackage{cleveref}

\allowdisplaybreaks

\usepackage[colorinlistoftodos, color=yellow]{todonotes}

\graphicspath{ {imagens/} {imagens/ClassMNIST/} {imagens/NLL-MNIST/} {imagens/ClassMNIST-mesmaTaxa/} {imagens/ClassMNIST-comp/} {imagens/NLL-MNIST-comp/} }

\definecolor{steelblue}{RGB}{70,130,180}
\definecolor{lightblue}{RGB}{120,180,250}

\begin{document}
	\let\WriteBookmarks\relax
	\def\floatpagepagefraction{1}
	\def\textpagefraction{.001}
	\shorttitle{Optimizing Connectivity through Network Gradients for Restricted Boltzmann Machines}
	\shortauthors{A.C.N. de Oliveira and D.R Figueiredo }
	
	\title [mode = title]{Optimizing Connectivity through Network Gradients for Restricted Boltzmann Machines}                      
	
	\tnotetext[1]{Work published at Neural Networks: \href{https://www.sciencedirect.com/science/article/pii/S089360802500365X}{www.sciencedirect.com}}
	\tnotetext[2]{This work received financial support through research grants provided by CNPq, FAPERJ and CAPES (Brazil).}
	\author[1]{Amanda Camacho Novaes de Oliveira}[orcid=0000-0003-0711-5970]
	\cormark[1]
	\fnmark[]
	\ead{amandacno@cos.ufrj.br}
	\credit{Conceptualization of this study, Methodology, Software, Writing}
	\affiliation[1]{organization={Systems Engineering and Computer Science (PESC), Federal University of Rio de Janeiro (UFRJ)},
		addressline={Av. Athos da Silveira Ramos 149, Bloco H-319, Cidade Universitária},  
		postcode={21945-970}, 
		city={Rio de Janeiro},
		state={RJ},
		country={Brazil}}
		
	\author[1]{Daniel Ratton Figueiredo}[orcid=0000-0001-9341-6619]
	\cormark[1]
	\fnmark[]
	\ead{daniel@cos.ufrj.br}
	\ead[URL]{https://www.cos.ufrj.br/~daniel/}
	\credit{Supervision, Conceptualization, Methodology, Writing}
	
	\cortext[cor1]{Corresponding author}
	
\begin{abstract}
	Leveraging sparse networks to connect successive layers in deep neural networks has recently been shown to provide benefits to large-scale state-of-the-art models. However, network connectivity also plays a significant role in the learning performance of shallow networks, such as the classic Restricted Boltzmann Machine (RBM). Efficiently finding sparse connectivity patterns that improve the learning performance of shallow networks is a fundamental problem. While recent principled approaches explicitly include network connections as model parameters that must be optimized, they often rely on explicit penalization or network sparsity as a hyperparameter. This work presents the Network Connectivity Gradients (NCG), an optimization method to find optimal connectivity patterns for RBMs. NCG leverages the idea of network gradients: given a specific connection pattern, it determines the gradient of every possible connection and uses the gradient to drive a continuous connection strength parameter that in turn is used to determine the connection pattern. 
	Thus, learning RBM parameters and learning network connections is truly jointly performed, albeit with different learning rates, and without changes to the model's classic energy-based objective function. The proposed method is applied to the MNIST and other data sets showing that better RBM models are found for the benchmark tasks of sample generation and classification. Results also show that NCG is robust to network initialization and is capable of both adding and removing network connections while learning. 
\end{abstract}


	\maketitle

\section{Introduction} \label{sec:intro}
	While most neural network architectures adopt a fully connected network between units of successive layers, it has been long recognized that the network's connectivity plays a fundamental role in the model, not only reducing the number of parameters but also leading to a more accurate model or to faster learning~\citep{pruningSurvey1993, pruningSurvey}. 
	This finding has recently reemerged in the context of deep neural networks, and while classic architectures such as ResNet~\citep{ResNets} and BERT~\citep{BERT} have millions of parameters that must be learned, recent works indicate that only a small fraction of them are necessary for the model to attain a similar performance under an equivalent training effort~\citep{pruningSurvey}. Finding the right sparse network connectivity that yields similar performance is known as {\it network pruning}~\citep{pruningSurvey1993, pruningSurvey}.  
	
	Most works on network pruning consider deep neural network models, given their large number of connections. However, connectivity patterns also play a fundamental role even on simple two-layer networks such as Restricted Boltzmann Machine (RBM). The RBM is a simple energy-based model for unsupervised learning. Originally designed for sample generation, it has been since applied to a myriad of tasks. The Boltzmann machines were inspired by physics models and their advancement as a powerful tool was recognized by this year's Nobel Prize in Physics (along with Hopfield networks)~\citep{nobel}. 

    While the reduction in the absolute number of parameters in an RBM may be small, recent works observe that an effective connectivity pattern can yield superior learning curves, allowing the model to learn faster and better~\citep{SET_mocanu, selfAnonymous}. 
    In fact, the connectivity of an RBM can be interpreted as a hyperparameter that influences its performance, just as the number of neurons (another hyperparameter) in its hidden layer~\citep{fischer2014trainingRBMintro, iRBM}. 
    
    Finding the best network connectivity for a given neural network is not a trivial problem, given its dependence on the input (training data), the discrete nature of the connections, and the exponentially large space of possible connection patterns (there are $2^{n^2}$ different networks between two layers with $n$ units each). 
    A common approach to tackle this problem is to construct the connectivity pattern while training the network, starting with a dense network and using some pruning strategy to remove connections in a sequence of rounds (train and prune)~\citep{han2015learningConnections, frankleLotteryTicket}.
    A less explored yet more principled approach is to explicitly include the network connections as parameters that must be optimized in the model~\citep{continuousSparsification,pmlr-chen-lotteryGNN,Zhou21learningStrcSparsScratch}. 
    Intuitively, the model should jointly learn the optimal network weights and network connections during training. 
    
    However, the discrete nature of the network connections poses a challenge to widely used continuous optimization frameworks such as (stochastic) gradient descent, since discrete connection parameters have no derivatives (and thus, no gradient). To circumvent this problem, recent approaches adopt continuous variables and functions to represent network connections while also adopting some form of threshold to prune connections~\citep{savarese2018_MCtoLV}, which is the approach adopted in our framework as well. 
   
    Moreover, in order to drive the model towards sparse networks, many approaches incorporate an explicit penalization term (representing the number of connections) into the objective function~\citep{pruning_SHI2024,ma2019transformed}.  
    Also, the discrete network connectivity is often only determined at the end of training (or at the end of a round of training), and therefore it does not evolve jointly with the optimization of other parameters~\citep{pruning_SHI2024,ma2019transformed,savarese2018_MCtoLV}. 
    In contrast, this work proposes a novel method tailored to RBMs based on the notion of \enquote{network gradients}. 
    
    In a nutshell, the Network Connectivity Gradient (NCG) method\footnote{
    	Code available at \url{https://github.com/AmieOliveira/NCG}.
    } computes the gradient for every possible network connection for any given connectivity pattern. 
    Moreover, NCG uses a continuous parameter to represent the strength of every possible network connection, which is updated according to the gradient as any other model parameter. Finally, the network strength is thresholded to yield a discrete connectivity pattern {\em during} optimization (i.e., at each training epoch), which in turn determines how information (probabilities in the RBM) and gradients flow on the model during training.
	
	Intuitively, the network gradient indicates the relevance of each possible connection given the current connection pattern. This gradient drives the connection strength parameter, which in turn determines if a connection should be present or absent, effectively adjusting the connection pattern as the model is trained. Thus, if the initial connectivity pattern is too sparse or too dense, NCG will enable or disable connections early during training, respectively. In essence, NCG truly learns the network connectivity jointly with other RBM parameters, albeit with possibly different learning rates. Note that no changes are required to the objective function of the RBM and no regularization is necessary. 

    The main ideas behind NCG are two novel model parameters: a connection activation binary matrix ($\textbf{A}$) and a connection strength matrix ($\textbf{A}'$). These parameters must be learned along with other RBM parameters. This work determines exactly how all such parameters are to be jointly learned in the context of RBMs (using approximations to the partial derivatives with respect to parameters using contrastive divergence). In this regard, NCG, as proposed in this paper, is specific to RBMs. However, its main ideas could be extended to other neural network models. 
    
    Beyond proposing NCG, this work evaluates the method using the MNIST data set (more information on Section \ref{sec:result}) on two orthogonal tasks often used to assess RBMs: sample generation (average NLL is the performance metric) and input classification (accuracy is the performance metric). For the classification task, two other data sets from the UCI Evaluation Suite~\citep{UCIdata} are considered (Mushrooms and Connect-4). In both tasks NCG shows superior learning curves, learning faster as well as more accurately than a classic fully connected RBM. The evaluation also shows that NCG removes and adds network connections during training, indicating its effectiveness in searching for optimal network patterns and robustness to initialization. Comparison with static patterns and the SET method~\citep{SET_mocanu} (also designed for the optimization of the network connectivity of RBMs) indicate the superiority of NCG, especially during the early phases of training. 
    
	The remainder of this work is organized as follows: Section \ref{sec:lit} has a cursory discussion of related works; Section \ref{sec:rbm} imparts a brief explanation of the RBM; Section \ref{sec:ncg} presents the NCG method here proposed; Section \ref{sec:result} shows the experimental results; and Section \ref{sec:conc} has the concluding remarks. 

\section{Related Work}
\label{sec:lit}

    The idea of removing network connections to improve the learning curve of a neural network model has been explored at least since the early 90s~\citep{lecun1990optimal,pruningSurvey1993}. However, the advancement of deep network models with tens of layers and billions of parameters (connections) has provided a novel motivation for network pruning along with strong evidence that networks with only a small fraction of the total connections can have similar learning performance~\citep{frankleLotteryTicket,pruningSurvey,pruning+quantizaztionSurvey,zhang_pruningTNNLS}. 

    However, finding an optimal connectivity pattern for two adjacent layers is not a trivial task. Most pruning approaches start with dense networks and iterate in rounds of training the model parameters and using the parameter values (and the input samples) to prune network connections. In these approaches, different methods are often used to determine which connections should be removed between rounds \citep{pruningSurvey,pruning+quantizaztionSurvey}. An orthogonal approach (not considered in this work) is to perform pruning even before training the network \citep{lee2019snip,de2021progressive}.
    
    A more principled approach explicitly includes the network connectivity as a parameter of the model, making it a part of the optimization problem. This often requires increasing the number of parameters and modifying the objective function to induce pruning of the connections. 
    A prominent example is Continuous Sparsification~\citep{continuousSparsification}, which uses continuous parameters and continuous functions to approximate the discrete nature of network connections, and adds a penalization term to the objective function. The discrete network connectivity is determined at the end of training rounds. \cite{pmlr-chen-lotteryGNN}, with its Unified GNN Sparsification (UGS), deploy a similar approach tailored to Graph Neural Networks (GNN). Akin to this present work, Discovering Neural Wirings (DNW) \citep{DNW} does not add penalization to the objective function, but keeps a constant number of edges (a hyperparameter) with the largest weight magnitude, discovering good sparse subnetworks in predefined network architectures. Another recent approach is the Sparse-Refined Straight-Through Estimator (SR-STE) \citep{Zhou21learningStrcSparsScratch} where the (discrete) connectivity pattern is updated at each training iteration. However, the method assumes that each input unit is connected to a fixed number of output units and thus sparsity is predefined (a hyperparameter). Another recent approach is to perform pruning in two phases (learning phase and fine-tuning phase) using two different regularization functions \citep{pruning_SHI2024}. 
	
    All prior works above focus on deep neural networks. However, network connectivity also plays a fundamental role in simple two-layer networks, including the Restricted Boltzmann Machine (RBM), a principled and probabilistic model that has been widely explored and applied in literature~\citep{fischer2014trainingRBMintro,Decelle_2021,rbmNew_AGLIARI2022}. RBMs' hyperparameters have a significant impact on the model's performance which has prompted different methods that choose adequate hyperparameters for a given task and input data~\citep{iRBM,hintonGuiaRBM,papa2015harmonySearchRBM}. A prominent example is the infinite RBM~\citep{iRBM}, a variation where the number of hidden units (a hyperparameter in the classic model) is an explicit model parameter that is determined during training. 
    
    The connectivity between layers has also been investigated for RBMs. For example, recent work has shown that crafted and fixed connectivity patterns can yield significantly better learning performance on RBMs~\citep{selfAnonymous}, and that sparse representations of RBMs can lead to more accurate and faster learning for Deep Belief Models (DBN) \citep{wang_rbmTNNLS}. 
    Note that \cite{selfAnonymous}, our prior work on designing sparse RBMs, did not consider optimization of the network connectivity but fixed connectivity patterns and their influence on the learning curve.
    
    The Sparse Boltzmann Machines (SBM) is a model where the connectivity is a sparse two-layer tree-like network~\citep{SBM}. Different tree-like networks can be learned from data and then used as hyperparameters when training the RBM, generating models that are less likely to overfit and that have better interpretability with respect to the dense RBM. In a more recent work, the network connectivity of the RBM is learned during training along with other model parameters. The approach proposed by \cite{SET_mocanu} (called SET method) removes connections with the smallest weights and adds the same number of randomly chosen new connections at each training round. Thus, network sparsity is predefined (a hyperparameter). 
    In contrast, NCG (to be presented) learns the connectivity and sparseness during training using the gradients of the unmodified objective function of the model.

\section{The Restricted Boltzmann Machine}
\label{sec:rbm}
	The Restricted Boltzmann Machine (RBM), first proposed by \cite{smolensky1986rbmProposal} under the name \textit{Harmonium}, is an energy-based model for unsupervised learning. It is a classic neural network model that has been applied to a number of different tasks. While initially designed for sample generation~\citep{Decelle_2021, leRoux2011_genRBM, Tang12robustboltzmann}, it has been used for classification tasks~\citep{tielemanLikelihoodGradient, classRBM} and as a preprocessing method for downstream tasks~\citep{midhun2014_deepNNusingRBM}. 
    RBMs are also the building blocks of Deep Belief Networks (DBN), which find numerous applications in modern problems~\citep{wang_rbmTNNLS, AIS, qiang2020_fMRI_DBN}. Furthermore, Boltzmann machines were recognized (along with Hopfield networks) by the 2024 Nobel Prize in Physics~\citep{nobel}, given the numerous applications and advancements in physics-related fields throughout the years \citep{physics_peng2008, physics_melko2019, physics_LIU2020109429, moro2023anomaly}. 
	
\begin{figure}[h]
		\centering
		\def\layersep{1.7cm}
\def\scaledist{1}
\begin{tikzpicture}[shorten >=1pt,->,draw=black!90, node distance=1.6*\layersep]
	\tikzstyle{every pin edge}=[<-,shorten <=1pt]
	\tikzstyle{neuron}=[circle,fill=black!25,minimum size=20pt,inner sep=0pt]
	\tikzstyle{visible unit}=[neuron, fill=steelblue];
	\tikzstyle{hidden unit}=[neuron, fill=gray!85];
	
	\foreach \name / \x in {1,...,5}
		\node[hidden unit] (h-\name) at (1.5*\x*\scaledist, 0) {\color{white} $ h_{\name} $};
	
	\foreach \name / \x in {1,...,4}
		\node[visible unit] (x-\name) at (1.5*\x*\scaledist+0.75*\scaledist, -\layersep) {\color{white} $ x_{\name} $};
	
	\node[left,align=left] at (1.5*5*\scaledist+.5*\scaledist,-1.2*\layersep/2){\textbf{W}};
	\foreach \name / \x in {1,...,5}
		\node[left,align=left] at (1.5*\x*\scaledist+.8*\scaledist,.35){\footnotesize $ b_{\name} $};
	\foreach \name / \x in {1,...,4}
		\node[left,align=left] at (1.5*\x*\scaledist+1.6*\scaledist, -1.2*\layersep){\footnotesize $ d_{\name} $};
	
	\foreach \source in {1,...,4}
	\foreach \dest in {1,...,5}
	\path[-] (h-\dest) edge (x-\source);
\end{tikzpicture}
		\caption{The RBM network for 4 visible ($ X $) and 5 hidden units ($ H $).}
		\label{fig:grafoRBM}
	\end{figure}
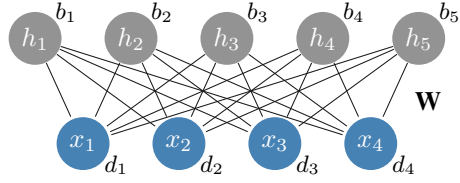

	An RBM is a probabilistic model composed of two layers of binary units: one visible \textbf{x} of size $ X $, representing the data, and one hidden (or latent) \textbf{h} of size $ H $, that extracts characteristics and increases learning ability. The two layers are fully connected through undirected weighted connections in a bipartite network. 
	Figure \ref{fig:grafoRBM} shows the example of an RBM network with $ X = 4 $ and $ H = 5 $.
	
	Each configuration $ (\textbf{x}, \textbf{h}) $ has the following associated energy:
	$ 
		E(\textbf{x}, \textbf{h}) = - \textbf{h}^\textsc{t}\textbf{W}\textbf{x} 
		- \textbf{x}^\textsc{t}\textbf{d} 
		- \textbf{h}^\textsc{t}\textbf{b} ,
	$ 
	where $ \textbf{W} \in \mathbb{R}^{H,X} $ is the weight matrix of the layers' connections ($ w_{ij} $ is the weight between visible unit $ x_j $ and hidden unit $ h_i $), 
	$ \textbf{d} \in \mathbb{R}^{X} $ is the visible units' bias vector ($ d_j $ is the bias 
	for $ x_j $) and $ \textbf{b}  \in \mathbb{R}^{H} $ is the hidden units' bias vector 
	($ b_i $ is the bias for $ h_i $). $ \textbf{W}$, $\textbf{d}$ and  $\textbf{b}$ are the model parameters, subsequently denoted by $\theta = (\textbf{W}, \textbf{d}, \textbf{b})$.
	The probability distribution of the RBM is defined as $P_\theta(\textbf{x}, \textbf{h}) = Z^{-1} e^{-E(\textbf{x}, \textbf{h})}$, with $ Z $ being the normalization constant (or partition function). 
	Note that this equation is in general not tractable due to the very large number of configurations ($2^{X+H}$, since all units are binary), and therefore for the most part one cannot know the exact probability of a given $ (\textbf{x}, \textbf{h}) $ configuration. 

\subsection{Training the Restricted Boltzmann Machine}
	The RBM is typically trained to minimize the Negative Log-Likelihood (NLL) of 
	the available data set, which is equivalent to maximizing the Log-Likelihood. In 
	this case, the average NLL is often adopted in order to simplify the learning procedure. Given a data set $ \{ \textbf{x}^{(t)} \}_{t=1}^T $ with $ T $ samples, the average NLL of the model is simply $ \frac{1}{T} \sum_{t=1}^T - \ln P_\theta(\textbf{x}^{(t)}) $. Note that the probability $P_\theta(\textbf{x}^{(t)})$ depends on the model parameters, $\theta$. 
	
	The RBM is trained by applying Stochastic Gradient Descent (SGD)~\citep{SGD}. Due to the intractability of the normalization constant, training methods such as Contrastive Divergence (CD)~\citep{hintonCD} approximate the gradient with the following expression: 
	\begin{equation}\label{eq:CDapprox}
		\begin{aligned}
			&\frac{1}{|\mathcal{B}|} \sum_{t \in \mathcal{B}} \mathbb{E}_\textbf{h}\left[ \nabla_\theta E(\textbf{x}, \textbf{h} ) \left| \textbf{x} = \textbf{x}^{(t)} \right. \right]  \\ 
			&\quad- \frac{1}{|\mathcal{B}|} \sum_{t \in \mathcal{B}} \mathbb{E}_{\textbf{h}}\left[ \nabla_\theta E(\textbf{x}, \textbf{h} ) \left| \textbf{x} = \tilde{\textbf{x}}^{(t)} \right. \right] , 
		\end{aligned}
	\end{equation}
	where $ \mathcal{B} $ corresponds to a batch of samples randomly chosen from the data, since \cite{hintonGuiaRBM} has shown that computing the gradients using batches rather than the full data often leads to better learning curves; 
	and $ \tilde{\textbf{x}}^{(t)} $ is a random sample of the RBM given its parameters. Note that equation \eqref{eq:CDapprox} requires generating a random sample from the RBM distribution for each data sample $ \textbf{x}^{(t)} $, which is done applying $ k $ steps of Gibbs Sampling~\citep{gibbsSampling} on the model, starting from the data sample $ \textbf{x}^{(t)} $. 
	
	While there are other methods that yield better approximations by changing how $ \tilde{\textbf{x}}^{(t)} $ is generated \citep{fischer2014trainingRBMintro}, this work uses the traditional CD method. Our goal is to improve the RBM training by adaptively choosing the connectivity, and this is agnostic to the sampling method. Thus, improving such approximation is marginal to the main theme of this work. 

	Calculating the corresponding expectations for each model parameter 
	$ w_{ij}, b_i, d_j $, the resulting parameter update rules are given by: 
	\begin{align}
		\label{eq:updateW}
        \textbf{W} &\leftarrow \textbf{W} + \alpha \frac{1}{|\mathcal{B}|} \sum_{t \in \mathcal{B}} \left( \hat{\textbf{h}}(\textbf{x}^{(t)}){\textbf{x}^{(t)}}^\textsc{t} - \hat{\textbf{h}}(\tilde{\textbf{x}}^{(t)} )\mbox{$ \tilde{\textbf{x}}^{(t)} $}^\textsc{t} \right) \\ 
        \label{eq:updateb}
		\textbf{b} &\leftarrow \textbf{b} + \alpha \frac{1}{|\mathcal{B}|} \sum_{t \in \mathcal{B}} \left( \hat{\textbf{h}}(\textbf{x}^{(t)}) - \hat{\textbf{h}}(\tilde{\textbf{x}}^{(t)} ) \right) \\
        \label{eq:updated}
		\textbf{d} &\leftarrow \textbf{d} + \alpha \frac{1}{|\mathcal{B}|} \sum_{t \in \mathcal{B}} \left( \textbf{x}^{(t)} - \tilde{\textbf{x}}^{(t)} \right)
	\end{align}
    where $ \alpha >0$ is the learning rate hyperparameter and $ \hat{\textbf{h}}\left(\textbf{x}\right) $ is a vector of size $ H $ such that $ \hat{h}_i\left(\textbf{x}\right) = P(h_i = 1 |\textbf{x} ) = \sigma(b_i + \textbf{W}_{i\cdot}\textbf{x}) $, with $ \sigma(y) = \frac{1}{1 + e^{-y}} $. 
	
\section{Network Connectivity Gradient}
\label{sec:ncg}
    The classic RBM considers a fully connected network between its input and hidden layers. However, this is not necessarily the best connectivity pattern for the model to learn a particular task, prompting the investigation of other patterns. 
	
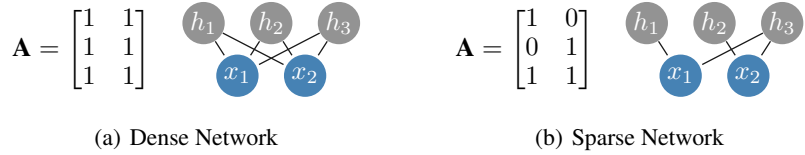
\begin{figure}[h]
		\begin{center}
			\def\layersep{10mm}
\def\circlesize{20pt}
\def\circdistscale{1}

	\begin{tabular}{cc}
		$ \textbf{A} = \left[ \begin{matrix} 
			1 & 1 \\
			1 & 1 \\
			1 & 1 \\
		\end{matrix} \right] $ & 
		\adjustbox{valign=c}{\begin{tikzpicture}[shorten >=1pt,->,draw=black!90, node distance=\layersep]
				\tikzstyle{every pin edge}=[<-,shorten <=1pt]
				\tikzstyle{neuron}=[circle,fill=black!25,minimum size=\circlesize,inner sep=0pt]
				\tikzstyle{visible unit}=[neuron, fill=steelblue];
				\tikzstyle{hidden unit}=[neuron, fill=gray!85];
				
				\foreach \name / \x in {1,...,3}
				\node[hidden unit] (h-\name) at (\circdistscale*\x, 0) {\color{white} $ h_{\name} $};
				
				\foreach \name / \x in {1,...,2}
				\node[visible unit] (x-\name) at (\circdistscale*\x+\circdistscale*.5, -\layersep) {\color{white} $ x_{\name} $};
				
				\foreach \source in {1,...,2}
				\foreach \dest in {1,...,3}
				\path[-, line width=0.5pt] (h-\dest) edge (x-\source);
		\end{tikzpicture}}
		\\
		& \\
	\end{tabular} 
	\begin{tabular}{cc}
		$ \textbf{A} = \left[ \begin{matrix} 
			1 & 0  \\
			1 & 1  \\
			0 & 1  \\
		\end{matrix} \right] $ & 
		\adjustbox{valign=c}{\begin{tikzpicture}[shorten >=1pt,->,draw=black!90, node distance=\layersep]
				\tikzstyle{every pin edge}=[<-,shorten <=1pt]
				\tikzstyle{neuron}=[circle,fill=black!25,minimum size=\circlesize,inner sep=0pt]
				\tikzstyle{visible unit}=[neuron, fill=steelblue];
				\tikzstyle{hidden unit}=[neuron, fill=gray!85];
				
				\foreach \name / \x in {1,...,3}
				\node[hidden unit] (h-\name) at (\circdistscale*\x, 0) {\color{white} $ h_{\name} $};
				
				\foreach \name / \x in {1,...,2}
				\node[visible unit] (x-\name) at (\circdistscale*\x+\circdistscale*.5, -\layersep) {\color{white} $ x_{\name} $};
				
				\path[-, line width=0.5pt] (h-1) edge (x-1)
				(h-2) edge (x-2)
				(h-2) edge (x-1)
				(h-3) edge (x-2);
		\end{tikzpicture}}
		\\
		& \\
	\end{tabular}
			\caption{Two examples of adjacency matrices $ \textbf{A} $ and the corresponding RBM networks: the classic fully connected network (top), and a model with the connections $ (x_1, h_2) $ and $ (x_2, h_1) $ suppressed (bottom).}
			\label{fig:inibicao}
		\end{center}
	\end{figure}

	Let $ \textbf{A} \in \mathbb{B}^{H,X} $ denote a binary matrix that represents a given connectivity pattern for the RBM, in the sense that $a_{ij} = \textbf{A}[i,j] = 1$ if hidden unit $h_i$ is connected to visible unit $x_j$, or $a_{ij} = \textbf{A}[i,j] = 0$ otherwise. 
	Figure \ref{fig:inibicao} shows examples of the adjacency matrix $\textbf{A}$ for two connectivity patterns. Note that it is generally intractable to enumerate them even in the case of small models ($2^{HX}$ possibilities of $\textbf{A}$). 
	In order to incorporate $\textbf{A}$ into the model, the weights in matrix $\textbf{W}$ must be zero on entries where a connection is not present. Thus, let $ \textbf{C} = \textbf{W} \odot \textbf{A}$ denote the acting weights of the model where $\odot$ is the element-wise matrix product such that $ c_{ij} = \textbf{C}[i,j] = w_{ij}a_{ij} $. The classic model parameters can be learned as before by using matrix $ \textbf{C} $ instead of $\textbf{W}$ to compute the gradients. 
    
    The novelty of the proposed method lies in computing a gradient for each possible element (connection) of $ \textbf{A} $. This can be analytically derived as with the other RBM parameters, where $\theta$ in equation \eqref{eq:CDapprox} also includes $ \textbf{A} $. The new set of parameters is equivalent to $\theta = (\textbf{W}, \textbf{A}, \textbf{d}, \textbf{b})$. 
    In particular, the expectation over the energy gradient is given by 
	\begin{equation}\label{eq:expectationA}
	\begin{aligned}
			\mathbb{E}_\textbf{h}\left[ \nabla_{a_{ij}} E(\textbf{x}, \textbf{h} ) \left| \textbf{x} \right. \right]  
			&= \mathbb{E}_{\textbf{h}} \left[ \left. \nabla_{a_{ij}} (- h_i a_{ij} w_{ij} x_j) \right| \textbf{x} \right] \\
			&= \mathbb{E}_{\textbf{h}} \left[ \left. - h_i w_{ij} x_j \right| \textbf{x} \right] \\
			&= - P_\theta(h_i = 1 | \textbf{x}) w_{ij} x_j \\ 
			&= - \sigma(\textbf{C}_{i\cdot}\textbf{x} + b_i) w_{ij} x_j  \\
			&= - \hat{h}_i(\textbf{x}) w_{ij} x_j
	\end{aligned}
	\end{equation}
    where $ \textbf{C}_{i\cdot} $ is the $ i $-th row in matrix $ \textbf{C} $. Note that under NCG, $\hat{\textbf{h}}(x)$ is now given by $\hat{h}_i\left(\textbf{x}\right) = \sigma(b_i + \textbf{C}_{i\cdot}\textbf{x})$. 
    
    This expectation is used to calculate the gradient (Equation \eqref{eq:CDapprox}). 
    Note that the gradient for a connection $ (x_i, h_j) $ can be non-zero even when $a_{ij} = 0$. 
    This is a key aspect of the methodology here proposed. It provides a gradient for absent connections and, consequently, the possibility for them to be enabled (or permanently disabled). 
    Furthermore, the weight $w_{ij}$ of a connection is not modified when the connection is deactivated ($a_{ij} = 0$ forces the gradient of $w_{ij}$ to zero without altering the weights themselves), which means that the RBM does not need to restart training from scratch if the connection were to be re-activated. 
	
	However $ a_{ij} $ is binary, and thus the usual continuous optimization framework that leverages the gradient to update its value does not apply. To circumvent this limitation, a continuous parameter denoting the connectivity strength is introduced in the model, and represented by $ \textbf{A}' \in [0,1]^{H,X}$ such that $ 0 \le a'_{ij} = \textbf{A}'[i,j] \le 1 $. Thus, the connection strength can be updated using the corresponding gradient (but saturating at 0 or 1), and the binary connection becomes a function of the connection strength. In particular, a simple threshold (step) function is used to determine the presence or absence of a connection. This idea leads to the following two-step update rule for the connection parameters: 
	\begin{equation}\label{eq:updateA}
		\begin{aligned}
			a'_{ij} \leftarrow& a'_{ij} + \frac{\alpha_A}{|\mathcal{B}|} \sum_{t \in \mathcal{B}} \left[ \hat{h}_i(\textbf{x}^{(t)}) w_{ij} x_j^{(t)} - \hat{h}_i(\tilde{\textbf{x}}^{(t)}) w_{ij} \tilde{x}^{(t)}_j \right] \\
			a_{ij} \leftarrow& \mathds{1}\left[ a_{ij}' \ge \gamma \right]
		\end{aligned} \,\,,
	\end{equation}
    where $ \gamma $ is the hyperparameter that denotes the threshold for enabling/disabling a connection based on the connection strength, $ \mathds{1}[\cdot] $ corresponds to the indicator/step function, and $ \alpha_A $ to the connectivity learning rate. Though not explicitly stated in the equation, when updated $a'_{ij}$ is saturated between $0$ and $1$. That is, after updating $a'_{ij}$ as shown in equation \eqref{eq:updateA}, the following rule is applied: 
    \begin{equation} \label{eq:saturationA}
        a'_{ij} \leftarrow \max \left\{ 0, \min \left\{ a'_{ij}, 1 \right\} \right\} \,\,.
    \end{equation}
    
\begin{figure}[h]
        \centering
        \includegraphics[width=\linewidth]{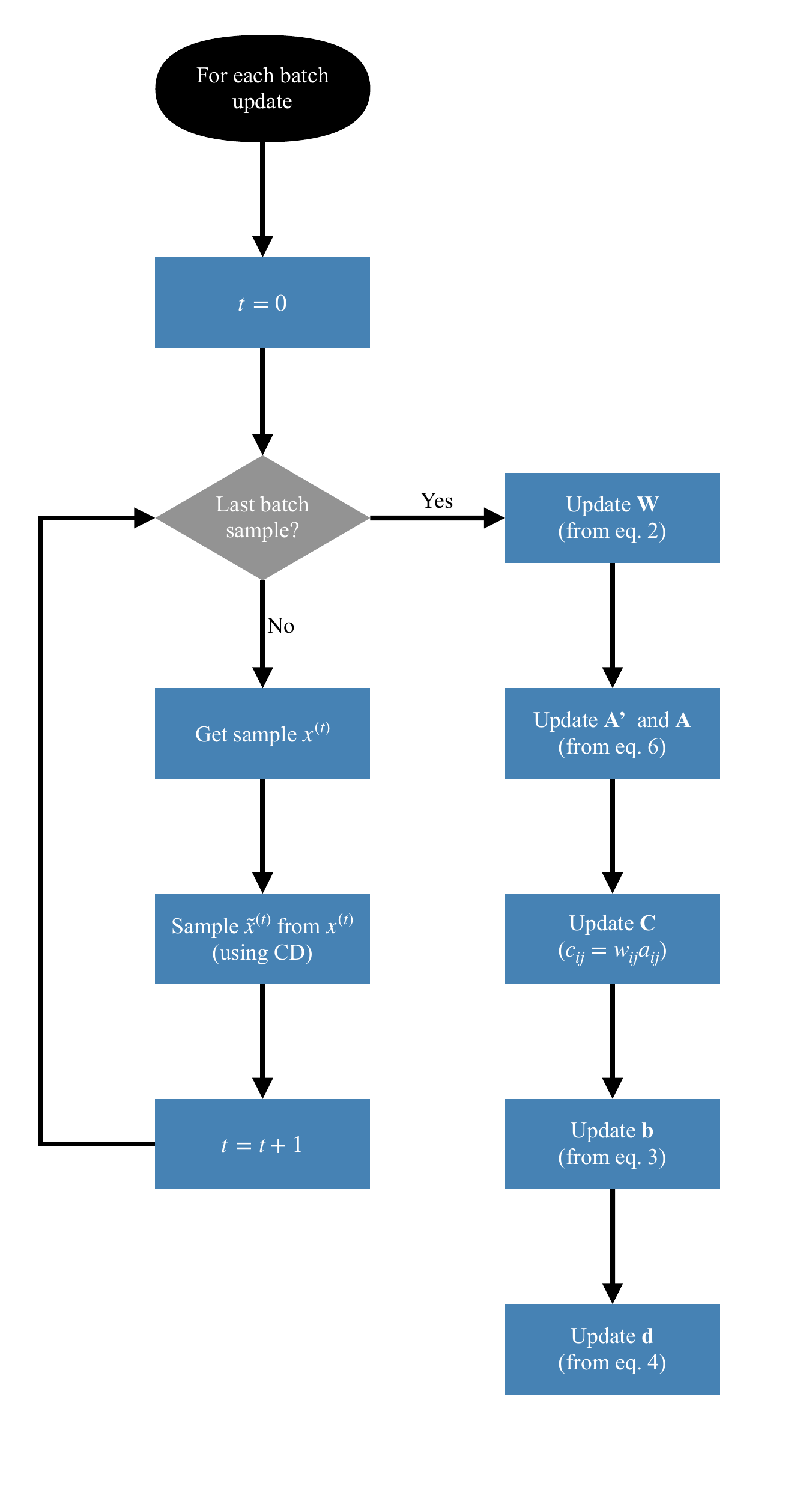}
        \caption{Algorithmic flowchart of the NCG batch update. For each batch sample $x$, a random $\tilde{x}$ sample is generated. The combined samples (batch and generated) are used to update each model parameter according to equations \labelcref{eq:updateW,eq:updateb,eq:updated,eq:updateA}.}
        \label{fig:flowchart}
    \end{figure}

    Note that the connectivity strength $a'_{ij}$ is updated according to the network parameters, being changed even when the connection is deactivated (that is, $a_{ij} = 0$). The strength, in turn, drives the connectivity itself, determining whether the connection should be activated (if the strength is greater than $\gamma$) or deactivated (if the strength is smaller than $\gamma$). 
    A flowchart of the full batch update of the NCG method is displayed in Figure~\ref{fig:flowchart}. 
    
    The method is called Network Connectivity Gradient (NCG) and jointly learns the connectivity pattern and classic model parameters for the RBM. Note that $ \alpha_A $ allows to decouple the learning rate of model parameters from the connectivity pattern, which can lead to better learning curves, as shown in Section \ref{sec:result:acc:learningRate}. 
    
	\subsection{Connectivity initialization} \label{sec:ncg:sec:init}
	A fundamental aspect of continuous optimization frameworks such as SGD is the initialization of the parameters that must be optimized. Being parameters, the connectivity pattern and the connection strength must also be initialized. 
	While the fully connected network is a possible initialization, intuitively it may not be the best pattern to start the optimization since it may take too many iterations to remove connections. 
	A common initialization method in the context of RBM (and other models) is choosing random (and small) values for the parameters. Thus, by following this method 
	each possible connection is initialized as active ($ a_{ij} = 1 $) with probability $ p $ or inactive ($ a_{ij} = 0 $) with probability $ 1 - p $.
	 Intuitively, $p$ will influence the learning performance of the RBM, since large/small $p$ can lead to dense/sparse networks that may require many iterations to evolve. Thus, $p$ is a hyperparameter of the initialization procedure. 
	
	Once the initial connection pattern has been determined, the connection strengths must also be defined. While initializing $a'_{ij} = a_{ij}$ is a possible initialization, this leads to connection strengths that are either 0 or 1, which may require too many iterations in order to cross the threshold to enable or disable the connection, respectively. 
	To avoid this cold start, connection strengths are randomly initialized as follows: 
	\begin{equation} \label{eq:initA'}
		a'_{ij} = \text{U}(0, \gamma)\,\,\mathds{1}\left[ a_{ij} = 0 \right] + \text{U}(\gamma,1)\,\,\mathds{1}\left[ a_{ij} = 1 \right] , 
	\end{equation}
	where $ \text{U}(a,b) $ is the continuous uniform random value in the interval $[a,b]$. Note that the random value of the connection strength depends on the threshold $\gamma$ for enabling/disabling the connection. Intuitively, a random value is chosen in the segment corresponding to the connection being absent (range $[0,\gamma]$) or present (range $[\gamma,1]$).

\section{Empirical Evaluation}
\label{sec:result}
	Two tasks will be considered to evaluate the learning performance of the RBM under the NCG methodology: sample generation (average NLL is the performance metric) and sample classification (accuracy is the performance metric). The main data set used is MNIST\,\footnote{
		Data set available at \url{http://yann.lecun.com/exdb/mnist/}.
	}, a frequently used benchmark in computer vision and the RBM literature~\citep{fischer2014trainingRBMintro}. 
	Also, two data sets from the UCI Evaluation Suite~\citep{UCIdata} are used for the classification task: Mushrooms and Connect-4\,\footnote{
		Data sets available at \url{https://www.csie.ntu.edu.tw/~cjlin/libsvmtools/datasets/}
	}. Within the UCI Evaluation Suite, the classification task in the 
    Mushrooms data set is relatively easy while classification in the Connect-4 dataset is seemingly difficult. These were chosen to elucidate the performance improvement of NCG in scenarios with different difficulties.  
	
\begin{figure}[h]
		\begin{center}
			\input{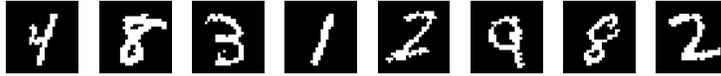}
			\caption{Examples of MNIST data set images after conversion to binary.}
			\label{fig:mnist}
		\end{center}
	\end{figure}

	The MNIST data set consists of gray-scale square images of $28 \times 28$ pixels. Each image contains a handwritten digit. 
	It has two separate sets of data: a train set, with $ 60\,\text{k} $ samples; and a test set, which has $ 10\,\text{k} $ samples. 
	The images were converted into black and white in order to be directly used as input to the RBM. The conversion was probabilistic such that each pixel was assigned a black color with probability proportional to its darkness (gray-scale) in the original image, a methodology commonly adopted~\citep{AIS, iRBM}. Some examples of the resulting data set are shown in Figure \ref{fig:mnist}.
	
	For MNIST, each image in the data set has 784 pixels, each of which corresponds to a visible unit of the RBM. The experiments used 500 hidden units, and training was achieved using CD with 10 steps of Gibbs sampling (CD-10). The learning rate for the model parameters was set to $ \alpha= 0.1 $ and mini-batches of 50 random samples. The connectivity learning rate was $ \alpha_A = 0.5 $, unless otherwise specified. 
	No momentum or weight decay was used. The RBM weight parameters were initialized with null biases and small random weights, uniformly distributed in the interval $ [-1, 1] $. For the connection threshold in NCG, $\gamma = 0.5$ is the default value as this is the midpoint value in the possible range for the connection strengths, not favoring either a more sparse ($\gamma > 0.5$) or dense network ($\gamma < 0.5$). In what follows, results with different values for $\gamma$ are also presented supporting its default value. 
	During training, one epoch corresponds to one iteration over the entire training data set, with the model's parameters being updated at every batch. Since the batch size was 50 data samples, an epoch corresponds to 1200 parameter updates necessary to iterate over the $ 60\,\text{k} $ samples in this data set. Batch elements are randomly determined for every epoch. 
	
	The Mushrooms data set contains characteristics of different types of mushrooms, subdivided into edible and poisonous categories. There are 21 attributes, converted into 112 binary features (each one associated with a visible unit, so that $X=112$), and 8124 samples (subdivided into $ 2\,\text{k} $ for training the rest for testing). 
	Finally, the Connect-4 data set contains board situations for the game of Connect-4, labeled by whether the first player wins, loses, or there is a draw. There are 67557 samples (with $ 16\,\text{k} $ separated for training), each with 42 board spaces. The board spaces were converted into 126 binary features ($X=126$). 
	
	Experiments for the UCI Evaluation Suite data sets used 100 hidden neurons, a batch of 10 random samples, $ \alpha = 0.01 $ and $ \alpha_A = 0.05 $, unless otherwise specified. 
	Other hyperparameters were the same as specified for MNIST. 
	
	Although fine-tuning these hyperparameters could improve the learning performance of the RBM, the goal of this work is to ascertain the improvement that the application of NCG can bring to the model, regardless of having the best possible set of hyperparameters for the task and instance at hand.

	\subsection{Generative Results}
	\label{sec:result:nll}
		In the sample generation task, a classic generative RBM is trained to generate random samples similar to the input examples. The performance is assessed using the average NLL across the training set. Since the exact average NLL cannot be computed due to the intractability of the normalization constant, the Annealed Importance Sampling (AIS) method is used as an approximation, as proposed by \cite{AIS}, with 100 runs and $14.5\,\text{k}$ intermediate distributions. Note that all experiments on this task are evaluated using the MNIST data set. 
		
\begin{figure}[h]
			\begin{center}
					\includegraphics[trim={4mm 2mm 0 4mm}, clip, width=1\linewidth]{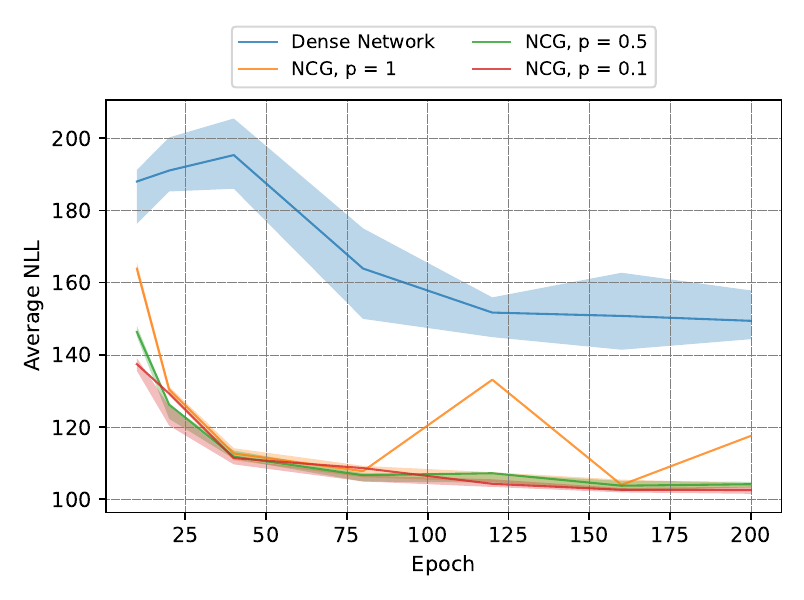}
				\caption{
					Average NLL over the training epochs -- lines are the means and shades the quartile uncertainty over 10 experiments. 
				}
				\label{fig:nllMNIST_nll}
			\end{center}
		\end{figure}

		Figure \ref{fig:nllMNIST_nll} shows the learning curve (evolution of the average NLL over the epochs) for the classic fully connected RBM and three initializations of the NCG method. 
		Clearly, the fully connected network exhibits a significantly inferior learning curve, both in terms of sample mean and variance. Interestingly, the three different initializations for NCG exhibit a very similar performance (with the exception of a few outliers for the case $p=1$)\footnote{
			While training NCG with $ p = 1 $, two of the 10 runs exhibited much higher than average NLL at epoch 120 and one of the 10 runs at epoch 200.
		}. While the mean performance for $p=0.1$ could be said to be slightly better, the overlapping quartiles show that the sparsity of the random initialization is not particularly important in this scenario. Indeed, the similar learning curves indicate that NCG can find effective networks independently of the (random) initial connectivity, and is therefore robust to connectivity initialization. 
		
		\begin{figure*}[t]
			\begin{center}
				\include{connectivity_nllMnist}
				\caption{
					Degree statistics (minimum, average, maximum) of the hidden units over the NCG training epochs -- lines are the sample mean and shades the sample quartile over 10 experiments. 
				}
				\label{fig:nllMNIST_conn}
			\end{center}
		\end{figure*}
		
		\begin{figure}[h]
			\begin{center}
				\includegraphics[trim={0 0 1cm 1cm}, clip, width=.8\linewidth]{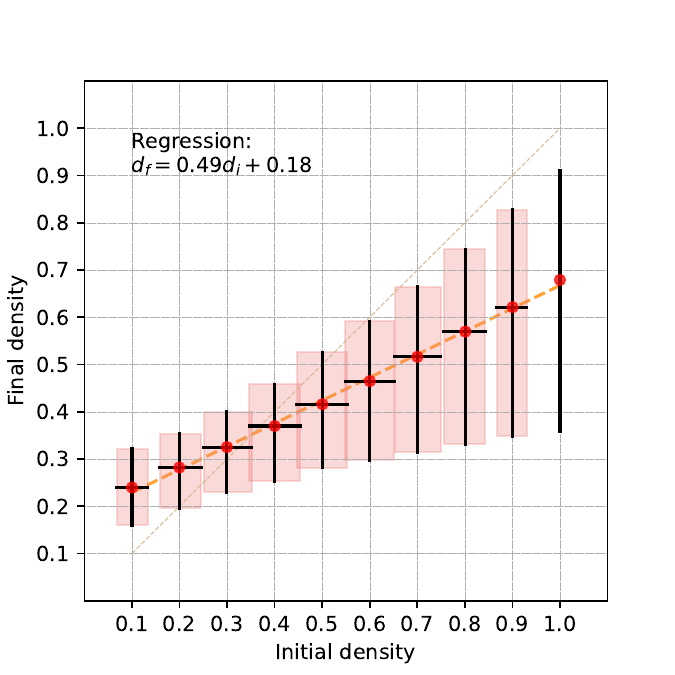}
				\caption{
					Comparison between initial and final fraction of active connections for  NCG training after 10 epochs -- the dots are the mean over 25 experiments, the shades the uncertainty (minimum and maximum values) and the orange line the linear fit.  
				}
				\label{fig:density}
			\end{center}
		\end{figure}
		
		Despite the similar learning curves, the evolution of the network degrees\footnote{The degree of a unit is the number of active connections it has with the other layer. Different units can have different degrees.} is very different across the different initializations. Figure \ref{fig:nllMNIST_conn} shows the evolution of the maximum, minimum, and average degree of the hidden units for the three initializations. For $p=1$ a sharp decrease is observed in all three statistics in the first 10 epochs, with the curves indicating a slight decay even after 200 epochs, while for $p=0.5$, the initial decrease is not as strong and the curves seem closer to convergence. Interestingly, the case $p=0.1$ shows an increase in all three statistics in the first 10 epochs and convergence after 200 epochs. This shows that NCG can not only prune connections but also {\em add} connections when the network is too sparse. 
		The similar learning curves but different network patterns indicate that the joint optimization of model parameters and network connectivity can compensate for one another, leading to similar performance even when the connectivity pattern is significantly different. 
		Indeed, the literature on network pruning suggests that different network patterns can often achieve similar performance \citep{pruningSurvey}. 
		
		\begin{figure}[h]
			\begin{center}
				\includegraphics[width=\columnwidth]{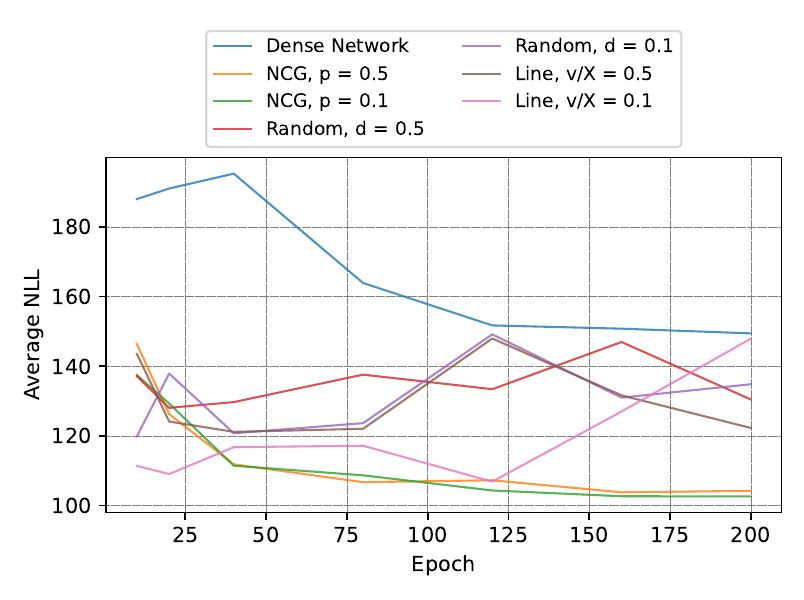}
				\caption{
					Average NLL over the training epochs -- lines are the mean values  over 10 experiments. Comparison of NCG with the Line and Random patterns. 
				}
				\label{fig:static_mnist_nll}
			\end{center}
		\end{figure}
		
		Further insight is provided by Figure \ref{fig:density}, which shows the relationship between initial network density (a hyperparameter) and final density. Note that NCG tends to increase the number of connections when the initial network is too sparse (up to 30\%) and decrease the number of connections when the initial network is too dense (40\% or more). Note that the randomness of the final density (vertical bars) is much larger than that of the initial density (horizontal bars), as the final density depends on the optimization. Interestingly, the average final and initial densities follow an almost perfect linear relationship, with its slope being approximately 0.5, as indicated in Figure \ref{fig:density}. 
		
		It is known that sparse networks can often outperform their dense counterparts. Indeed, even simple fixed patterns such as the line pattern (which connects each hidden unit to $ v $ consecutive visible units -- that is, a hidden unit $h_i$ is connected to visible units $x_j$ such that $i \le j \le i+v$) can achieve better learning curves \citep{selfAnonymous}. Figure \ref{fig:static_mnist_nll} shows the learning curves of NCG together with both the line pattern and a random pattern (the latter corresponds to a bipartite Erdös-Rényi random graph, created by applying a $d$ probability of each connection being present in the network). 
        Note that these patterns are intrinsically different because the first has pre-determined and deterministic connections, while the latter has stochastic connectivity. 
        For fairness of comparison, RBMs with the same (average) number of initial connections are created for all patterns: 50\% and 10\% densities. Note that NCG shows better performance than any other pattern after 50 training epochs (for both initializations) while also showing less noisy learning curves. Interestingly, all patterns outperformed the fully connected RBM. 
		
		\begin{figure}[h]
			\begin{center}
				\includegraphics[width=\columnwidth]{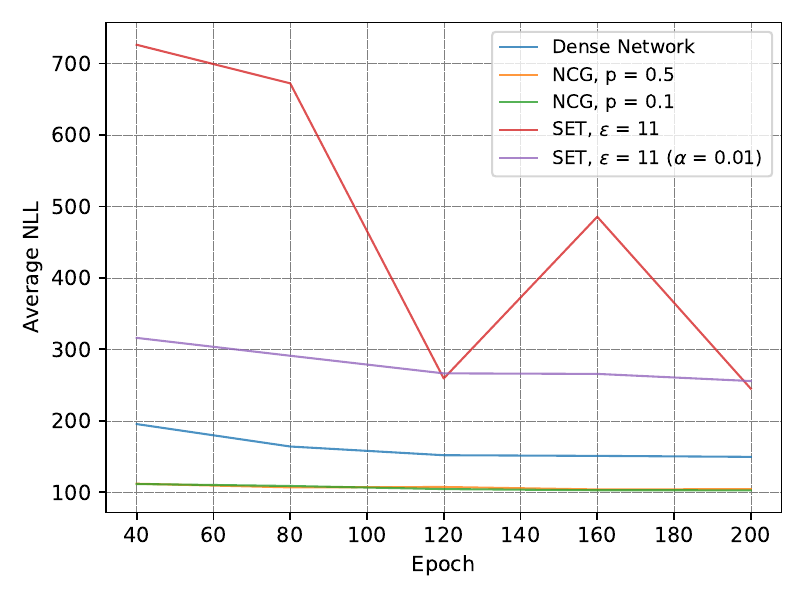}
				\caption{
					Average NLL over the training epochs -- lines are the mean values  over 10 experiments and shades are the quartiles. Comparison of NCG with the SET method~\cite{SET_mocanu}. 
				}
				\label{fig:setComp}
			\end{center}
		\end{figure}
		
		Lastly, Figure \ref{fig:setComp} shows a direct comparison between NCG and SET~\citep{SET_mocanu}, trained with 2500 hidden units, sparsity parameter $ \epsilon = 11 $ and fraction of removed edges $ \zeta = 0.3 $ (parameters reported in the original paper for the MNIST dataset, using their publicly available code). 
        For the learning rate, experiments were performed with both $ 0.1 $ and $ 0.01 $ values. The first allows for a direct comparison with the NCG experiments (same learning rate) and the second reduces the fluctuations during learning. However, note that SET did not outperform the fully connected network (on this task), and was thus significantly outperformed by NCG. Moreover, SET induced a very sparse network (see Table \ref{tab:results}) which probably hindered its performance. 
        
        Table \ref{tab:results} shows a summary of the generative performance and sparsity obtained by each reported method after 200 training epochs. Note that both NCG initializations exhibit the lowest NLL across all methods while also having different final densities with respect to its initial density.  
        
        Plots with the sample quartiles can be found in Appendix \ref{apd:quartile}. Further comparisons with the SET method are also available in Appendix \ref{apd:appendix_SET}. 
        This includes the learning curves of both SET and NCG using the same hyperparameters ($H=500$ and the same sparsity values). 
        Under these conditions, SET yielded poorer performance, which is the reason that the results reported in this section use the hyperparameters reported in its original paper \cite{SET_mocanu}. 
        Furthermore, Appendix \ref{apd:biggerH} shows an evaluation of the effect of increasing the number of hidden neurons $H$. Results indicate that increasing the size of the model leads to worse performance for a fixed number of training epochs. Intuitively, larger models require more training epochs to achieve the same NLL performance. 
        Lastly, some examples of samples generated by the trained RBMs are portrayed in Appendix \ref{apd:genSamples}.

	\subsection{Classification Results}
	\label{sec:result:acc}
		In the classification task, the RBM is trained to classify the given input (the digit in the image, for MNIST). The RBM is expanded to have additional visible (input) units in order to encode the label of the image during training, as per done in the RBM literature~\citep{fischer2014trainingRBMintro, classRBM}. There is one extra visible unit per class, and only one is activated for each input sample, corresponding to the sample class (the image digit, from 0 to 9, for MNIST). The connections between hidden units and the label units are fixed and not subjected to optimization, as they are crucial for the classification task. Moreover, the RBM is trained using Contrastive Divergence as in the generative task, and is not a priori aware of the classification task (no changes are made to the objective function). 
		
		While the generative task relied on the approximate NLL to measure the learning performance of the RBM, the classification task used the accuracy as performance, given by the fraction of data samples correctly classified. For MNIST, a sample corresponds to an image. Classification is performed by presenting the image to the input layer of the RBM, setting all label units to $ 0.5 $, calculating the probabilities of each hidden unit being activated, and finally selecting the label unit (digit) with the higher probability of being activated. This digit is the predicted label for the image. 
		
\begin{figure*}[t!]
			\vskip -2mm
			\begin{center}
				\subfloat[MNIST, Test set]{ \label{fig:MNISTacc}
					\includegraphics[width=.315\linewidth]{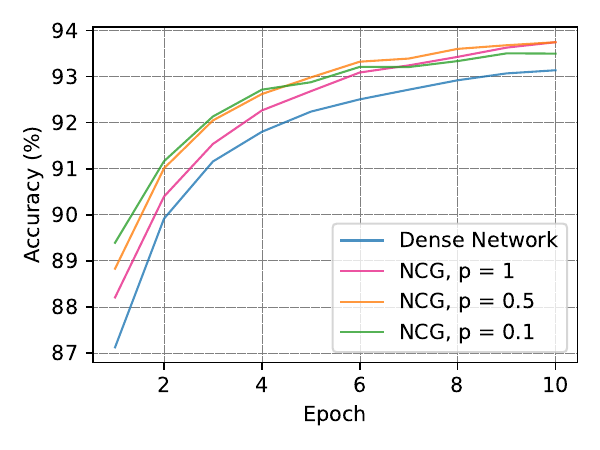}
				}
				\subfloat[Mushrooms, Test set]{ \label{fig:MUSHacc}
					\includegraphics[width=.315\linewidth]{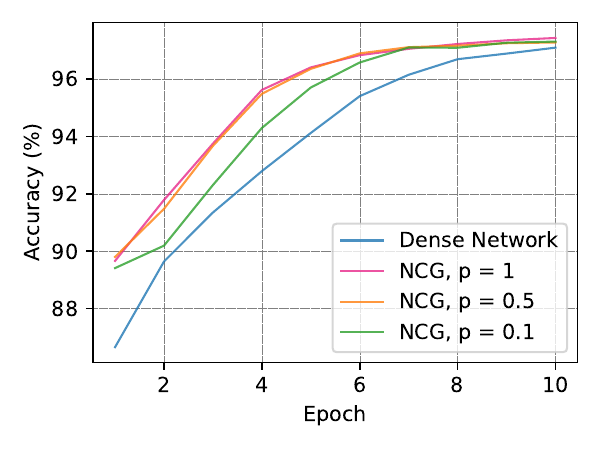}
				}
				\subfloat[Connect-4, Test Set]{ \label{fig:CON4acc}
					\includegraphics[width=.315\linewidth]{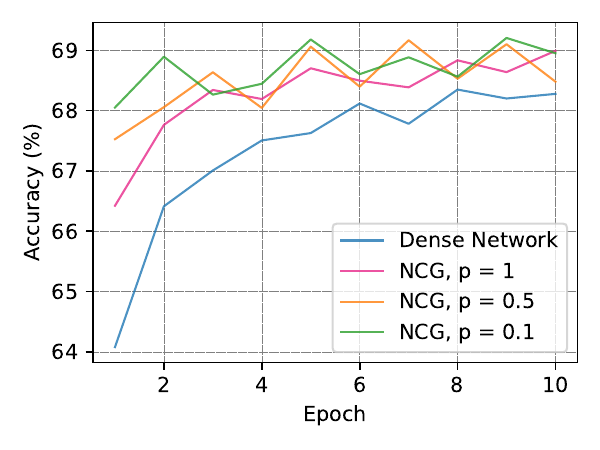}
				} 
				\caption{
					Test set classification accuracy for NCG and the classical (dense) RBM for the MNIST, Mushrooms and Connect-4 data sets; lines are the sample means over 25 runs. 
				}
				\label{fig:class_main}
			%
				\subfloat[MNIST, Test set]{ \label{fig:class_mesmaTaxa_MNIST}
					\includegraphics[trim={0 4mm 0 3.5mm}, clip, width=.315\linewidth]{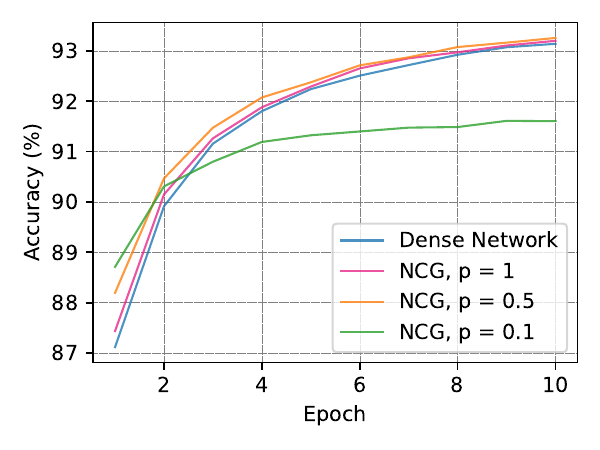}
				}
				\subfloat[Mushrooms, Test set]{ \label{fig:class_mesmaTaxa_mushrooms}
					\includegraphics[trim={0 4mm 0 3.5mm}, clip, width=.315\linewidth]{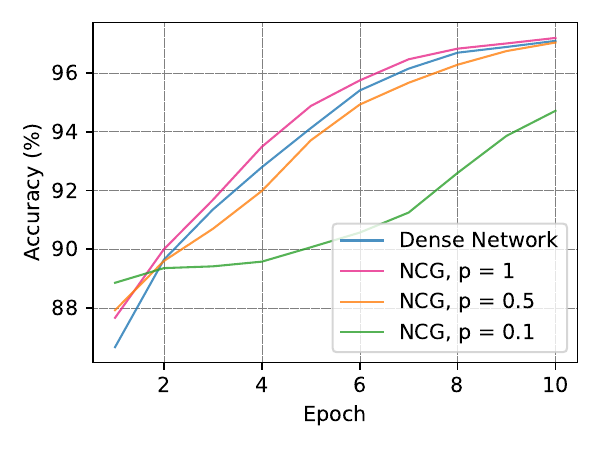}
				}
				\subfloat[Connect-4, Test set]{ \label{fig:class_mesmaTaxa_connect-4}
					\includegraphics[trim={0 4mm 0 3.5mm}, clip, width=.315\linewidth]{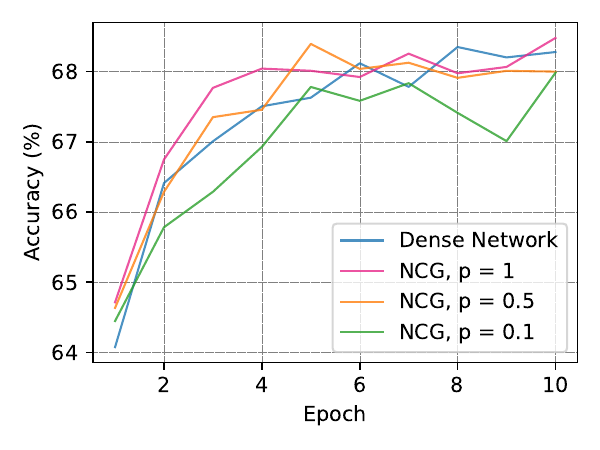}
				} 
				\caption{
					Classification accuracy over the training epochs of NCG for the MNIST, Mushrooms and Connect-4 data sets with $\alpha_A = \alpha$ ($ 0.1 $ for MNIST and $ 0.01 $ for the other data sets); lines correspond to the sample mean over 25 experimental runs. 
				}
				\label{fig:class_mesmaTaxa}
		%
				\subfloat[MNIST, Test set]{ \label{fig:class_gamma25_MNIST}
					\includegraphics[trim={0 4mm 0 3.5mm}, clip, width=.315\linewidth]{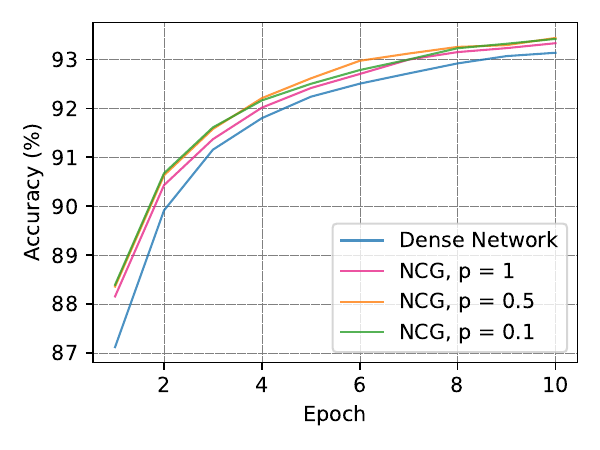}
				}
				\subfloat[Mushrooms, Test set]{ \label{fig:class_gamma25_mushrooms}
					\includegraphics[trim={0 4mm 0 3.5mm}, clip, width=.315\linewidth]{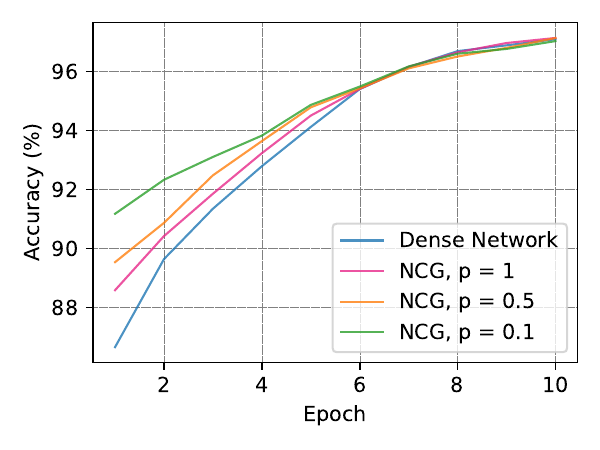}
				}
				\subfloat[Connect-4, Test set]{ \label{fig:class_gamma25_connect-4}
					\includegraphics[trim={0 4mm 0 3.5mm}, clip, width=.315\linewidth]{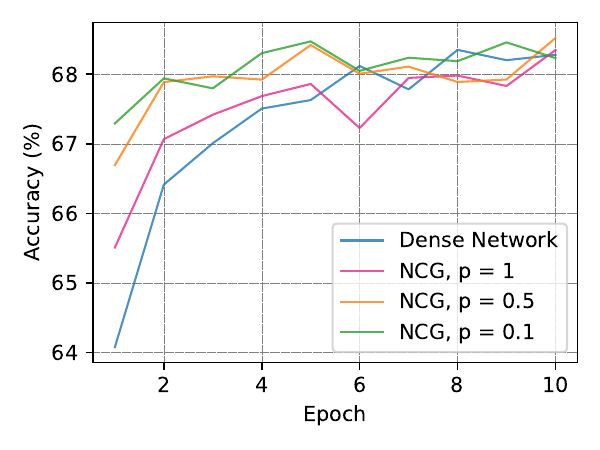}
				}
				\caption{
					Classification accuracy over the training epochs of NCG for the test and train sets of the MNIST, Mushrooms and Connect-4 data sets with $\gamma = 0.25$; lines correspond to the sample mean over 25 experimental runs. 
				}
				\label{fig:class_gamma25}
			%
				\subfloat[MNIST, Test set]{ \label{fig:class_gamma75_MNIST}
					\includegraphics[trim={0 4mm 0 3.5mm}, clip, width=.315\linewidth]{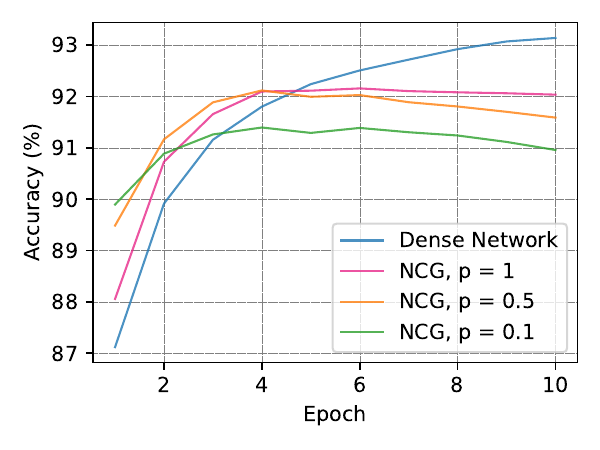}
				}
				\subfloat[Mushrooms, Test set]{ \label{fig:class_gamma75_mushrooms}
					\includegraphics[trim={0 4mm 0 3.5mm}, clip, width=.315\linewidth]{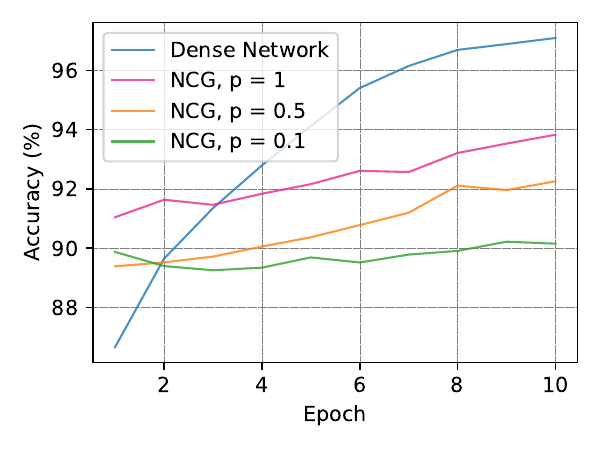}
				}
				\subfloat[Connect-4, Test set]{ \label{fig:class_gamma75_connect-4}
					\includegraphics[trim={0 4mm 0 3.5mm}, clip, width=.315\linewidth]{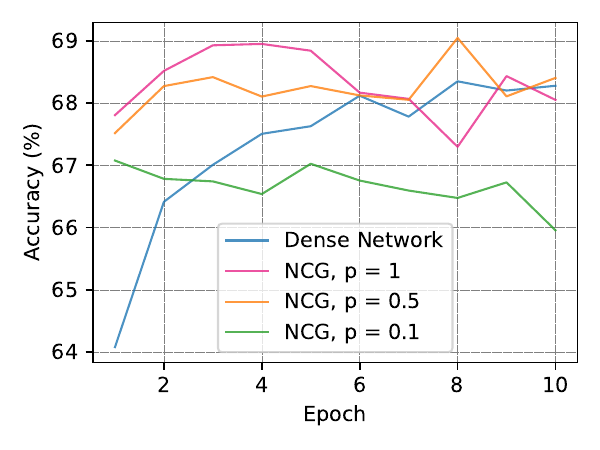}
				} 
				\caption{
					Classification accuracy over the training epochs of NCG for the test and train sets of the MNIST, Mushrooms and Connect-4 data sets with $\gamma = 0.75$; lines correspond to the sample mean over 25 experimental runs. 
				}
				\label{fig:class_gamma75}
			\end{center}
		\end{figure*}

		Figure~\ref{fig:MNISTacc} shows the evolution of the classification accuracy on the test set over the epochs for different NCG initializations (results for the train set can be found in Appendix \ref{apd:train}). 
		Note that all three initializations generate models that are consistently better than the fully connected RBM in the early stages of training. 
		
		Interestingly, Figure~\ref{fig:MNISTacc} shows that accuracy is inversely proportional to the initial density during the first epochs of training: initializing the network with fewer connections yields superior accuracy in early stages of training. However, as the number of epochs increases, the accuracy between the NCG models becomes more similar. This indicates that NCG is capable of overcoming a poorly initialized connectivity pattern by adjusting both the connections and model weights. Note that the degrees of the hidden units of different initializations are still very different after 10 epochs, as shown in Figure \ref{fig:density}, despite the similar accuracy performances. 
		
		Figure \ref{fig:class_main} shows the accuracy of these different models for the first 10 epochs of training for both the MNIST data set as well as Mushrooms and Connect-4 (uncertainty measures were removed from the plots to avoid visual clutter, but quartile information for these and the following plots can be found in Appendix \ref{apd:quartile}). 
        Note that only the first 10 epochs of training are shown because the RBM training is designed to minimize the NLL, even in the classification task. However, reducing the NLL does not necessarily improve the classification accuracy. Indeed, in all models considered (original RBM or NCG), accuracy performance started to decay after 10 epochs. Thus, since the focus of the classification task is accuracy, the results report performance only for this range. 
        
		For all data sets, results indicate that NCG has a superior performance, outperforming the dense network, in particular during the beginning of training. Note that the Connect-4 data set appears to be more difficult for RBMs, with increased training noise and smaller overall performance, while the Mushrooms data set is markedly easier, reaching the high accuracy values. Despite differences in the kind of data and difficulty of correct classification, NCG manages to maintain a higher performance than all the baselines considered. 
		
		A comparison of the RBMs classification performance (test set accuracy) and network sparsity after 10 training epochs is shown in Table \ref{tab:results}. 
        Across all three data sets, NCG shows better performance than the other connectivity patterns. In particular, NGC is the only method that manages to surpass the dense RBM performance in the classification task. 
		Plots with the accuracy on the training set over the epochs between NCG and the line and random pattern can be found in Appendix \ref{apd:conn_patterns}. 
		
		\begin{table*}[width=.97\textwidth,pos=h]
			\centering
			\caption{Performance of different methods on different tasks and data sets. For generative task, the average NLL and standard deviation on the test set over 10 experiments each having 200 training epochs is reported (lower is better). For the classification task, the average classification accuracy and standard deviation on the test set over 25 experiments each having 10 training epochs is reported (higher is better). The sparsity values are equivalent to the percentage of active connections of the trained model (the dense network has all connections active).}
			\label{tab:results}
			\begin{tabular*}{\tblwidth}{@{} LLLLLL @{} }
				\toprule
				\multirow{2}{*}{\bfseries Data set} & \multirow{2}{*}{\bfseries Model} & \multicolumn{2}{c}{\bfseries Generative} & \multicolumn{2}{c}{\bfseries Classification} \\
				& 										  & \textbf{NLL}                &   \textbf{Sparsity}             & \textbf{Accuracy}          & \textbf{Sparsity} \\
				\toprule 
				\multirow{6}{*}{MNIST}  & Dense Network & 149 $\pm$ 11 & 1 & 93.1 $\pm$ 0.3 & 1 \\
				& NCG p = 0.5 & 104 $\pm$ 1 & 0.300 $\pm$ 0.001 & \textbf{93.7 $\pm$ 0.3} & 0.4574 $\pm$ 0.0007 \\
				& NCG p = 0.1 & \textbf{103 $\pm$ 1} & 0.205 $\pm$ 0.001 & 93.5 $\pm$ 0.4 & 0.278 $\pm$ 0.001 \\
				& Line & 122 $\pm$ 8 & 0.5 & 93.0 $\pm$ 0.5 & 0.5 \\
				& Random & 130 $\pm$ 14 & 0.50 $\pm$ 0.25 & 92.5 $\pm$ 0.4 & 0.50 $\pm$ 0.25 \\
				& SET$^*$ & 244 $\pm$ 302 & 0.00627 $\pm$ 0.00003 & --- & --- \\
				& SET$^{*,\dagger}$ & 162 $\pm$ 2 & 0.00640 $\pm$ 0.00003 & --- & --- \\
				\midrule
				\multirow{6}{*}{Mushrooms}  & Dense Network & --- & --- & 97.1 $\pm$ 0.5 & 1 \\
				& NCG p = 0.5 & --- & --- & \textbf{97.3 $\pm$ 0.4} & 0.498 $\pm$ 0.004 \\
				& NCG p = 0.1 & --- & --- & \textbf{97.3 $\pm$ 0.5} & 0.270 $\pm$ 0.003 \\
				& Line & --- & --- & 96.3 $\pm$ 0.7 & 0.5  \\
				& Random & --- & --- & 96.6 $\pm$ 0.9 & 0.50 $\pm$ 0.25 \\
				\midrule
				\multirow{6}{*}{Connect-4}  & Dense Network & --- & --- & 68.3 $\pm$ 1.0 & 1 \\
				& NCG p = 0.5 & --- & --- & 68.5 $\pm$ 1.6 & 0.513 $\pm$ 0.003 \\
				& NCG p = 0.1 & --- & --- & \textbf{69.0 $\pm$ 1.8} & 0.343 $\pm$ 0.004 \\
				& Line & --- & --- & 67.5 $\pm$ 2.2 & 0.5 \\
				& Random & --- & --- & 67.0 $\pm$ 2.6 & 0.50 $\pm$ 0.25 \\
				\bottomrule
				\multicolumn{6}{l}{\tiny $^*$ SET method proposed by \cite{SET_mocanu} and run with their code implementation} \\
				\multicolumn{6}{l}{\tiny $^\dagger$ Method run for 480 epochs with learning rate $\alpha=0.01$ (other models in the MNIST data set report results for 200 epochs and $\alpha=0.1$)}
			\end{tabular*}
		\end{table*}

		\subsection{Learning Rate Analysis} 
		\label{sec:result:acc:learningRate}
			Empirical results showed that adopting a higher learning rate for the connectivity parameters allows the network connectivity to evolve faster in the early stages of training. Intuitively, this allows NCG to quickly adjust for poor initial network patterns before other model parameters start to converge. As discussed below, this decoupling of the learning rates (and being faster for network connectivity and slower for network weights and biases) is crucial for NCG. 
			
			Figure \ref{fig:class_mesmaTaxa} shows the classification accuracy when using the same learning rate for both the connectivity and the other model parameters, $ \alpha_A = \alpha = 0.1 $ for MNIST and $ \alpha_A = \alpha = 0.01 $ for the Mushrooms and the Connect-4 data sets. 
            The smaller learning rates were chosen for the UCI Suite data sets because, with them being smaller and less complex, their overall training showed good performance even when starting with such small learning rates. 
	        In comparison to Figure \ref{fig:class_main}, note the significant decrease in accuracy for all three initialization and all epochs of training (the dense network -- the blue line -- is the same across all experiments, it can be used as a reference to compare the performances across the plots). 
	        Moreover, only the networks initialized with all connections activated ($p=1$) managed to achieve higher mean accuracy than the traditional RBM (fully connected network) for all three data sets. This is in marked contrast to the previous results, for which NCG shows better performance than the traditional RBM even for initializations with 50\% and 10\% of the connections activated. 
			
			Interestingly, the initialization with 10\% of the connections activated ($ p = 0.1 $) often shows superior performance after the first epoch of training, having the best mean performance for both the MNIST and Mushrooms data sets. However, the model fails to continue improving its accuracy and falls behind the other models, including the fully connected network. Intuitively, the model cannot adjust its connection pattern fast enough and the connectivity gradient becomes subdued by other model parameters. This example highlights the importance of decoupling the learning rates when jointly optimizing network connectivity and other model parameters. 
		
	\subsection{Threshold Analysis}
	\label{sec:result:acc:threshold}
		Besides the initialization parameter $p$ and the connectivity learning rate $\alpha_A$, the connectivity threshold parameter $\gamma$, responsible for determining how strong a connection needs to be in order to be enabled, also plays a fundamental role. Therefore, experiments with different values of $\gamma$ were performed. For these experiments, the connectivity learning rate was set to the previous values ($\alpha_A = 0.5$ for MNIST and $\alpha_A = 0.05$ for the other data sets). 
		
		Figure \ref{fig:class_gamma25} shows the resulting accuracy throughout the epochs when using $\gamma=0.25$. Intuitively, a lower threshold would tend to generate a more dense network, since smaller connectivity strengths can activate a connection. Results in Figure \ref{fig:class_gamma25} corroborate this conclusion: in this regimen, the learning curves become more similar to the dense network learning curve. Intuitively, as the threshold decreases the corresponding learning curves would converge to the traditional RBM (dense network). Thus, NCG is not effective if the threshold is too small. 
		
		Figure \ref{fig:class_gamma75} on the other hand, shows the learning curve for a higher threshold, $\gamma = 0.75$. Note that this scenario induces sparser networks and also much lower classification accuracy. Intuitively, a higher threshold makes it more difficult for a connection to be activated and consequently more difficult to learn (due to high sparsity). As the threshold increases, the network will simply have no connections and nothing meaningful can be learned. Thus, NCG is not effective if the threshold is too large. In essence, a moderate threshold yields a good tradeoff between sparsity and learnability (e.g., accuracy) of the model.  

\section{Conclusion}
\label{sec:conc}
	This work presented Network Connectivity Gradient (NCG), a principled optimization method tailored to RBMs that learns the optimal connectivity network jointly with other model parameters (weights and biases). NCG computes gradients for each possible network connection given a connectivity pattern. The gradients are used to drive the continuous connectivity strength parameter that in turn determines to maintain, add, or remove the connection in each training epoch. NCG requires no change in RBM's objective function nor its classic optimization framework, and thus it requires no regularization to induce sparse networks. Evaluation of NCG on generative and classification tasks using the MNIST and other data sets demonstrated its effectiveness in learning better models (learning faster and better performance) than the dense RBM, other static connectivity patterns, and the SET method (also a dynamic method for network connectivity). Results also indicate that NCG is robust with respect to its (randomized) initialization for network connections and weights and biases. However, NCG's hyperparameters such as the connectivity learning rate ($\alpha_A$) and the threshold on connection strength ($\gamma$), play a fundamental role in the model's performance. 

    While NCG has been specifically designed for RBMs, its core ideas could be applied to other neural network models. In particular, NCG introduces two new matrices as part of the model parameters: $A$ (connection mask) and $A'$ (connection strength). In this work, the learning of these parameters, along with the learning of other model parameters, are specific to RBMs (Equations~\ref{eq:expectationA} and \ref{eq:updateA}). However, extending this idea to other neural network models would require specifying how these parameters (as well as other model parameters) can be learned, for example, by adapting backpropagation. 
    
    Finally, as with other recent works, network pruning during the initialization phase~\citep{lee2019snip,de2021progressive} might also be leveraged to design more effective initial networks for NCG. 

{
	\bibliographystyle{elsarticle-num-names}
	\bibliography{bibliography}
}

\printcredits

\section*{Declaration of competing interest}
This work received financial support through research grants from the
National Council for Scientific and Technological Development (CNPq)
and the Rio de Janeiro State Research Foundation (FAPERJ) both from
Brazil.

\section*{Data availability}
Data is available online, partially on the NCG repository itself (\url{https://github.com/AmieOliveira/NCG}).

{
	\bio{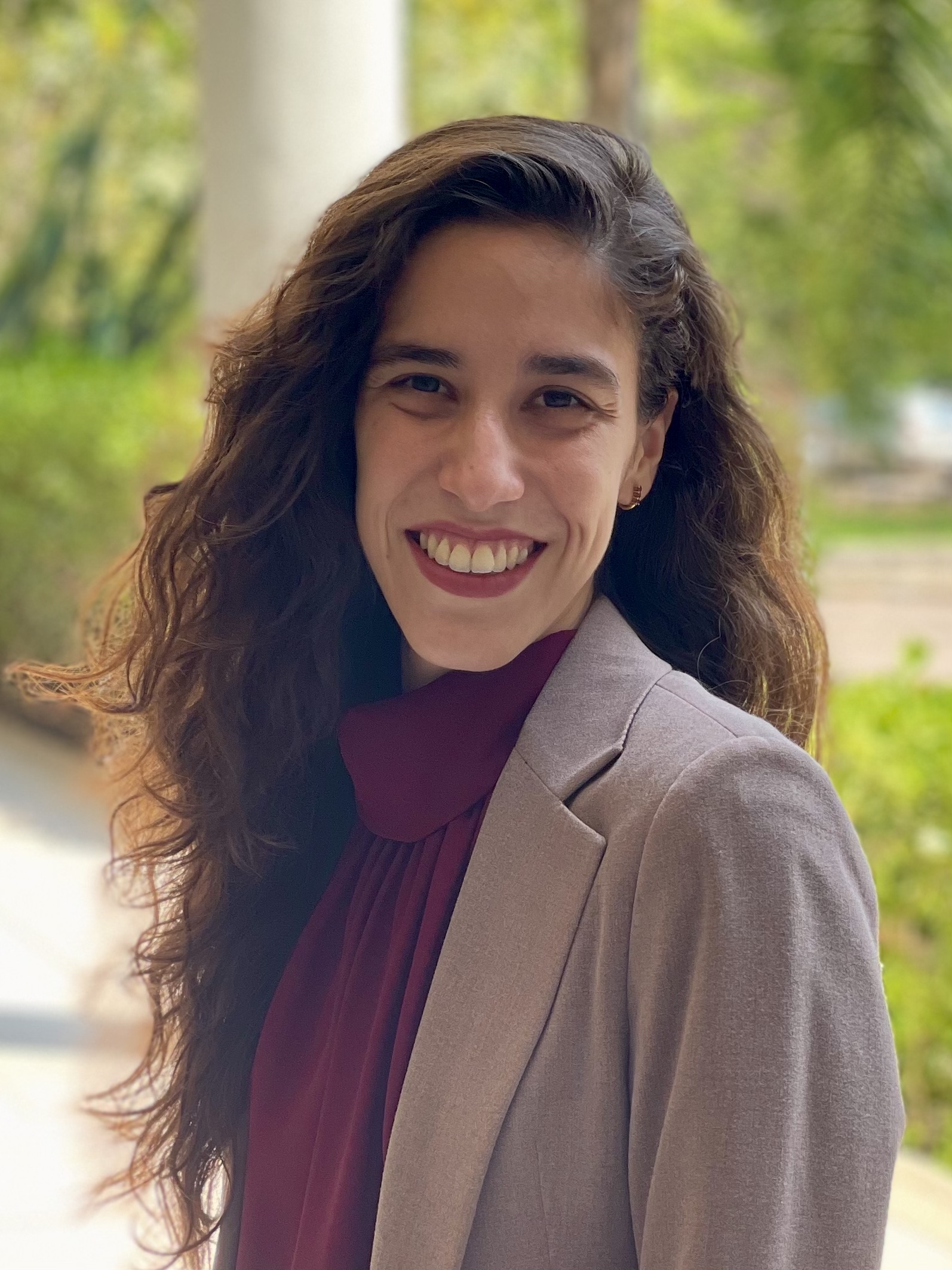} 
	\textbf{Amanda Camacho Novaes de Oliveira} is currently pursuing her D.Sc. degree in Computer Science and Systems Engineering at the Federal University of Rio de Janeiro (UFRJ), Brazil. In 2022, she received her M.Sc. degree in the same field, and in 2020 her Magma cum laude bachelor's degree in Control and Automation Engineering at UFRJ. In 2018 she did an internship at the European Organization for Nuclear Research (CERN) and in 2023 she spent time as a visiting scholar in the Computer Sciences department at the University of Massachusetts Amherst (UMass). 
	She was awarded a special scholarship during her masters and doctorate, in 2021  and 2024, respectively, the \enquote{Bolsa Nota 10} scholarship from FAPERJ.
	Her research interests include artificial intelligence, reinforcement learning, machine learning and automation. 
	\endbio
	
	\bio{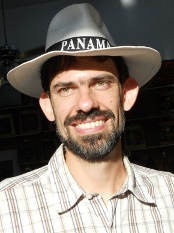} 
	\textbf{Daniel Ratton Figueiredo} received a BS cum laude degree in Computer Science and MSc degree in Computer and Systems Engineering from the Federal University of Rio de Janeiro (UFRJ) Brazil, in 1996 and 1999, respectively. He received a MSc and PhD degrees in Computer Science from the University of Massachusetts Amherst (UMass) in 2005. He worked as a post-doc researcher at the Swiss Federal Institute of Technology, Lausanne (EPFL). In 2007, he joined the Department of Computer and Systems Engineering (PESC/COPPE) at the Federal University of Rio de Janeiro (UFRJ), Brazil where he currently holds an Associate Professor position. He holds a Research Productivity Fellowship granted by CNPq (since 2009) and was member of the Young Scientists Program granted by FAPERJ (from 2010 to 2018). He was a Visiting Fulbright Scholar at Columbia University (6 months in 2017). His current main interests are in Network Science and Graph Learning.
	\endbio
}

\clearpage

\appendix
\onecolumn

\section{Train and test set comparison} \label{apd:train}
The classification results portrayed in section \ref{sec:result} show only the test set accuracy obtained during training. In this section, the train set accuracy plots are added for comparison. The overall similarity between train and test set results attests to the lack of overfitting of the trained models. 

\begin{figure*}[h]
	\begin{center}
		\subfloat[MNIST, Test set]{ \label{fig:MNISTacc_test}
			\includegraphics[width=.315\linewidth]{{acc-Te_mnist_base_H500_CD-10_l0.1_b50_i10-25rep-2}.pdf}
		}
		\subfloat[Mushrooms, Test set]{ \label{fig:MUSHacc_test}
			\includegraphics[width=.315\linewidth]{{acc-Te_mushrooms_sing_H100_CD-10_l0.01_b10_i10-25rep-2}.pdf}
		}
		\subfloat[Connect-4, Test Set]{ \label{fig:CON4acc_test}
			\includegraphics[width=.315\linewidth]{{acc-Te_connect-4_sing_H100_CD-10_l0.01_b10_i10-25rep-2}.pdf}
		} \\
		\subfloat[MNIST, Train set]{ \label{fig:MNISTacc_train}
			\includegraphics[width=.315\linewidth]{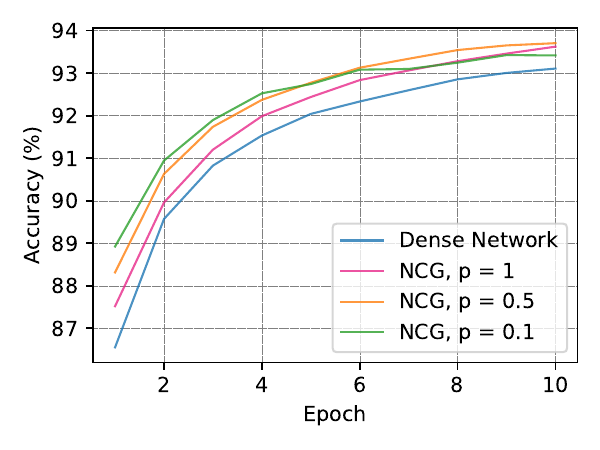}
		}
		\subfloat[Mushrooms, Train set]{ \label{fig:MUSHacc_train}
			\includegraphics[width=.315\linewidth]{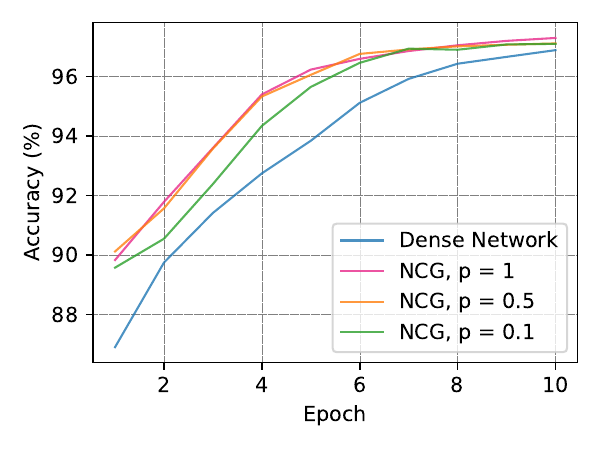}
		}
		\subfloat[Connect-4, Train Set]{ \label{fig:CON4acc_train}
			\includegraphics[width=.315\linewidth]{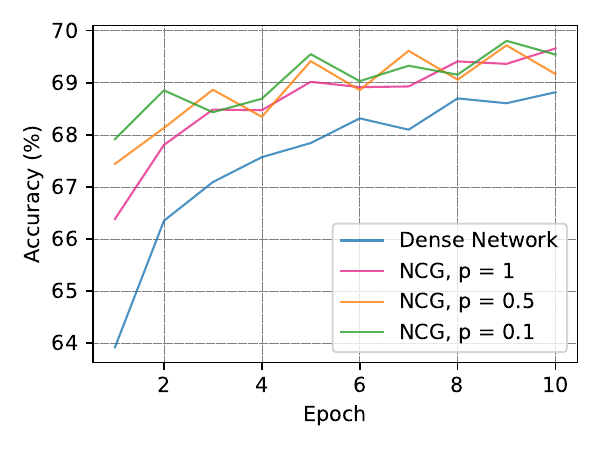}
		} 
		\caption{
			Test and train set classification accuracy comparison for NCG and the classical (dense) RBM for the MNIST, Mushrooms and Connect-4 data sets; lines are the sample means over 25 runs. (Test set images are equivalent to the ones in figure \ref{fig:class_main}.)
		}
		\label{fig:class_main_train}
	\end{center}
\end{figure*}

Figure \ref{fig:class_main_train} shows the evolution of the classification accuracy over the epochs for different NCG initializations, for both the test and train sets. This is the same as figure \ref{fig:class_main}, now adding the train set information. 
Note that the performance between training and test sets are qualitative and quantitatively similar, indicating there is likely no overfitting at this point. 

\begin{figure*}[!h]
	\vskip -3mm
	\begin{center}
		\subfloat[MNIST, Test set]{ \label{fig:class_mesmaTaxa_MNIST_test}
			\includegraphics[trim={0 4mm 0 3.5mm}, clip, width=.315\linewidth]{{acc-Te_mnist_H500_CD-10_l0.1_b50_i10-25r}.pdf}
		}
		\subfloat[Mushrooms, Test set]{ \label{fig:class_mesmaTaxa_mushrooms_test}
			\includegraphics[trim={0 4mm 0 3.5mm}, clip, width=.315\linewidth]{{acc-Te_mushrooms_msmTaxa_H100_CD-10_l0.01_b10_i10-25r}.pdf}
		}
		\subfloat[Connect-4, Test set]{ \label{fig:class_mesmaTaxa_connect-4_test}
			\includegraphics[trim={0 4mm 0 3.5mm}, clip, width=.315\linewidth]{{acc-Te_connect-4_msmTaxa_H100_CD-10_l0.01_b10_i10-25r}.pdf}
		} \\
		\subfloat[MNIST, Train set]{ \label{fig:class_mesmaTaxa_MNIST-train}
			\includegraphics[trim={0 4mm 0 3.5mm}, clip, width=.315\linewidth]{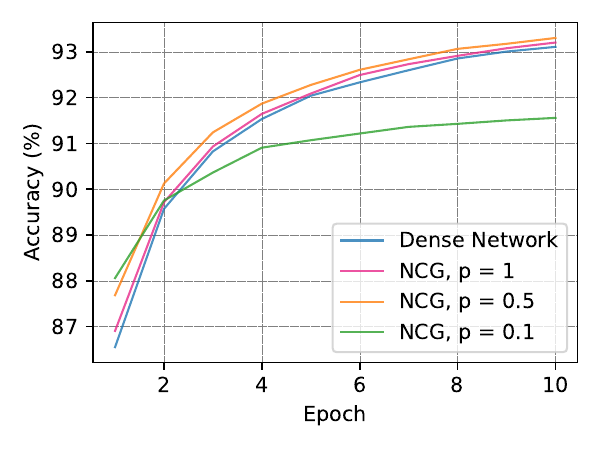}
		} 
		\subfloat[Mushrooms, Train set]{ \label{fig:class_mesmaTaxa_mushrooms-train}
			\includegraphics[trim={0 4mm 0 3.5mm}, clip, width=.315\linewidth]{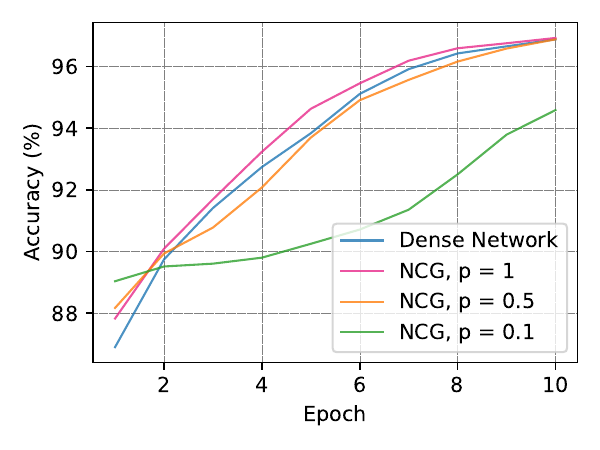}
		} 
		\subfloat[Connect-4, Train set]{ \label{fig:class_mesmaTaxa_connect-4-train}
			\includegraphics[trim={0 4mm 0 3.5mm}, clip, width=.315\linewidth]{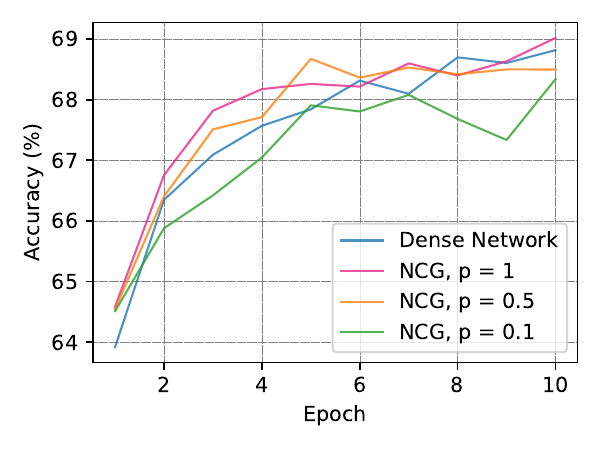}
		}
		\caption{
			Classification accuracy over the training epochs of NCG for the train and test sets of the MNIST, Mushrooms and Connect-4 data sets with $\alpha_A = \alpha$ ($ 0.1 $ for MNIST and $ 0.01 $ for the other data sets); lines correspond to the sample mean over 25 experimental runs. 
			(Test set images are equivalent to the ones in figure \ref{fig:class_mesmaTaxa}.)
		}
		\label{fig:class_mesmaTaxa_train}
	\end{center}
%
	\begin{center}
		\subfloat[MNIST, Test set]{ \label{fig:class_gamma25_MNIST-test}
			\includegraphics[trim={0 4mm 0 3.5mm}, clip, width=.315\linewidth]{{acc-Te_mnist_g25_H500_CD-10_l0.1_b50_i10-25r}.pdf}
		}
		\subfloat[Mushrooms, Test set]{ \label{fig:class_gamma25_mushrooms-test}
			\includegraphics[trim={0 4mm 0 3.5mm}, clip, width=.315\linewidth]{{acc-Te_mushrooms_g25_H100_CD-10_l0.01_b10_i10-25r}.pdf}
		}
		\subfloat[Connect-4, Test set]{ \label{fig:class_gamma25_connect-4-test}
			\includegraphics[trim={0 4mm 0 3.5mm}, clip, width=.315\linewidth]{{acc-Te_connect-4_g25_H100_CD-10_l0.01_b10_i10-25r}.pdf}
		} \\
		\subfloat[MNIST, Train set]{ \label{fig:class_gamma25_MNIST-train}
			\includegraphics[trim={0 4mm 0 3.5mm}, clip, width=.315\linewidth]{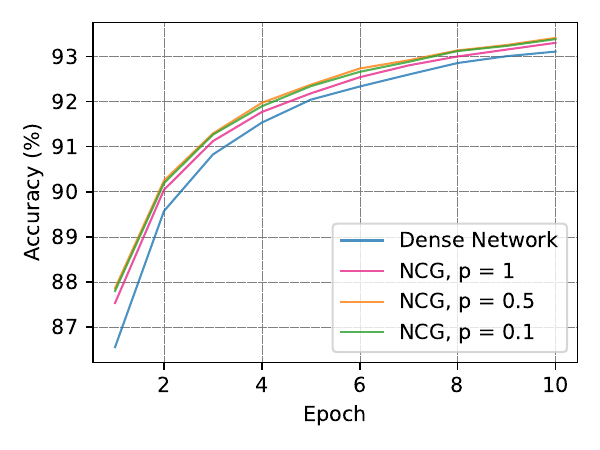}
		}
		\subfloat[Mushrooms, Train set]{ \label{fig:class_gamma25_mushrooms-train}
			\includegraphics[trim={0 4mm 0 3.5mm}, clip, width=.315\linewidth]{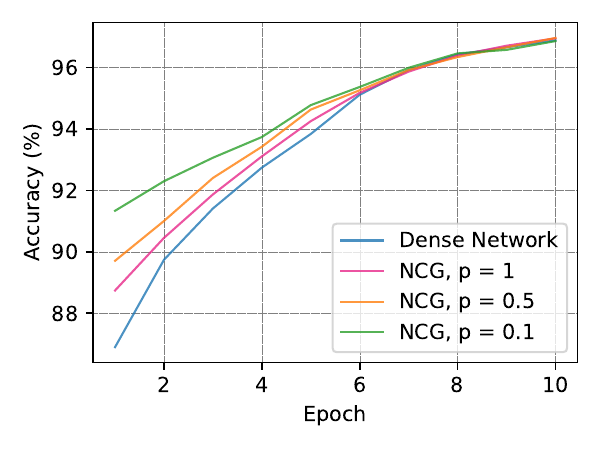}
		}
		\subfloat[Connect-4, Train set]{ \label{fig:class_gamma25_connect-4-train}
			\includegraphics[trim={0 4mm 0 3.5mm}, clip, width=.315\linewidth]{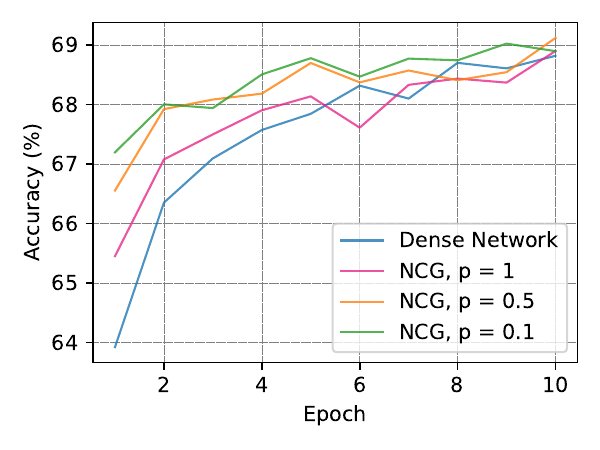}
		}
		\caption{
			Classification accuracy over the training epochs of NCG for the test and train sets of the MNIST, Mushrooms and Connect-4 data sets with $\gamma = 0.25$; lines correspond to the sample mean over 25 experimental runs. 
			(Test set images are equivalent to the ones in figure \ref{fig:class_gamma25}.)
		}
		\label{fig:class_gamma25_train}
	\end{center}
	\vskip -7mm
\end{figure*}

Figure \ref{fig:class_mesmaTaxa_train} corresponds to the evolution of the classification accuracy over the epochs for different NCG initializations considering that training is achieved using a smaller connectivity learning rate: $\alpha_A = \alpha$. For the MNIST data set, $\alpha = 0.1$ and for the Mushrooms and Connect-4 data sets $\alpha = 0.01$. This is equivalent to figure \ref{fig:class_mesmaTaxa}, but now adding the train set learning curves. 

Figure \ref{fig:class_gamma25_train} has the evolution of the classification accuracy over the epochs for different NCG initializations, now using a smaller connectivity threshold of $\gamma = 0.25$. Note that this makes disabling connections harder, inducing denser networks. Therefore, all curves become more similar to the dense network (blue line). It is equivalent to figure \ref{fig:class_gamma25}, with the added train set learning curves. 

\begin{figure*}[!h]
	\vskip -4mm
	\begin{center}
		\subfloat[MNIST, Test set]{ \label{fig:class_gamma75_MNIST-test}
			\includegraphics[trim={0 4mm 0 3.5mm}, clip, width=.315\linewidth]{{acc-Te_mnist_g75_H500_CD-10_l0.1_b50_i10-25r}.pdf}
		}
		\subfloat[Mushrooms, Test set]{ \label{fig:class_gamma75_mushrooms-test}
			\includegraphics[trim={0 4mm 0 3.5mm}, clip, width=.315\linewidth]{{acc-Te_mushrooms_g75_H100_CD-10_l0.01_b10_i10-25r}.pdf}
		}
		\subfloat[Connect-4, Test set]{ \label{fig:class_gamma75_connect-4-test}
			\includegraphics[trim={0 4mm 0 3.5mm}, clip, width=.315\linewidth]{{acc-Te_connect-4_g75_H100_CD-10_l0.01_b10_i10-25r}.pdf}
		} \\
		\subfloat[MNIST, Train set]{ \label{fig:class_gamma75_MNIST-train}
			\includegraphics[trim={0 4mm 0 3.5mm}, clip, width=.315\linewidth]{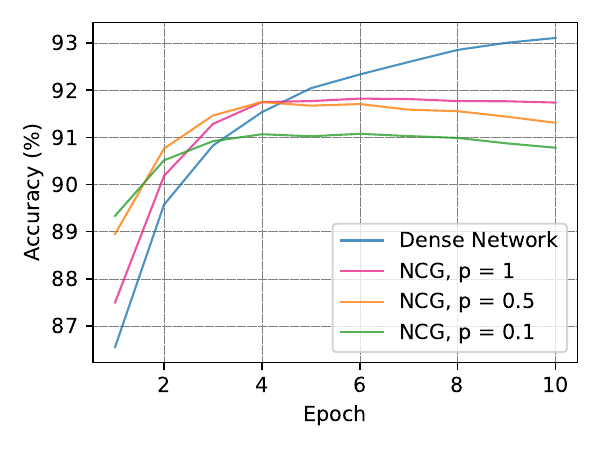}
		}
		\subfloat[Mushrooms, Train set]{ \label{fig:class_gamma75_mushrooms-train}
			\includegraphics[trim={0 4mm 0 3.5mm}, clip, width=.315\linewidth]{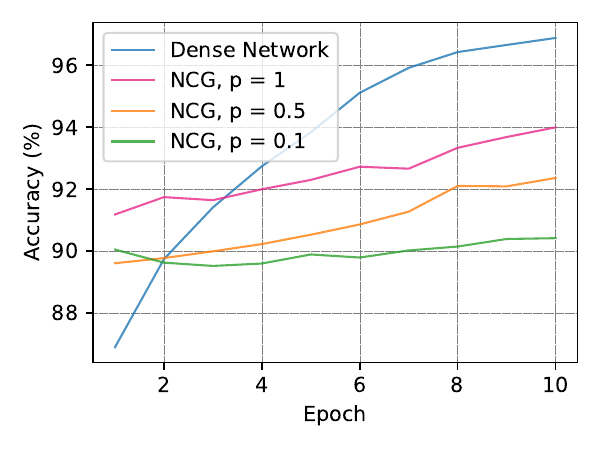}
		}
		\subfloat[Connect-4, Train set]{ \label{fig:class_gamma75_connect-4-train}
			\includegraphics[trim={0 4mm 0 3.5mm}, clip, width=.315\linewidth]{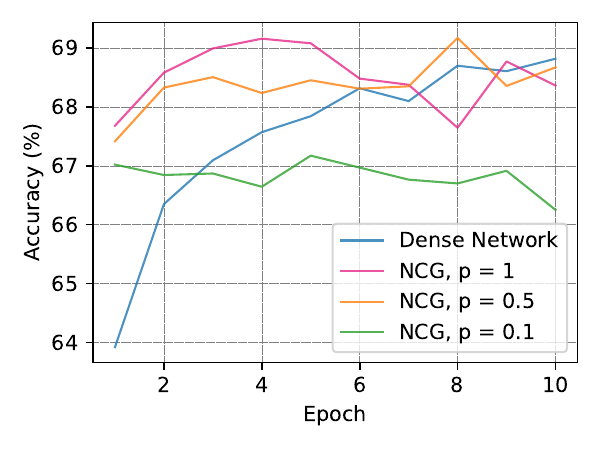}
		}
		\caption{
			Classification accuracy over the training epochs of NCG for the test and train sets of the MNIST, Mushrooms and Connect-4 data sets with $\gamma = 0.75$; lines correspond to the sample mean over 25 experimental runs. 
			(Test set images are equivalent to the ones in figure \ref{fig:class_gamma75}.)
		}
		\label{fig:class_gamma75_train}
%
		\subfloat[MNIST, Test set]{ \label{fig:compMNISTacc}
			\includegraphics[trim={2mm 3mm 2mm 5mm}, clip, width=.31\linewidth]{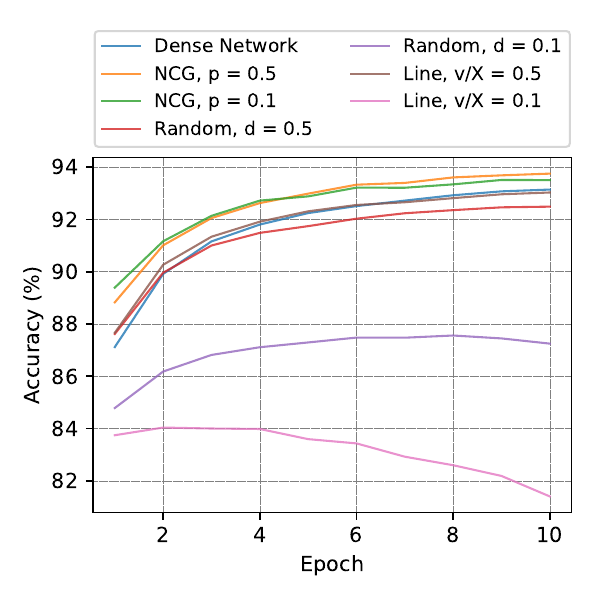}
		}
		\subfloat[Mushrooms, Test set]{ \label{fig:compMUSHacc}
			\includegraphics[trim={2mm 3mm 2mm 5mm}, clip, width=.31\linewidth]{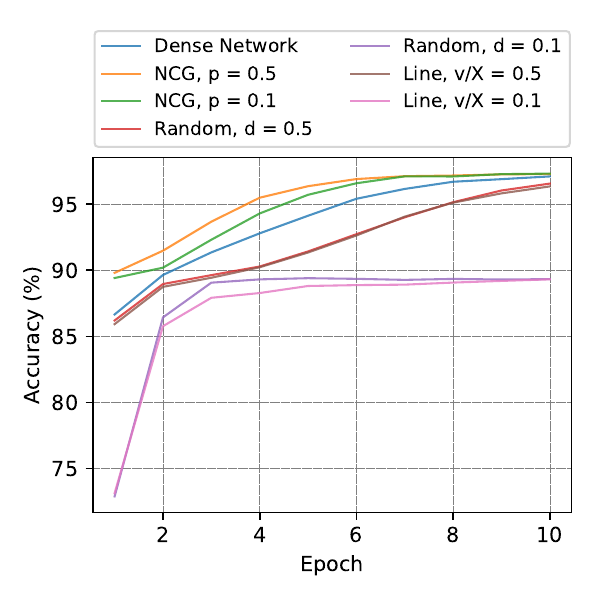}
		}
		\subfloat[Connect-4, Test Set]{ \label{fig:compCON4acc}
			\includegraphics[trim={2mm 3mm 2mm 5mm}, clip, width=.31\linewidth]{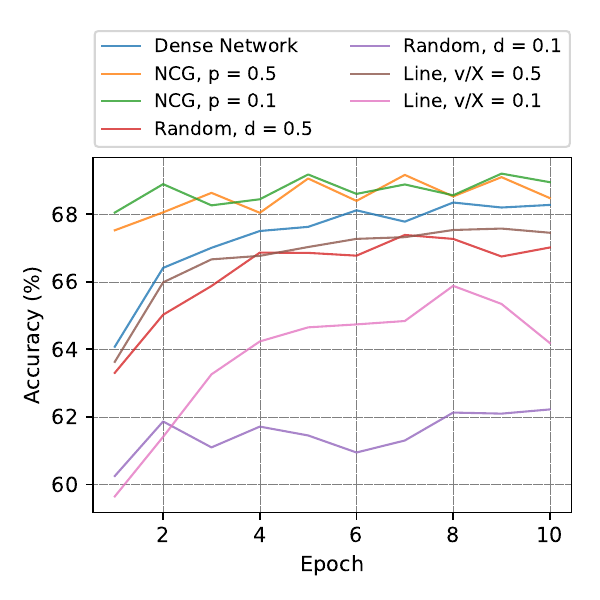}
		} \\
		\subfloat[MNIST, Train set]{ \label{fig:compMNISTacc_train}
			\includegraphics[trim={2mm 3mm 2mm 5mm}, clip, width=.31\linewidth]{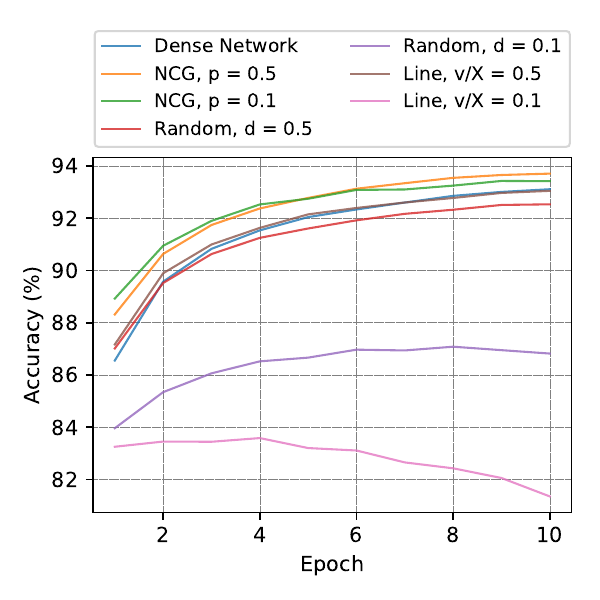}
		}
		\subfloat[Mushrooms, Train Set]{ \label{fig:compMUSHacc_train}
			\includegraphics[trim={2mm 3mm 2mm 5mm}, clip, width=.31\linewidth]{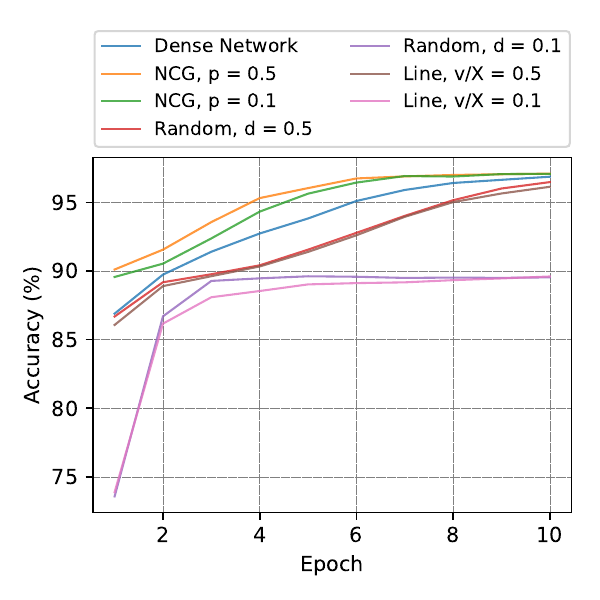}
		}
		\subfloat[Connect-4, Train Set]{ \label{fig:compCON4acc_train}
			\includegraphics[trim={2mm 3mm 2mm 5mm}, clip, width=.31\linewidth]{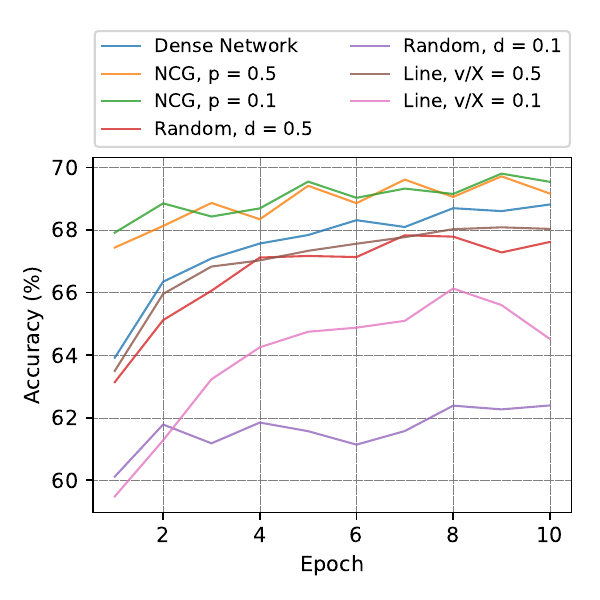}
		}
		\caption{
			Test and train sets classification accuracy for NCG, the classical RBM, and the Line and Random patterns for the MNIST, Mushrooms and Connect-4 data sets; lines are the sample means over 25 runs. 
		}
		\label{fig:class_compStatPats}
	\end{center}
	\vskip -7mm
\end{figure*}

Figure \ref{fig:class_gamma75_train} has the evolution of the classification accuracy over the epochs for different NCG initializations, now using a larger connectivity threshold of $\gamma = 0.75$. Note that this induces sparsity. It is equivalent to figure \ref{fig:class_gamma75}, with the added train set learning curves. 

\section{Connectivity Patterns Comparison} \label{apd:conn_patterns}
Besides the comparison with the dense network (the traditional RBM connectivity), the NCG method was compared with two other other fixed connectivity patterns: the line and the random pattern \citep{selfAnonymous}. As per described in the main body of the paper, the line pattern corresponds to a network with each hidden unit connected to $v$ consecutive visible units. That means that the percentage of activated connections is given by $v/X$, which is displayed in the following plots for easiness of comparison between the connectivity networks. The random pattern corresponds to a bipartite Erdös-Rényi random graph. for this pattern, the connectivity network is created by activating each connection with a given probability $p$. This also means that, on average, a percentage of $p$ connections should be active for any generated graph. 

Figure \ref{fig:class_compStatPats} shows the comparison of the accuracy learning curves between NCG, the dense network, the line and the random pattern. For fairness of comparison, networks with the same expected sparsity were considered (for NCG, the initialized connectivity had the same sparsity). 
Note that other connectivity patterns showed markedly worse performance, even when compared to the dense network, illustrating that connectivity pattern is more important than network sparsity.%

\section{Classification Degree Statistics} \label{apd:degreeAcc}
\begin{figure*}[h]
	\begin{center}
		\subfloat[$ p = 1 $]{\includegraphics[trim={2mm 3mm 2mm 3mm}, clip, width=.31\linewidth]{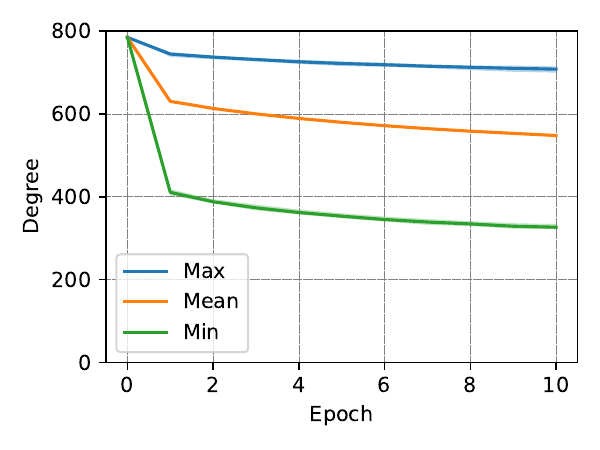}} 
		\subfloat[$ p = 0.5 $]{\includegraphics[trim={2mm 3mm 2mm 3mm}, clip, width=.31\linewidth]{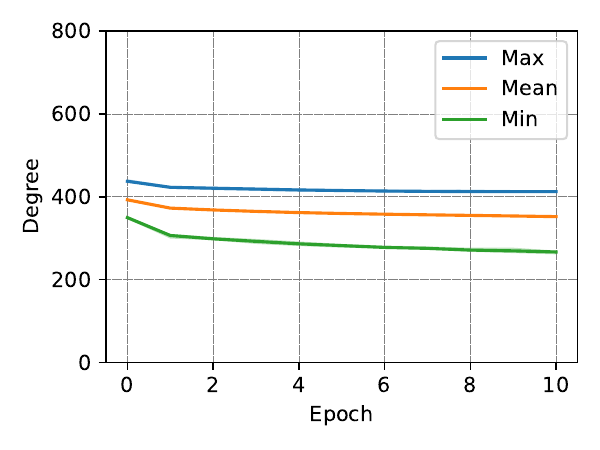}} 
		\subfloat[$ p = 0.1 $]{\includegraphics[trim={2mm 3mm 2mm 3mm}, clip, width=.31\linewidth]{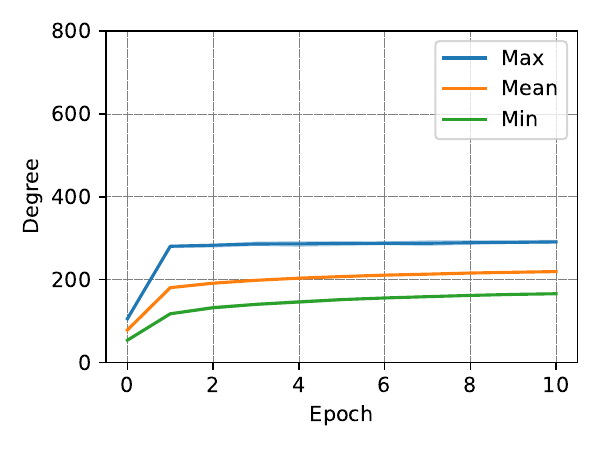}}
		\caption{
			Degree statistics (minimum, average, maximum) of the hidden units over the training epochs for the MNIST data set -- lines correspond to the sample mean and shades to the sample quartiles over 25 experimental runs. 
		}
		\label{fig:classMnist_conn}
%
	
	\subfloat[Mushrooms, $ p = 0.5 $]{ \label{fig:classMushrooms_conn-p05}
		\includegraphics[trim={2mm 3mm 2mm 3mm}, clip, width=.23\linewidth]{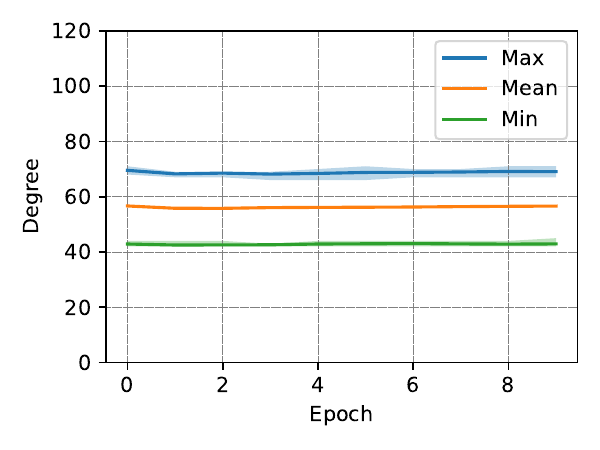}
	} 
	\subfloat[Mushrooms, $ p = 0.1 $]{ \label{fig:classMushrooms_conn-p01}
		\includegraphics[trim={2mm 3mm 2mm 3mm}, clip, width=.23\linewidth]{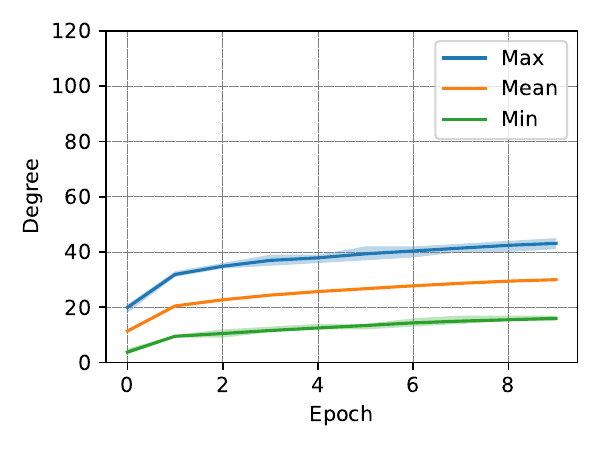}
	} 
	\subfloat[Connect-4, $ p = 0.5 $]{ \label{fig:classConnect-4_conn-p05}
		\includegraphics[trim={2mm 3mm 2mm 3mm}, clip, width=.23\linewidth]{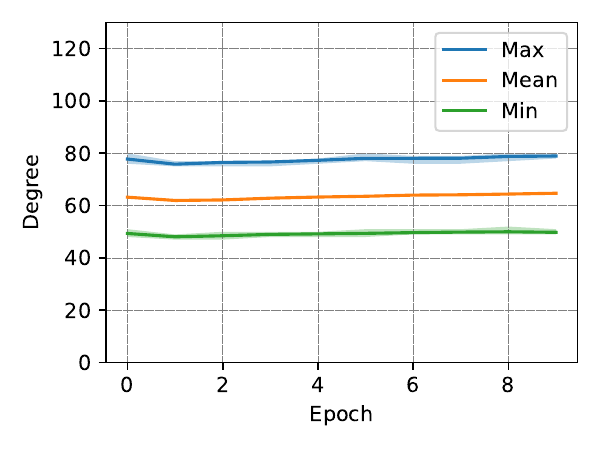}
	} 
	\subfloat[Connect-4, $ p = 0.1 $]{ \label{fig:classConnect-4_conn-p01}
		\includegraphics[trim={2mm 3mm 2mm 3mm}, clip, width=.23\linewidth]{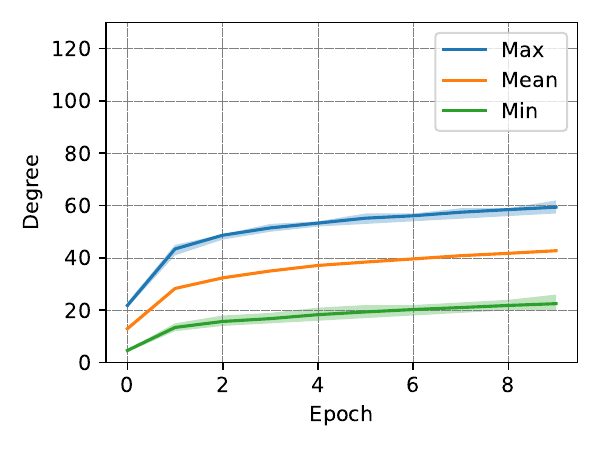}
	} 
	\caption{
		Degree statistics (minimum, average, maximum) of the hidden units over the training epochs for the Mushrooms (a and b) and Connect-4 (c and d) data sets -- lines correspond to the sample mean and shades to the sample quartiles over 25 experimental runs. 
	}
	\label{fig:classUCISuite_conn}
		\subfloat[$ p = 1 $]{\includegraphics[trim={2mm 3mm 2mm 3mm}, clip, width=.31\linewidth]{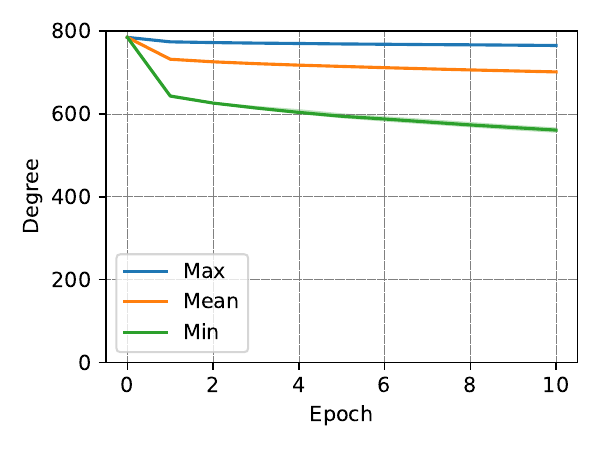}} 
		\subfloat[$ p = 0.5 $]{\includegraphics[trim={2mm 3mm 2mm 3mm}, clip, width=.31\linewidth]{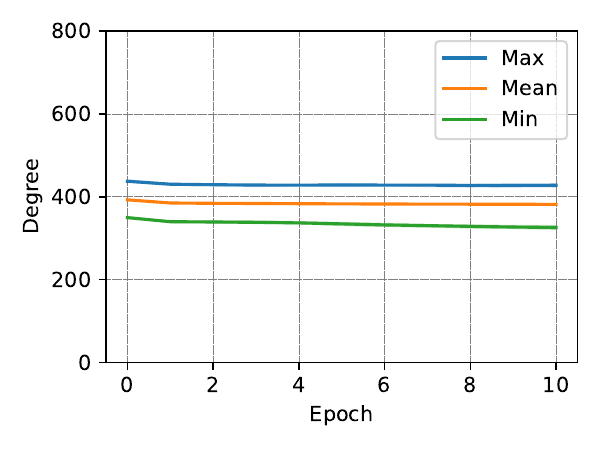}} 
		\subfloat[$ p = 0.1 $]{\includegraphics[trim={2mm 3mm 2mm 3mm}, clip, width=.31\linewidth]{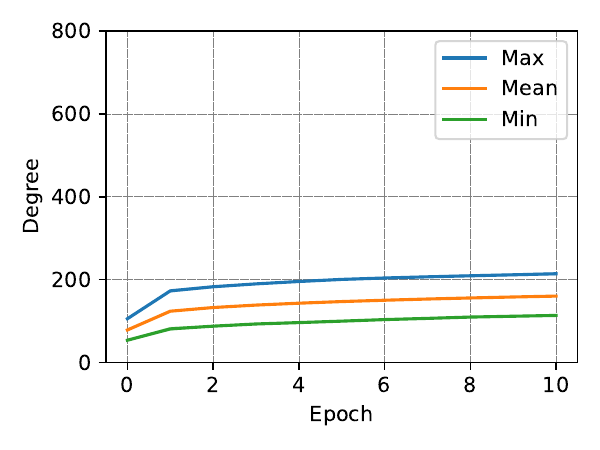}}
		\caption{
			Degree statistics (minimum, average, maximum) of the hidden units over the training epochs for the MNIST data set trained with $ \alpha_A = \alpha = 0.1 $ -- lines correspond to the sample mean and shades to the sample quartiles over 25 experimental runs. 
		}
		\label{fig:classMnist_conn_mesmaTaxa}
	\end{center}
\end{figure*}

Figure \ref{fig:classMnist_conn} portrays the degree statistics (minimum, average, and maximum) of the network's hidden units over the epochs for the NCG classification experiments in the MNIST data set (figures \ref{fig:MNISTacc} and \ref{fig:MNISTacc_train} show the accuracy). Note that for $p=1$ all degrees are 784 at time zero, and NCG significantly reduces the degrees of the network; the average degree is reduced by 30\% after 10 epochs. On the other hand, for $p=0.1$, NCG significantly increases the degrees of the network; the average degree is 2.5 times larger after 10 epochs. Finally, for $p=0.5$ NCG shows a relatively small change in the degrees. Moreover, while the degrees change and converge over the epochs, the initialization density has a strong influence: the average degree of the three models after 10 epochs reflects their initial density. 
Furthermore, figure \ref{fig:classUCISuite_conn} shows the degree statistics of the networks trained for the Mushrooms (figures \ref{fig:classMushrooms_conn-p05} and \ref{fig:classMushrooms_conn-p01}) and the Connect-4 (figures \ref{fig:classConnect-4_conn-p05} and \ref{fig:classConnect-4_conn-p01}) data sets. These experiments pertain to the accuracy performances portrayed in figure \ref{fig:class_compStatPats}. Once again, one can see that the final sparsity reflects the initial sparsity, and the networks end training with very different connectivity patterns after 10 epochs, despite the similar accuracy performances. 

%
\begin{figure*}[h!]
	\begin{center}
		\subfloat[$ p = 1 $]{\includegraphics[trim={2mm 3mm 2mm 3mm}, clip, width=.31\linewidth]{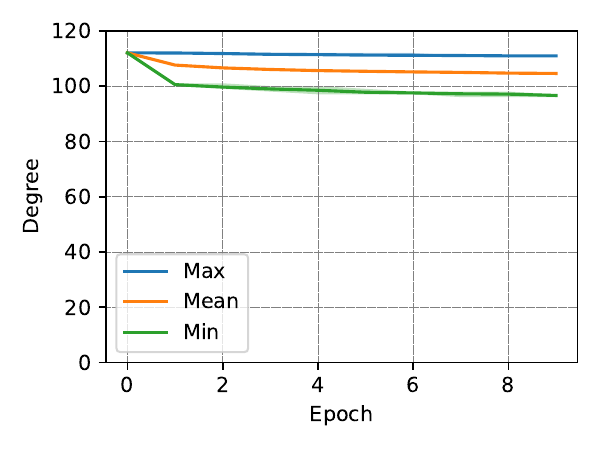}} 
		\subfloat[$ p = 0.5 $]{\includegraphics[trim={2mm 3mm 2mm 3mm}, clip, width=.31\linewidth]{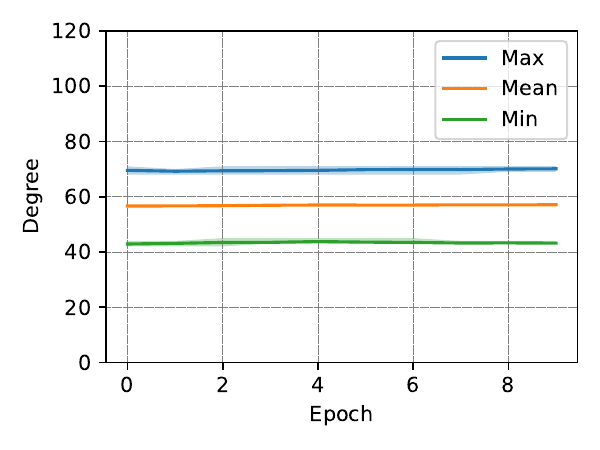}} 
		\subfloat[$ p = 0.1 $]{\includegraphics[trim={2mm 3mm 2mm 3mm}, clip, width=.31\linewidth]{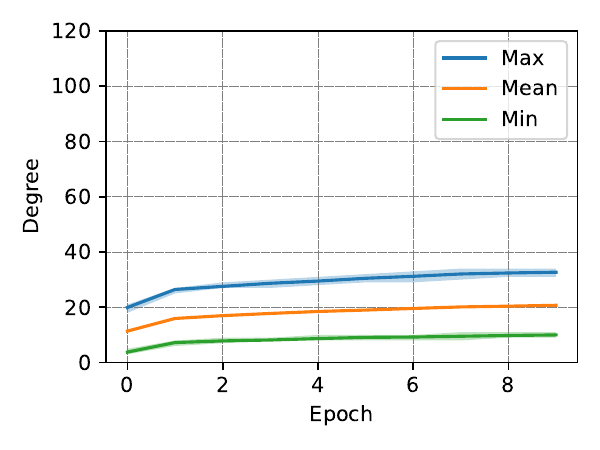}}
		\caption{
			Degree statistics (minimum, average, maximum) of the hidden units over the training epochs for the Mushrooms data set trained with $ \alpha_A = \alpha = 0.01 $ -- lines correspond to the sample mean and shades to the sample quartiles over 25 experimental runs. 
		}
		\label{fig:classMushrooms_conn_mesmaTaxa}
	\end{center}
	\begin{center}
		\subfloat[$ p = 1 $]{\includegraphics[trim={2mm 3mm 2mm 3mm}, clip, width=.31\linewidth]{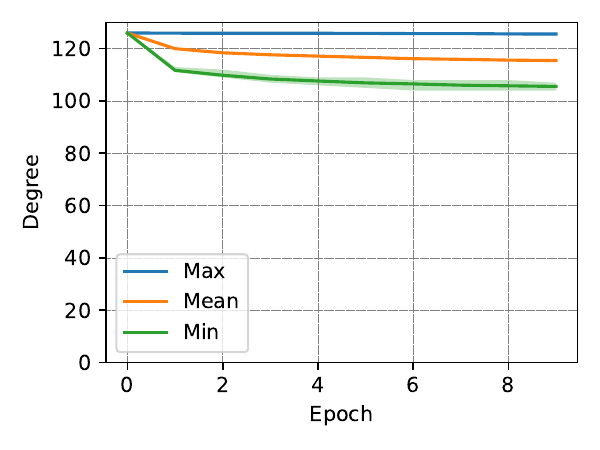}} 
		\subfloat[$ p = 0.5 $]{\includegraphics[trim={2mm 3mm 2mm 3mm}, clip, width=.31\linewidth]{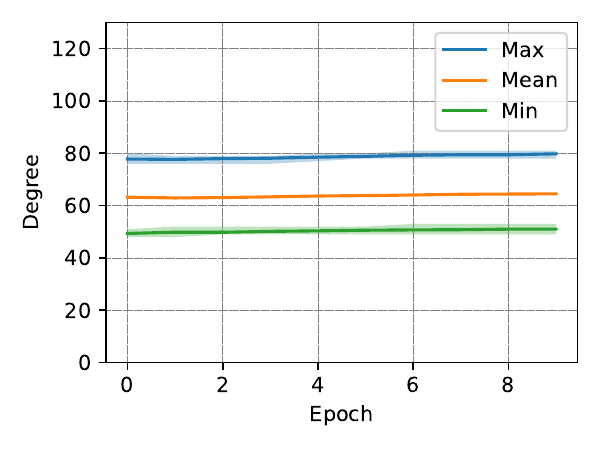}} 
		\subfloat[$ p = 0.1 $]{\includegraphics[trim={2mm 3mm 2mm 3mm}, clip, width=.31\linewidth]{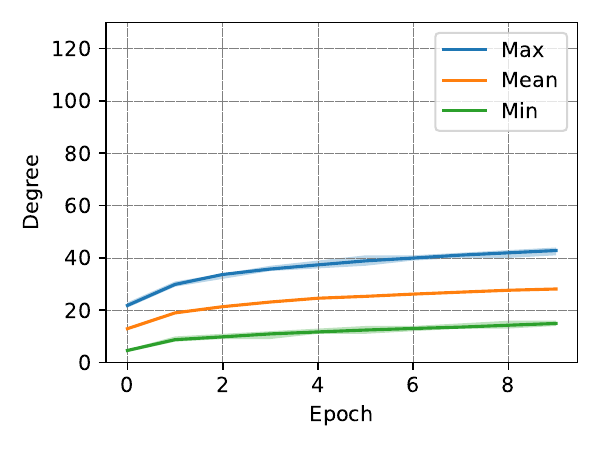}}
		\caption{
			Degree statistics (minimum, average, maximum) of the hidden units over the training epochs for the Mushrooms data set trained with $ \alpha_A = \alpha = 0.01 $ -- lines correspond to the sample mean and shades to the sample quartiles over 25 experimental runs. 
		}
		\label{fig:classConnect-4_conn_mesmaTaxa}
	\end{center}
	\begin{center}
		\subfloat[]{ 
			\label{fig:static_mnist_nll_quartile}
			\includegraphics[width=.48\columnwidth]{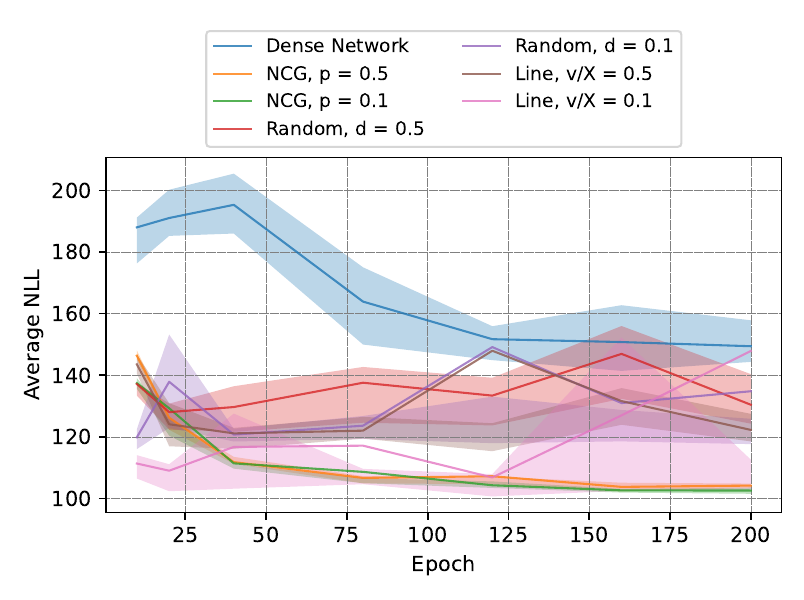}
		}
		\subfloat[]{ 
			\label{fig:setComp_qe}
			\includegraphics[width=.48\columnwidth]{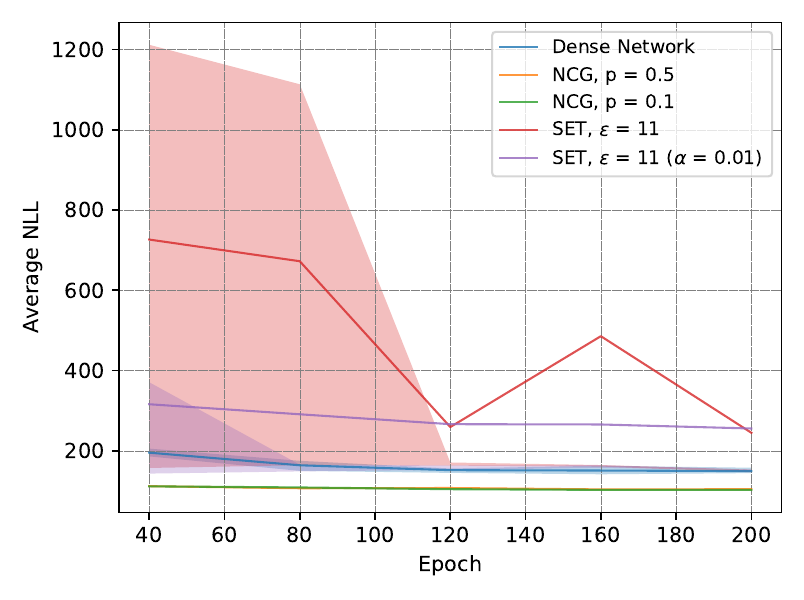}
		}
		\caption{
			Average NLL over the training epochs -- lines are the mean values and shades are the quatiles over 10 experiments. 
			(a) shows the comparison of NCG with the Line and Random patterns (regarding figure \ref{fig:static_mnist_nll}). 
			(b) shows the comparison of NCG with the SET method~\cite{SET_mocanu} (regarding figure \ref{fig:setComp}). 
		}
		\label{fig:nll_ncg_comparisons}
	\end{center}
\end{figure*}

Lastly, figures \ref{fig:classMnist_conn_mesmaTaxa}, \ref{fig:classMushrooms_conn_mesmaTaxa} and \ref{fig:classConnect-4_conn_mesmaTaxa} show the degree statistics for training experiments with a lower connectivity learning rate. The plots correspond, respectively, to experiments performed with the MNIST, Mushrooms and Connect-4 data sets, and pertain to the accuracy results presented in figure \ref{fig:class_mesmaTaxa}. As expected, one can see much less change in the connectivity with regards to the previous experiments, which use a higher learning rate. This is especially notable for the MNIST $ p=1 $ experiments.

\section{Quartile Figures} \label{apd:quartile}
Some plots in the main body had their uncertainties removed to avoid clutter. The plots with quartile representing the uncertainty are shown here. Figure \ref{fig:static_mnist_nll_quartile} shows the learning curves of NCG and the classic RBM compared to the line and random patterns, as in Figure \ref{fig:static_mnist_nll}, and figure \ref{fig:setComp_qe} shows the comparison with the SET method, as in figure \ref{fig:setComp}, for the generative task. 

\begin{figure*}[h]
	\begin{center}
		\def\pltsize{.315\textwidth}
		\subfloat[MNIST, Test set]{ \label{fig:compMNISTacc_qe-test}
			\includegraphics[width=\pltsize]{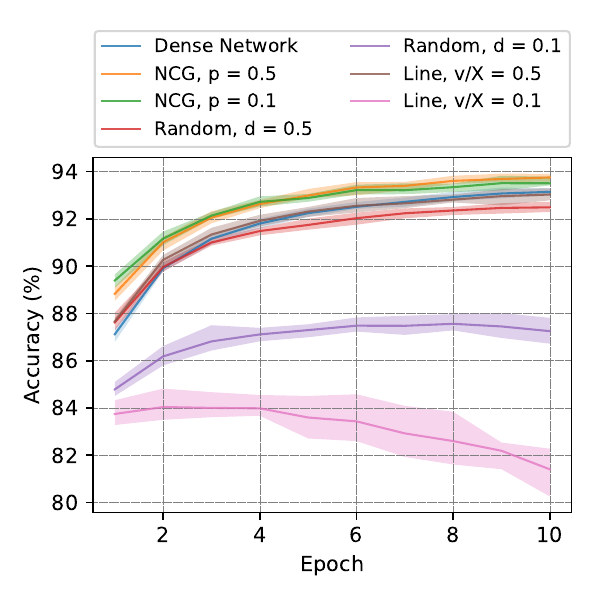}
		} 
		\subfloat[Mushrooms, Test set]{ \label{fig:compMUSHacc_qe-test}
			\includegraphics[width=\pltsize]{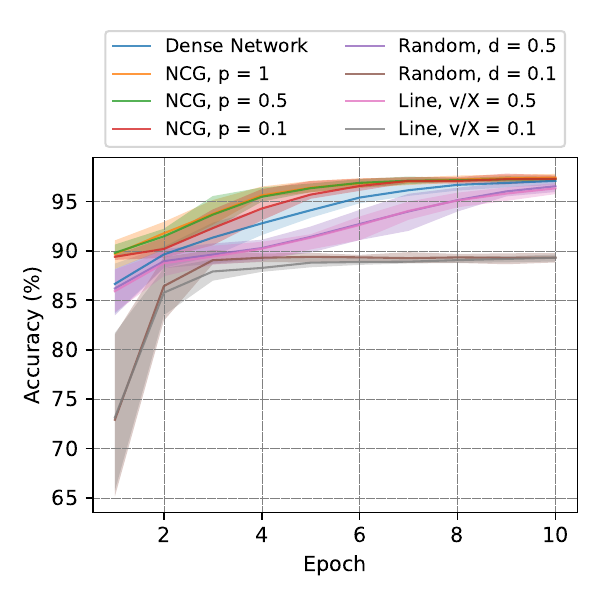}
		} 
		\subfloat[Connect-4, Test set]{ \label{fig:compCON4acc_qe-test}
		\includegraphics[width=\pltsize]{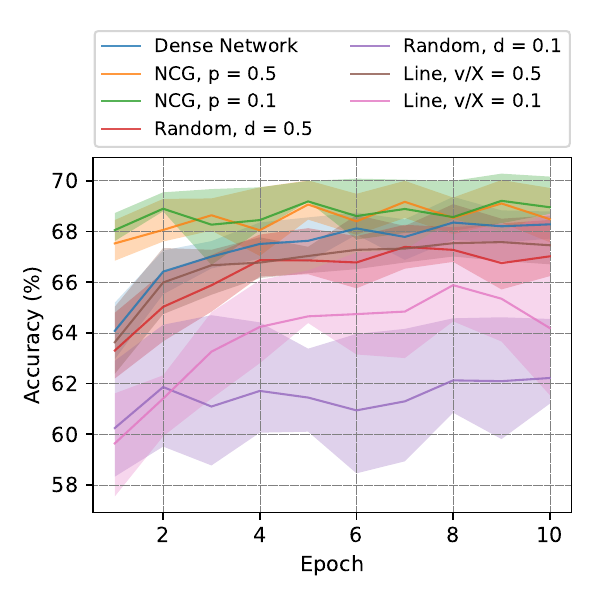}
		} \\
		\subfloat[MNIST, Train set]{ \label{fig:compMNISTacc_qe-train}
		\includegraphics[width=\pltsize]{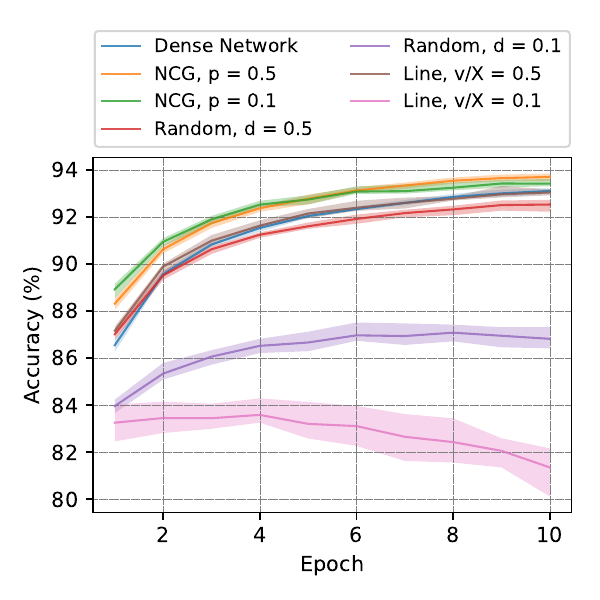}
		} 
		\subfloat[Mushrooms, Train set]{ \label{fig:compMUSHacc_qe-train}
			\includegraphics[width=\pltsize]{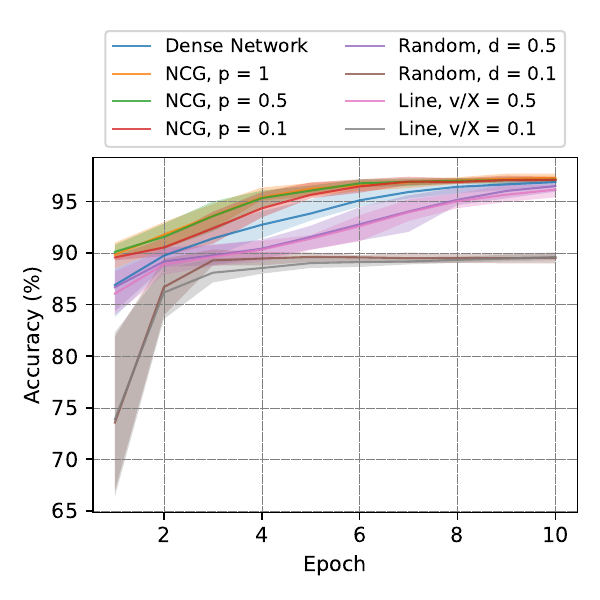}
		}
		\subfloat[Connect-4, Train set]{ \label{fig:compCON4acc_qe-train}
			\includegraphics[width=\pltsize]{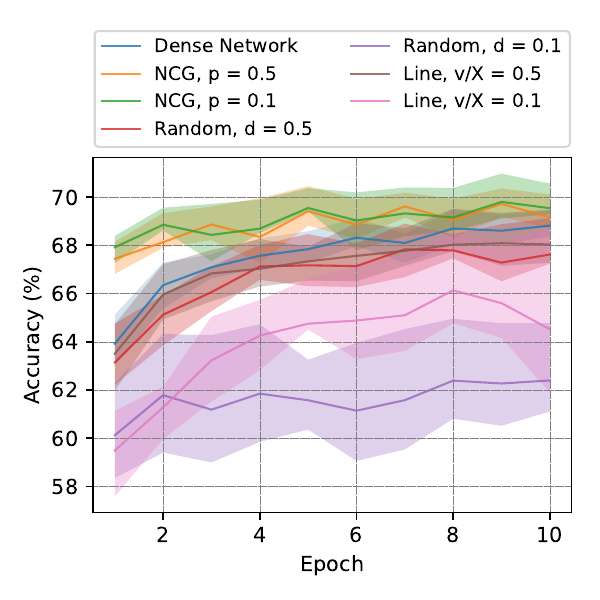}
		}
		\caption{
			(Regarding figure \ref{fig:class_compStatPats}. Contains the curves from figures \ref{fig:class_main} and \ref{fig:class_main_train} as well.) Classification accuracy for NCG, the classical RBM, and the Line and Random patterns for the test and train sets of MNIST, Mushrooms and Connect-4 data sets; lines are the sample means and shades are the quatiles over 25 runs. 
		}
		\label{fig:class_compStatPats_qe}
	\end{center}
\end{figure*}

Meanwhile, Figure \ref{fig:class_compStatPats_qe} presents the classification results for training of NCG compared with the fully connected RBM, the line and the random patters, for data sets MNIST, Mushrooms and Connect-4, as in Figure \ref{fig:class_compStatPats}. (Note that the quartiles of figure \ref{fig:class_main} are also represented in figure \ref{fig:class_compStatPats_qe}, since it contains all curves in the original figure as well as the other connectivity patterns learning curves.)

\begin{figure*}[!h]
	\vskip -3mm
	\begin{center}
		\def\pltsize{.315\textwidth}
		\subfloat[MNIST, Test set]{ \label{fig:compMNISTacc_mesmaTaxa_qe-test}
			\includegraphics[width=\pltsize]{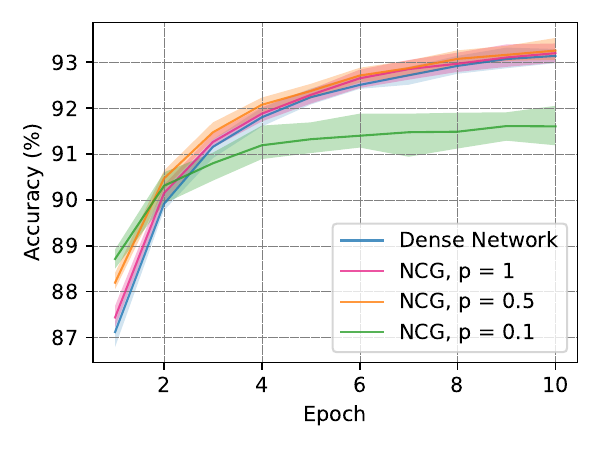}
		} 
		\subfloat[Mushrooms, Test set]{ \label{fig:compMUSHacc_mesmaTaxa_qe-test}
		\includegraphics[width=\pltsize]{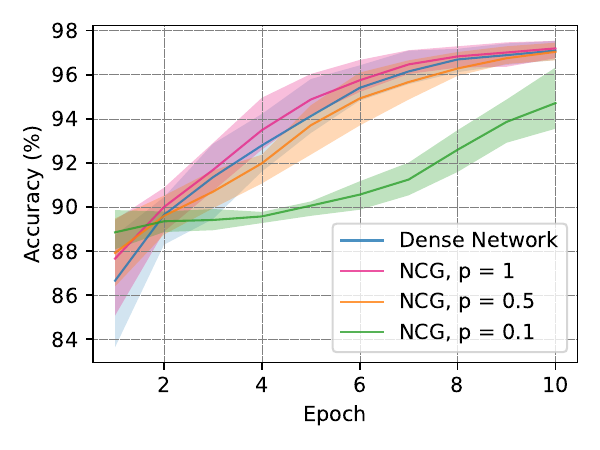}
		} 
		\subfloat[Connect-4, Test set]{ \label{fig:compCON4acc_mesmaTaxa_qe-test}
		\includegraphics[width=\pltsize]{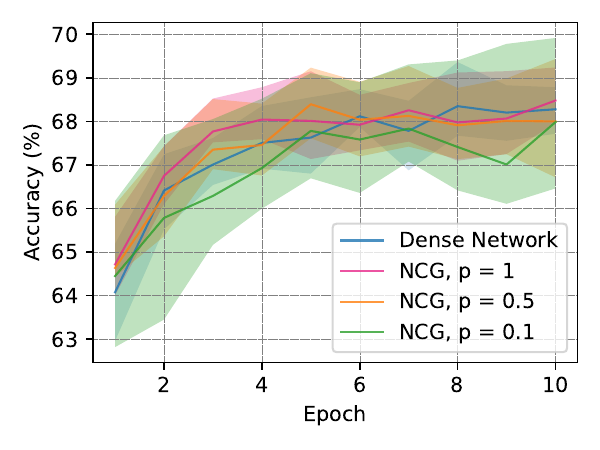}
		} \\
		\subfloat[MNIST, Train set]{ \label{fig:compMNISTacc_mesmaTaxa_qe-train}
			\includegraphics[width=\pltsize]{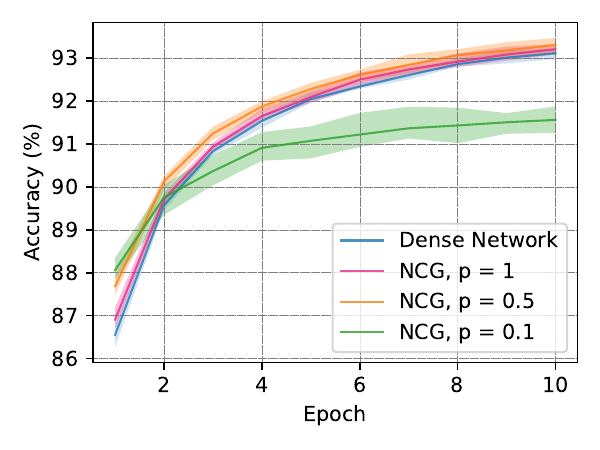}
		} 
		\subfloat[Mushrooms, Train set]{ \label{fig:compMUSHacc_mesmaTaxa_qe-train}
			\includegraphics[width=\pltsize]{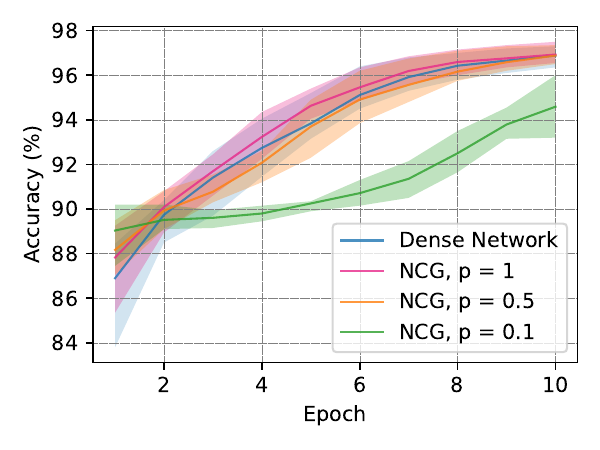}
		}
		\subfloat[Connect-4, Train set]{ \label{fig:compCON4acc_mesmaTaxa_qe-train}
			\includegraphics[width=\pltsize]{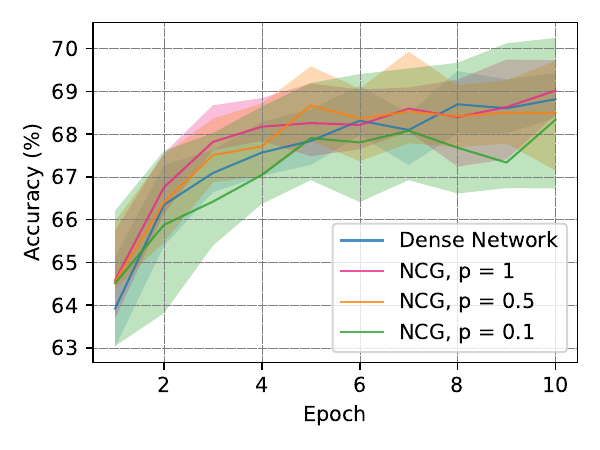}
		}
		\caption{
			(Regarding figures \ref{fig:class_mesmaTaxa} and \ref{fig:class_mesmaTaxa_train}.) Classification accuracy over the training epochs of NCG, for the test and training sets of MNIST, Mushrooms and Connect-4 data sets with $ \alpha_A = \alpha $; lines are the sample means and shades are the quatiles over 25 runs. 
		}
		\label{fig:class_mesmaTaxa_qe}
	\end{center}
%
	\begin{center}
		\subfloat[MNIST, Test set]{ \label{fig:class_gamma25_MNIST_qe_test}
			\includegraphics[trim={0 4mm 0 3.5mm}, clip, width=.315\linewidth]{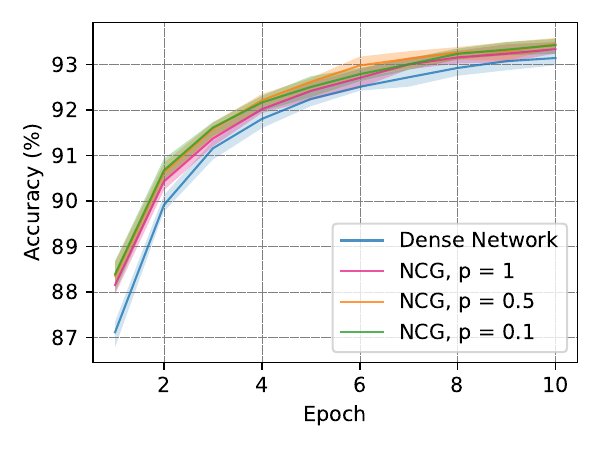}		}
		\subfloat[Mushrooms, Test set]{ \label{fig:class_gamma25_mushrooms_qe_test}
			\includegraphics[trim={0 4mm 0 3.5mm}, clip, width=.315\linewidth]{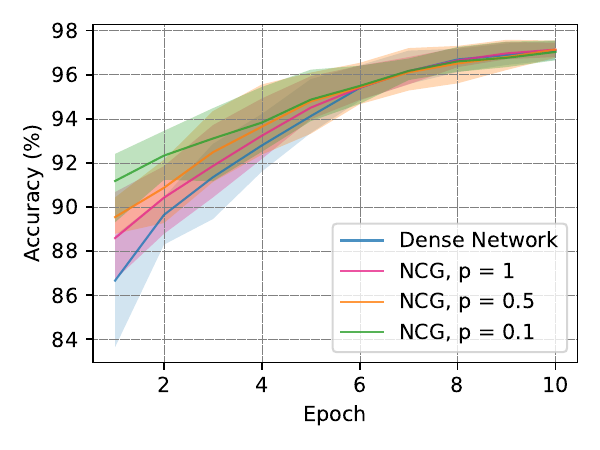}
		}
		\subfloat[Connect-4, Test set]{ \label{fig:class_gamma25_connect-4_qe_test}
			\includegraphics[trim={0 4mm 0 3.5mm}, clip, width=.315\linewidth]{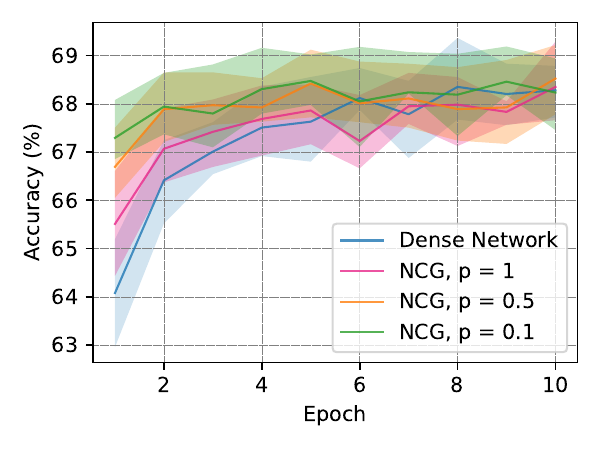}
		} \\
		\subfloat[MNIST, Train set]{ \label{fig:class_gamma25_MNIST_qe_train}
		\includegraphics[trim={0 4mm 0 3.5mm}, clip, width=.315\linewidth]{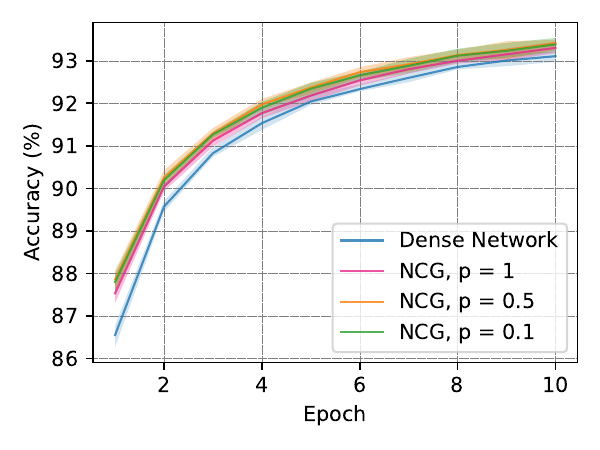}
		}
		\subfloat[Mushrooms, Train set]{ \label{fig:class_gamma25_mushrooms_qe_train}
		\includegraphics[trim={0 4mm 0 3.5mm}, clip, width=.315\linewidth]{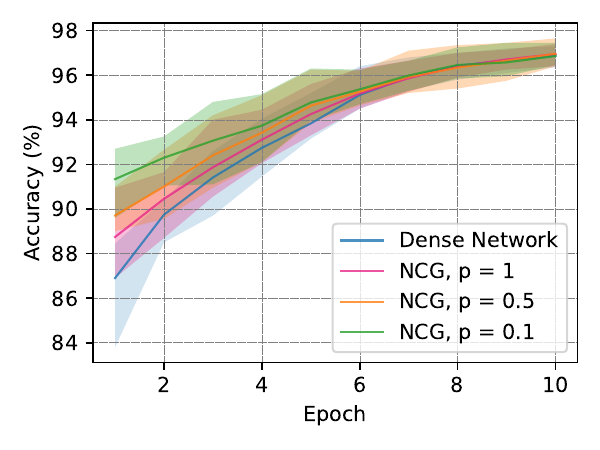}
		}
		\subfloat[Connect-4, Train est set]{ \label{fig:class_gamma25_connect-4_qe_train}
		\includegraphics[trim={0 4mm 0 3.5mm}, clip, width=.315\linewidth]{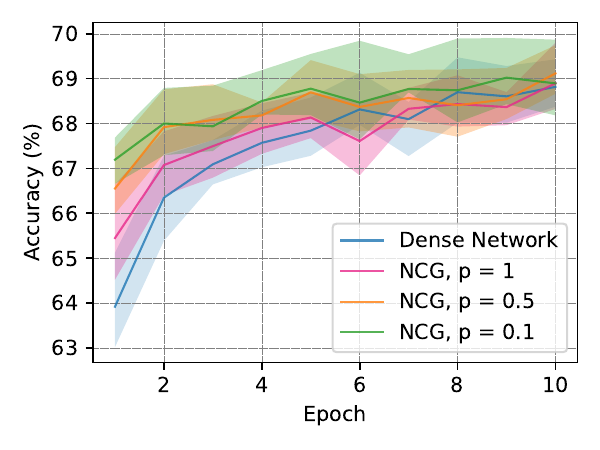}
		} \\
		\caption{
			(Regarding figures \ref{fig:class_gamma25} and \ref{fig:class_gamma25_train}.) Classification accuracy over the training epochs of NCG for the test and train sets of the MNIST, Mushrooms and Connect-4 data sets with $\gamma = 0.25$; lines correspond to the sample mean and shades to the quartiles over 25 experimental runs  
		}
		\label{fig:class_gamma25_qe}
	\end{center}
	\vskip -7mm
\end{figure*}

Figure \ref{fig:class_mesmaTaxa_qe} shows the classification performances of NCG for training with lower connectivity learning rate (same learning rate as the other model parameters, $ \alpha_A = \alpha $), as in figure \ref{fig:class_mesmaTaxa}, in the main body of the paper, and \ref{fig:class_mesmaTaxa_train} in the appendix, which adds the train set learning curves. 

\begin{figure*}[h]
	\begin{center}
		\subfloat[MNIST, Test set]{ \label{fig:class_gamma75_MNIST_qe_test}
			\includegraphics[trim={0 4mm 0 3.5mm}, clip, width=.315\linewidth]{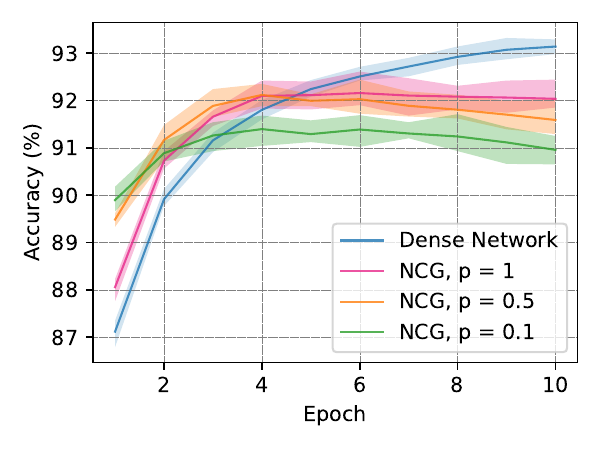}
		}
		\subfloat[Mushrooms, Test set]{ \label{fig:class_gamma75_mushrooms_qe_test}
			\includegraphics[trim={0 4mm 0 3.5mm}, clip, width=.315\linewidth]{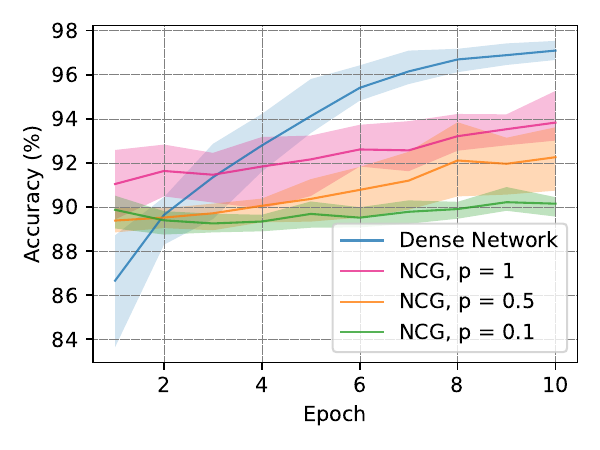}
		}
		\subfloat[Connect-4, Test set]{ \label{fig:class_gamma75_connect-4_qe_test}
			\includegraphics[trim={0 4mm 0 3.5mm}, clip, width=.315\linewidth]{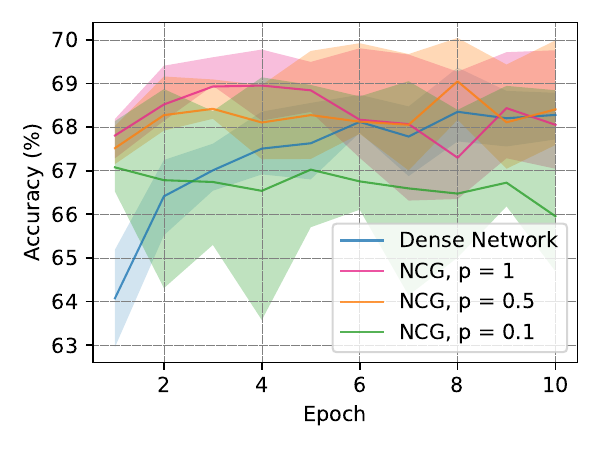}
		} \\
		\subfloat[MNIST, Train set]{ \label{fig:class_gamma75_MNIST_qe_train}
			\includegraphics[trim={0 4mm 0 3.5mm}, clip, width=.315\linewidth]{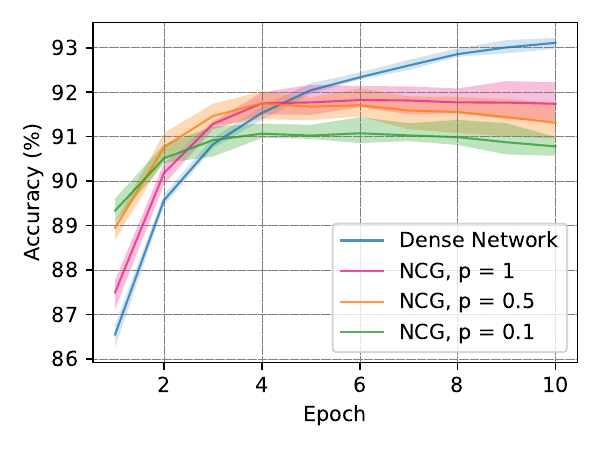}
		}
		\subfloat[Mushrooms, Train set]{ \label{fig:class_gamma75_mushrooms_qe_train}
			\includegraphics[trim={0 4mm 0 3.5mm}, clip, width=.315\linewidth]{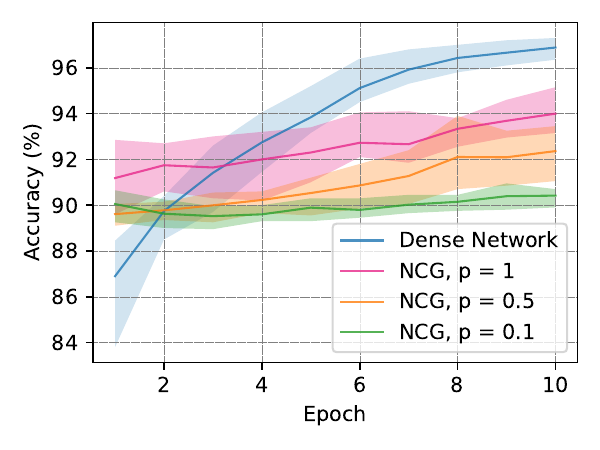}
		}
		\subfloat[Connect-4, Train est set]{ \label{fig:class_gamma75_connect-4_qe_train}
			\includegraphics[trim={0 4mm 0 3.5mm}, clip, width=.315\linewidth]{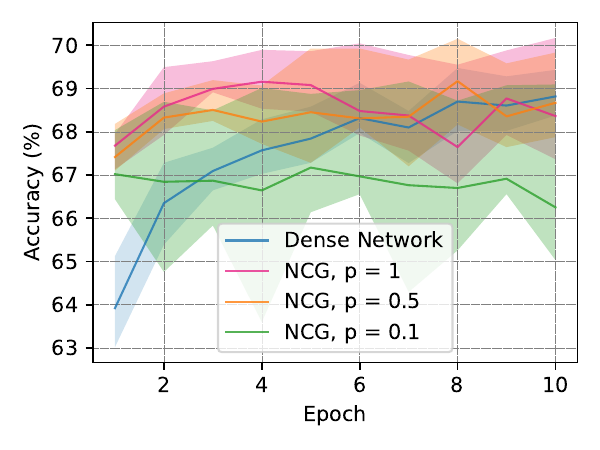}
		} \\
		\caption{
			(Regarding figures \ref{fig:class_gamma75} and \ref{fig:class_gamma75_train}.) Classification accuracy over the training epochs of NCG for the test and train sets of the MNIST, Mushrooms and Connect-4 data sets with $\gamma = 0.75$; lines correspond to the sample mean and shades to the quartiles over 25 experimental runs. 
		}
		\label{fig:class_gamma75_qe}
	\end{center} 
\end{figure*}

Figures \ref{fig:class_gamma25_qe} and \ref{fig:class_gamma75_qe} show the classification performances of NCG trained with smaller (\ref{fig:class_gamma25_qe}) or higher (\ref{fig:class_gamma75_qe}) connectivity activation threshold. They show the quartile information removed from figures \ref{fig:class_gamma25} and \ref{fig:class_gamma75}, in the main body of the paper, and figures \ref{fig:class_gamma25_train} and \ref{fig:class_gamma75_train} in the appendix. 

\section{Contrastive Divergence Approximation}
The experiments on this article applied Contrastive Divergence training using 10 steps of Gibbs Sampling (CD-10). However, changing the number of steps (and the way of obtaining the sample $ \tilde{x} $ entirely) can deeply affect results. To exemplify this, some evaluations using CD-1 (only one step of Gibbs Sampling) were performed. This creates a poorer gradient approximation, which usually affects training negatively. 

\subsection{Generative Results}
\begin{figure}[h]
	\centering
	\includegraphics[trim={4mm 2mm 0 4mm}, clip, width=.6\linewidth]{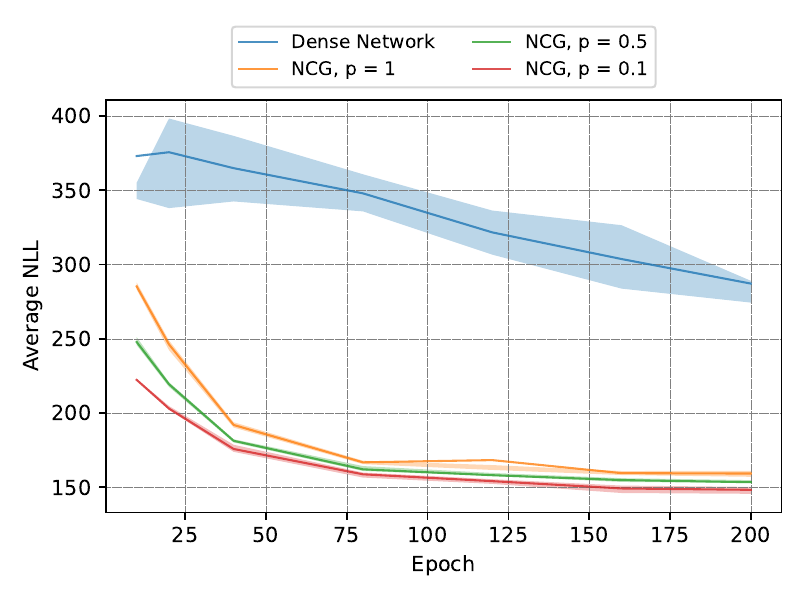}
	\vskip -2mm
	\caption{
		Average NLL over the training epochs for the MNIST dataset with CD-1 -- lines correspond to the sample mean and shades to the sample quartiles over 10 experimental runs. 
	}
	\label{fig:ncg:nll:mnistCD1}
	\vskip -3mm
\end{figure}

\begin{figure*}[h!]
	\def\sizefigs{.3\linewidth}
	
	\centering
	\subfloat[$ p = 1 $]{
		\includegraphics[trim={2mm 2mm 2mm 2mm}, clip, width=\sizefigs]{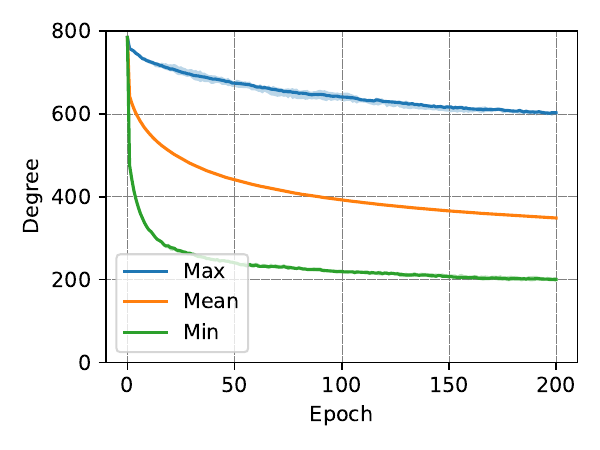}  
	} 
	\subfloat[$ p = 0.5 $]{
		\includegraphics[trim={2mm 2mm 2mm 2mm}, clip, width=\sizefigs]{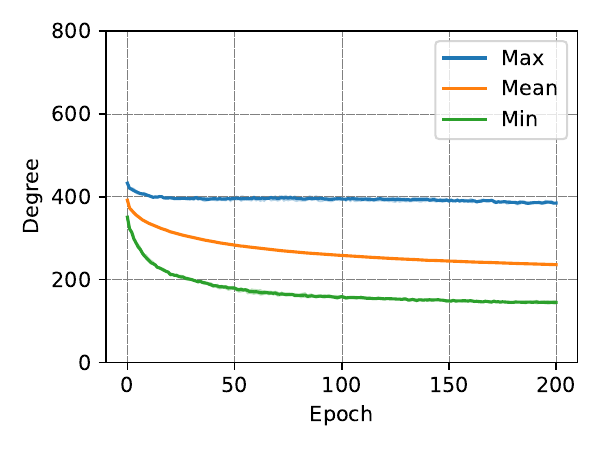}
	}
	\subfloat[$ p = 0.1 $]{
		\includegraphics[trim={2mm 2mm 2mm 2mm}, clip, width=\sizefigs]{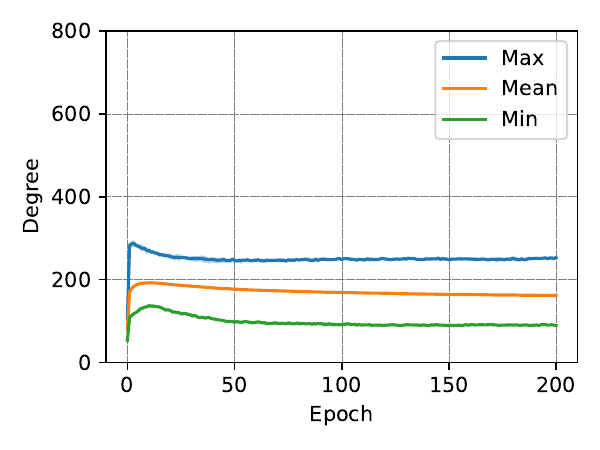}
	}
	\caption{
		Degree statistics (minimum, average, maximum) of the hidden units over the training epochs. Generative results on MNIST, trained using CD-1. Lines correspond to the sample mean and shade corresponds to the sample quartiles over 10 experimental runs. 
	}
	\label{fig:ncg:nll:mnistCD1_conn}
\end{figure*}

Figure \ref{fig:ncg:nll:mnistCD1} portrays the learning curves for dense RBM and NCG, and Figure \ref{fig:ncg:nll:mnistCD1_conn} the corresponding degree statistics. 
It is clear that CD-1 causes a major performance drop for all models considered: 
by the end of training the fully connected RBM has the average 
NLL around 290 in comparison to the 150 seen in Figure \ref{fig:nllMNIST_nll}, and the models 
trained with the NCG method reach at most 150, when before the worse average did 
not surpass 120. The degree statistics for $ p = 1 $ and $ p = 0.5 $ appear to show less 
change along the epochs than what was observed for CD-10, but the differences do not appear to be significant. 


Interestingly, the NCG models' NLL increased less than the classical RBM. The dense network's final NLL is double the value of what was obtained when using CD-10, and the relative increase in performance derived from optimizing the connectivity (NCG) is much larger. Therefore, despite the poorer overall performance, adding NCG generated a greater relative advantage in the CD-1 setting than in CD-10. 

\subsection{Classification Results}
While the goal in the classification task is to maximize accuracy, the objective function 
used during training with aims to minimize the NLL, using CD as an approximation to the gradient. 
Therefore NCG trains the connectivity network (and all the other 
trained parameters) for a slightly inaccurate objective. It stands to reason, therefore, that in 
worsening the approximation for the gradients, its performance will suffer. 

\begin{figure}[h]
	\def\sizefigs{.48\columnwidth}
	\centering
	\subfloat[Train Set]{
		\includegraphics[trim={3.5mm 4mm 3.5mm 3.5mm}, clip, width=\sizefigs]{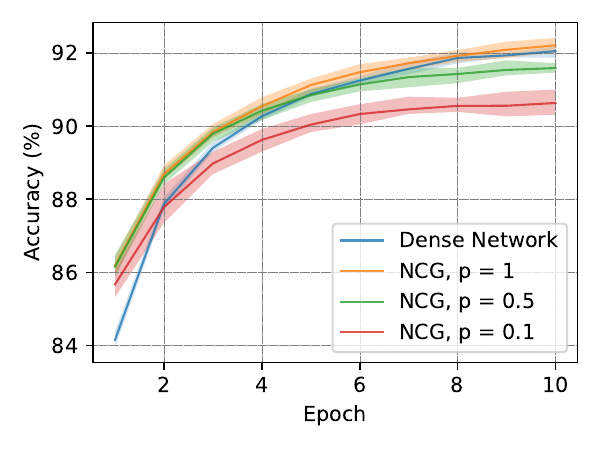}
	} 
	\subfloat[Test Set]{
		\includegraphics[trim={3.5mm 4mm 3.5mm 3.5mm}, clip, width=\sizefigs]{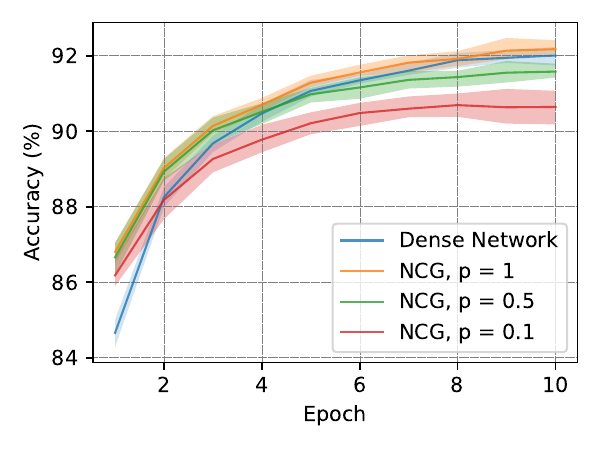}
	}
	\caption{
		Classification accuracy over the training epochs for the train (a) and test (b) sets of the MNIST data set. Trained with CD-1. Lines correspond to the sample mean and shades to the sample quartiles over 25 experimental runs. 
	}
	\label{fig:ncg:class:mnistCD1}
\end{figure}

\begin{figure*}[h]
	\def\sizefigs{.3\linewidth}
	\centering
	\subfloat[$ p = 1 $]{
		\includegraphics[trim={2mm 3mm 2mm 3mm}, clip, width=\sizefigs]{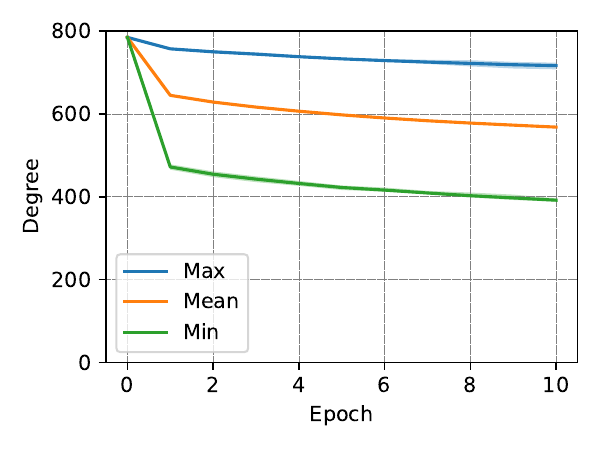}
	}
	\subfloat[$ p = 0.5 $]{
		\includegraphics[trim={2mm 3mm 2mm 3mm}, clip, width=\sizefigs]{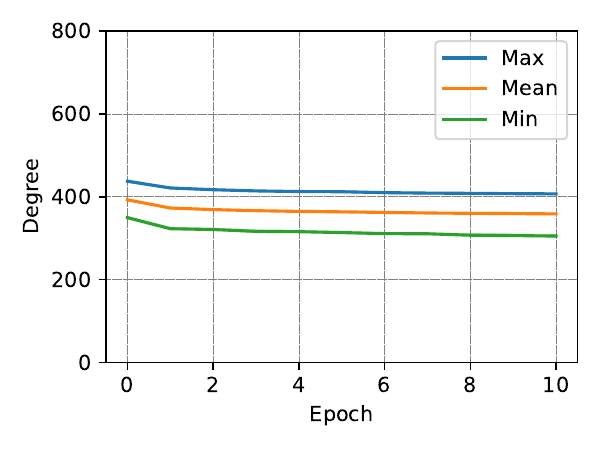}
	}
	\subfloat[$ p = 0.1 $]{
		\includegraphics[trim={2mm 3mm 2mm 3mm}, clip, width=\sizefigs]{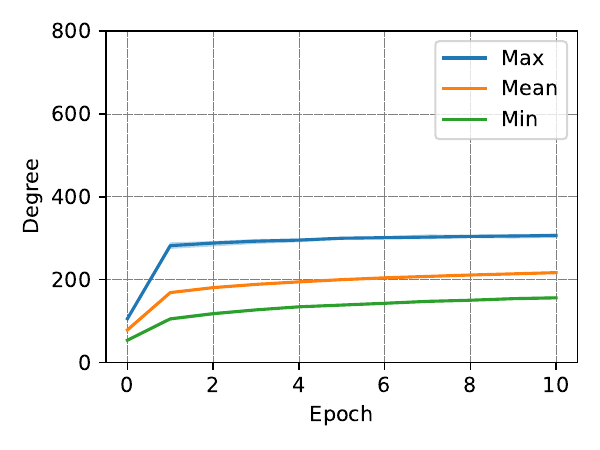}
	}
	\caption{
		Degree statistics (minimum, average, maximum) of the hidden units over the training epochs. Classification results on MNIST, trained with CD-1. Lines correspond to the sample mean and shades to the sample quartiles over 25 experimental runs. 
	}
	\label{fig:ncg:class:mnistCD1_conn}
\end{figure*}

Figure \ref{fig:ncg:class:mnistCD1} shows the evolution of the accuracy over 
epochs for the dense RBM as well as three initializations for the NCG model, for 
both the train and test sets. Figure \ref{fig:ncg:class:mnistCD1_conn} presents the 
corresponding degree statistics evolution, giving an idea of how the connectivity changes 
with training. Once again, the degree statistics do not show much difference from their 
CD-10 counterparts, except that they suffer less change throughout training. 


Although all RBMs have a worse accuracy when training with CD-1, it is clear from 
the results that the relative performance between NCG and the fully connected model diminishes. 
In these circumstances, only NCG initializing with all connections 
activated ($ p = 1 $) manages to surpass the traditional RBM, and even then they have 
very close results. It is not clear that the difference is statistically significant. 
The $ p = 0.1 $ training seems to suffer the most, not showing a 
better accuracy even in the first epoch of training. Overall, the results indicate a very 
different scenario in comparison to the one observed in the generative task, for which the 
addition of connectivity optimization resulted only in positive results, regardless of the CD 
approximation used. 

\section{SET Training}\label{apd:appendix_SET}
\begin{figure}[!h]
	\begin{center}
		\includegraphics[width=.6\columnwidth]{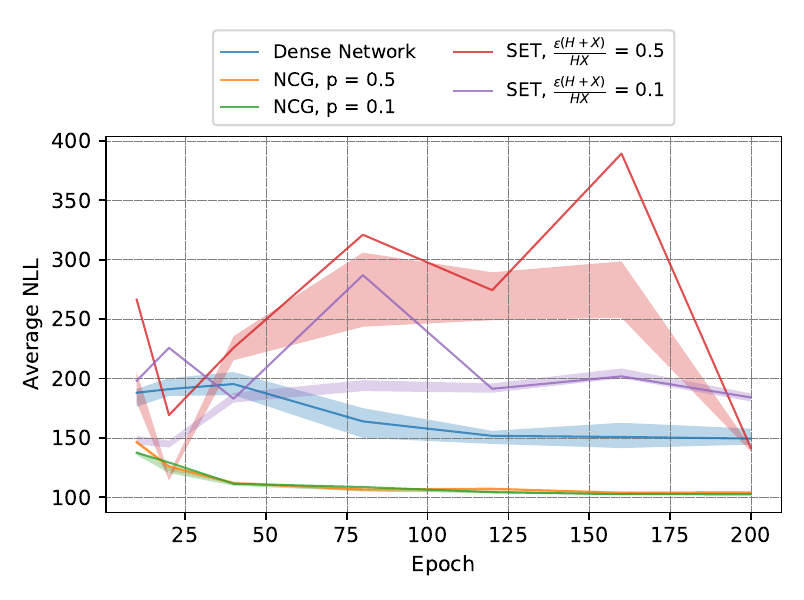}
		\caption{
			Average NLL over the training epochs -- lines are the mean values and shades are the quatiles over 10 experiments. Comparison of NCG with the SET method. 
		}
		\label{fig:set_original_params}
	\end{center}
\end{figure}

As mentioned in Section~\ref{sec:result:nll}, the learning curves reported for the SET method used the hyperparameters reported by \cite{SET_mocanu}. Figure \ref{fig:set_original_params} shows the comparison of NCG with SET, in which SET was trained with the same parameters as the previous NCG experiments. That is, with 500 hidden units, learning rate of $\alpha=0.1$, and the same initial sparsity. Note that the sparsity parameter $ \epsilon $ was chosen so as to create SET networks with nominal sparsity of $ 50\% $ and $ 10\% $ of all possible connections activated, which corresponds to values used for NCG intialization. 

This choice of hyperparameters for the SET method did not yield good performance, which shows noisy learning curves without apparent convergence. Since this scenario did not show good results for SET, experiments with other hyperparameters were performed and reported in Figure \ref{fig:setComp}. In any case, NCG showed significantly superior and more robust performance (less noisy learning curves).

\section{Number of hidden units evaluation}
\label{apd:biggerH}
Throughout this work, NCG has been evaluated utilizing $H=500$ hidden units, which shows good performance for the method. However, the question remains on whether increasing the number of hidden units could potentially increase its performance, especially since the best SET results reported in \cite{SET_mocanu} utilized $H=2500$, a significant increase in the model size. 

\begin{figure}[h]
    \centering
    \includegraphics[width=0.6\linewidth]{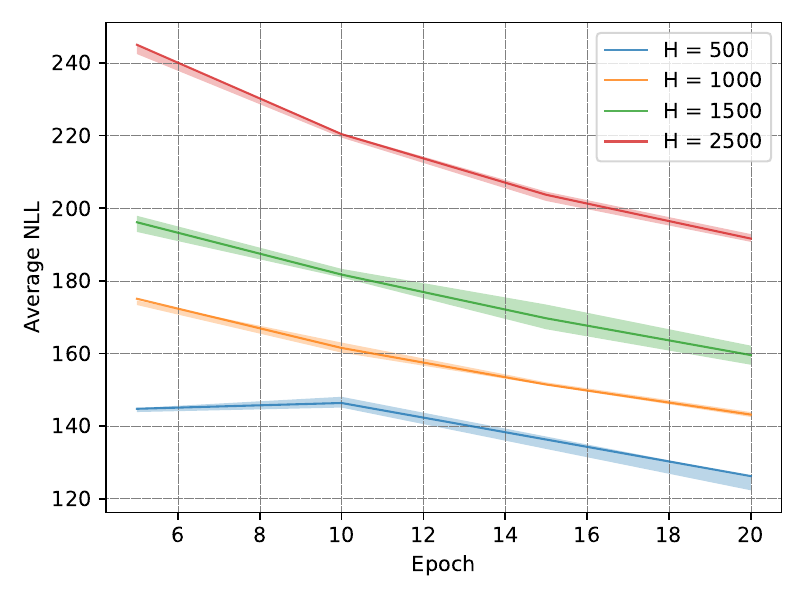}
    \caption{Average NLL over the training epochs -- lines are the mean values and shades are the quatiles over 10 experiments. Comparison of NCG using different numbers of hidden units $H$. NCG was trained with $p=0.5$, $\alpha=0.1$, $\gamma=0.5$.}
    \label{fig:ncg_nll_biggerH}
%
    \includegraphics[width=0.6\linewidth]{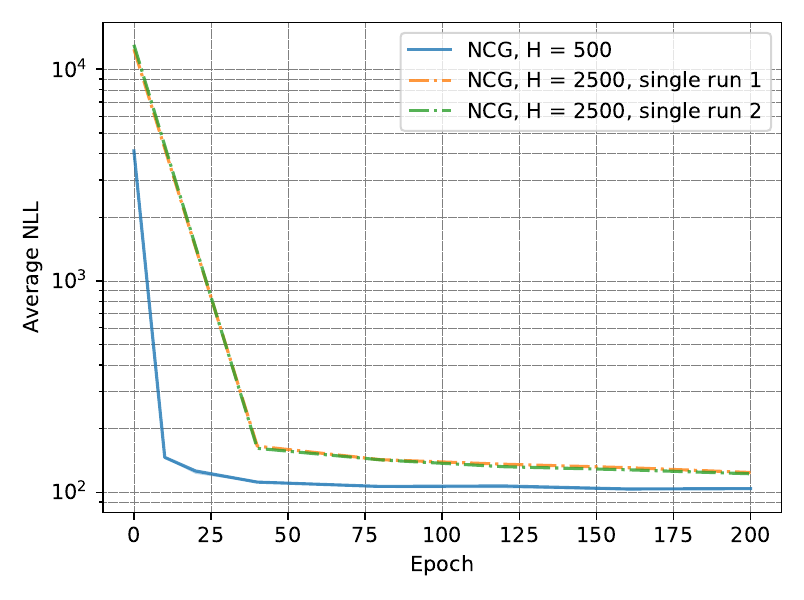}
    \caption{Average NLL over the training epochs. The two dash-dotted lines correspond to single RBM learning curves, which were parameterized with 2500 hidden units ($H=2500$). The full line is the mean value of the NLL performance for RBMs with $H=500$, with the corresponding to the quatiles over 10 experiments. All experiments were initialized with $p=0.5$.}
    \label{fig:nll_singleRuns_h2500}
\end{figure}

In that sense, some experiments were performed utilizing a greater number of hidden units in the trained RBMs. Figure \ref{fig:ncg_nll_biggerH}
shows the overall generative results for the first 20 training epochs of NCG training using the original $500$ units, $1000$, $1500$ and $2500$ hidden units, and Figure \ref{fig:nll_singleRuns_h2500} compares the original NCG results, trained with $H=500$, to the learning curves of two RBMs trained with $H=2500$. 

One can note that the increase in the model size consistently increases the NLL loss as well. As would be expected: the higher number of hidden units greatly increases the number of parameters to be tuned, which requires more training, and also more data. Although it could be that with more training epochs the larger models could eventually surpass the smaller models' performance, the increase in computational power needed to train the extra parameters does not seem to justify their usage. 

\section{Generated Samples for MNIST}
\label{apd:genSamples}

\begin{figure}[h]
    \centering
    \def\sizeNumbers{0.09\textwidth}
    \subfloat[Dense Network]{ \label{fig:genSamples:dense}
        \includegraphics[width=\sizeNumbers]{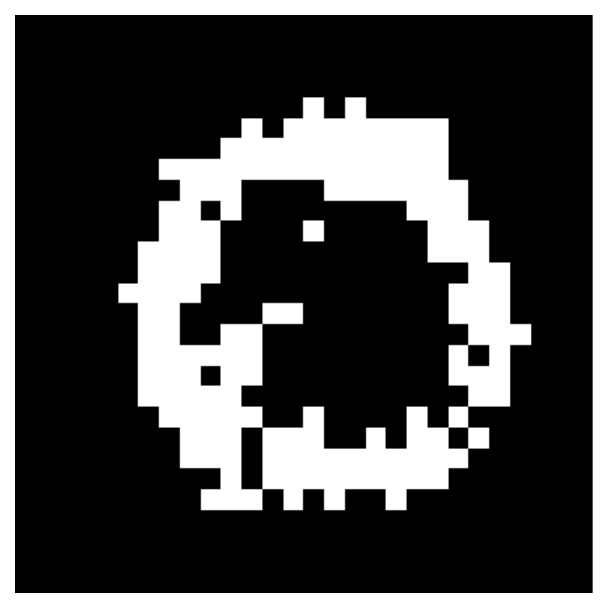} 
        \includegraphics[width=\sizeNumbers]{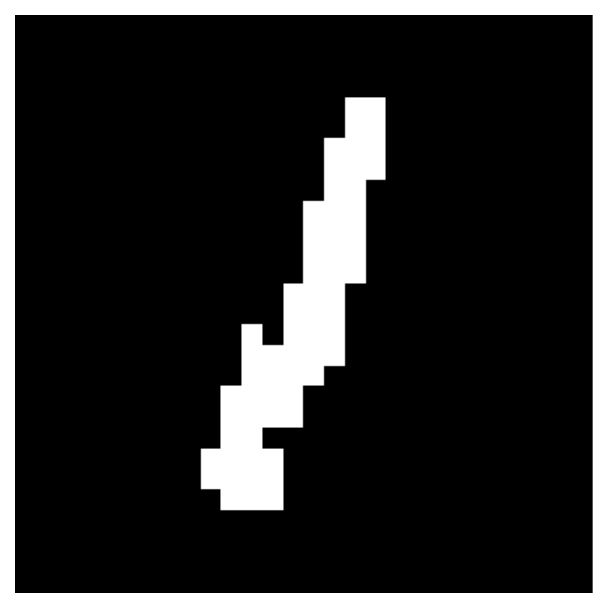}
        \includegraphics[width=\sizeNumbers]{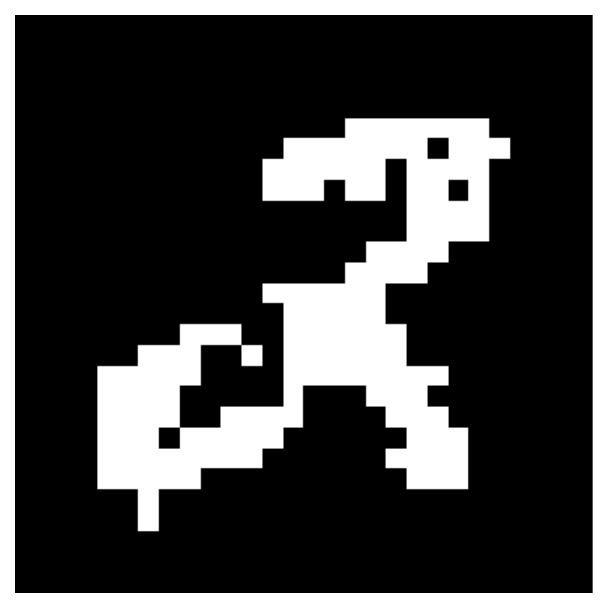}
        \includegraphics[width=\sizeNumbers]{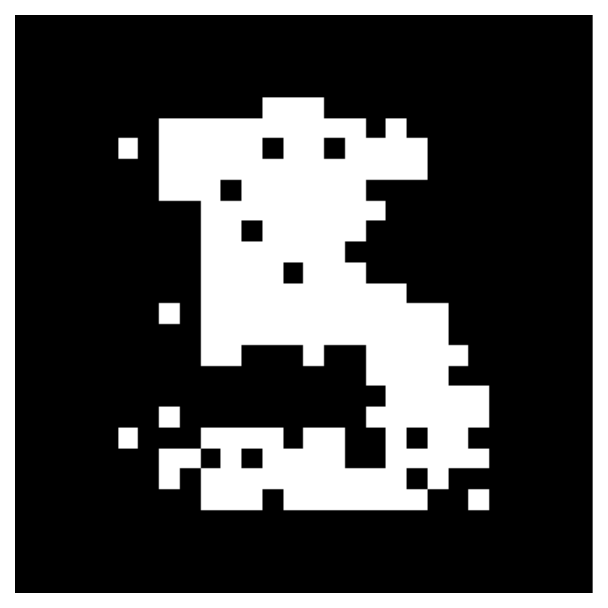}
        \includegraphics[width=\sizeNumbers]{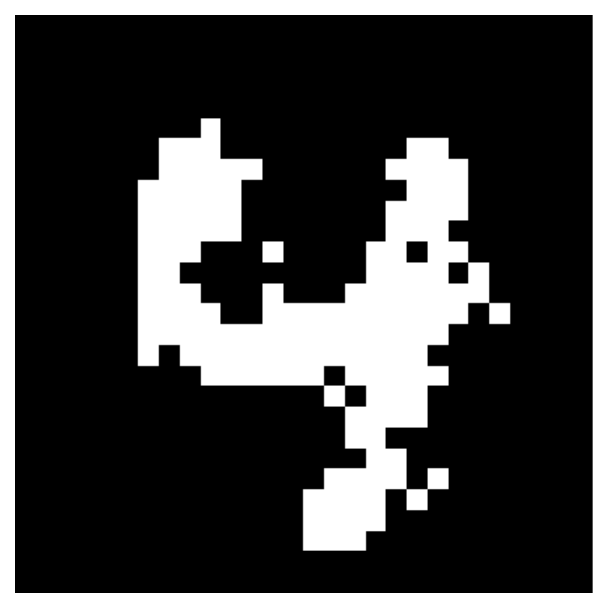}
        \includegraphics[width=\sizeNumbers]{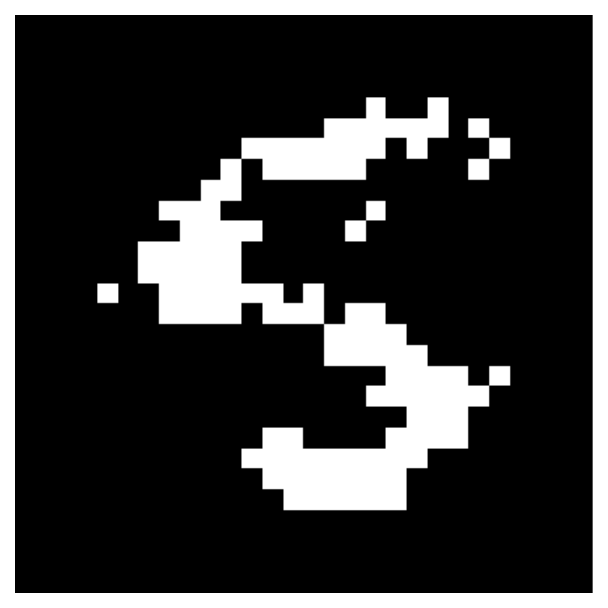}
        \includegraphics[width=\sizeNumbers]{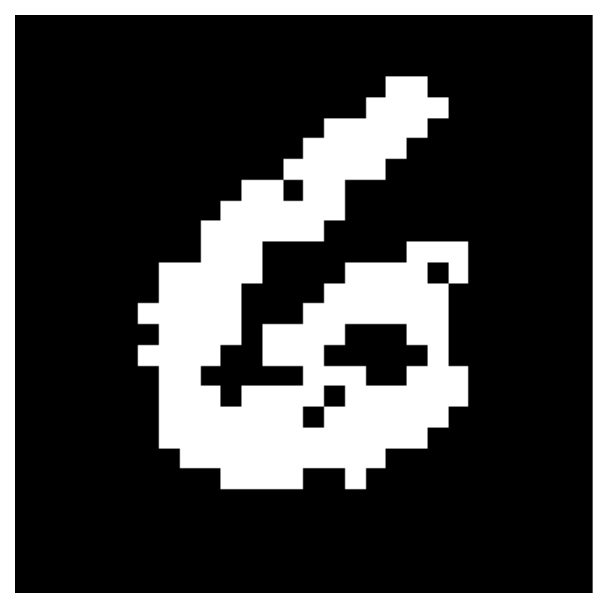}
        \includegraphics[width=\sizeNumbers]{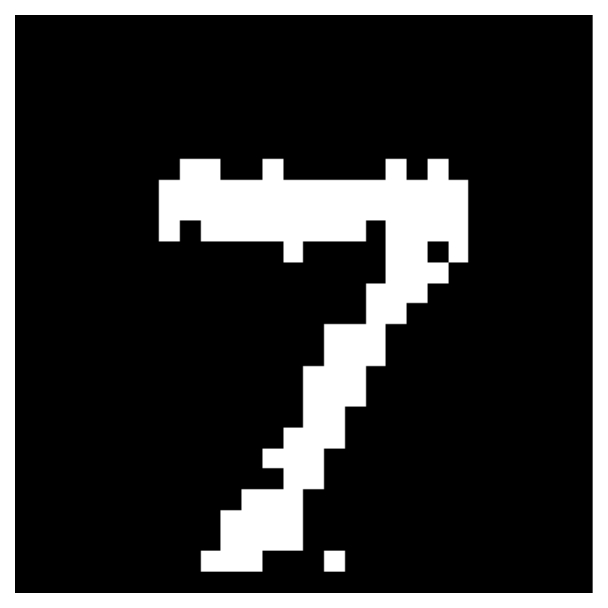}
        \includegraphics[width=\sizeNumbers]{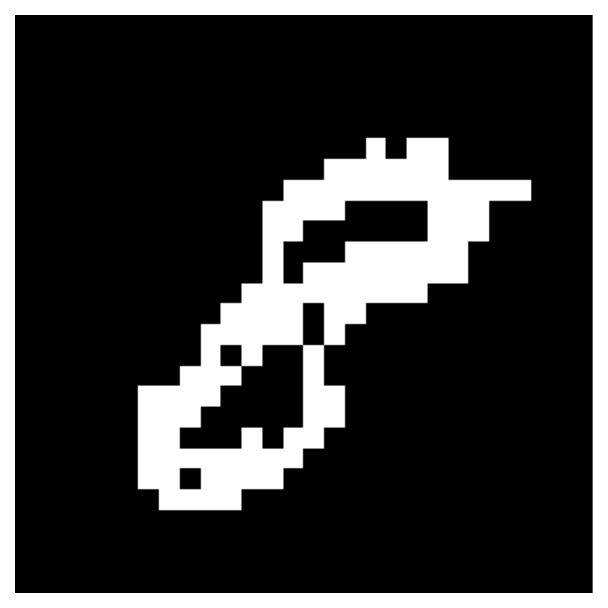}
        \includegraphics[width=\sizeNumbers]{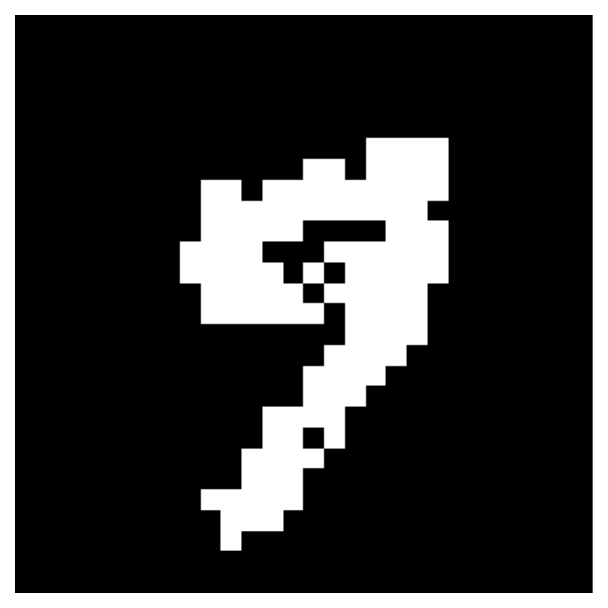}
    }
    \\
    \subfloat[NCG, $p = 0.5$]{ \label{fig:genSamples:ncg}
        \includegraphics[width=\sizeNumbers]{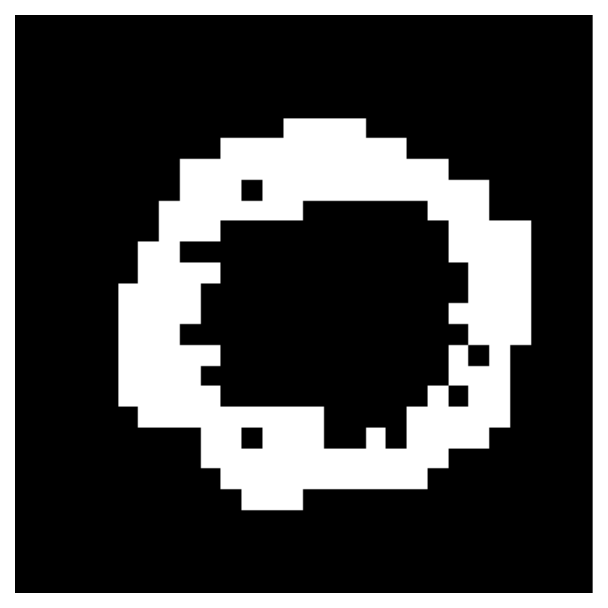}
        \includegraphics[width=\sizeNumbers]{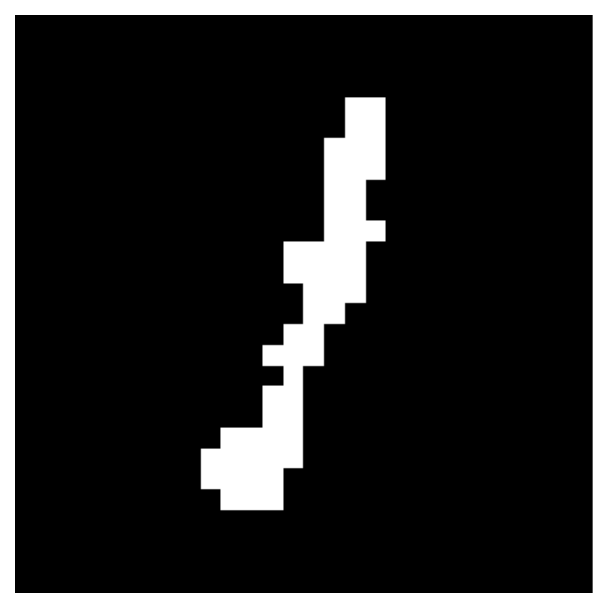}
        \includegraphics[width=\sizeNumbers]{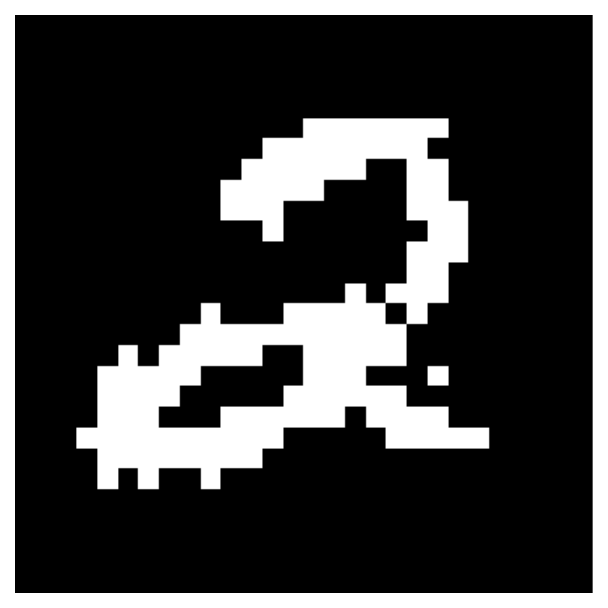}
        \includegraphics[width=\sizeNumbers]{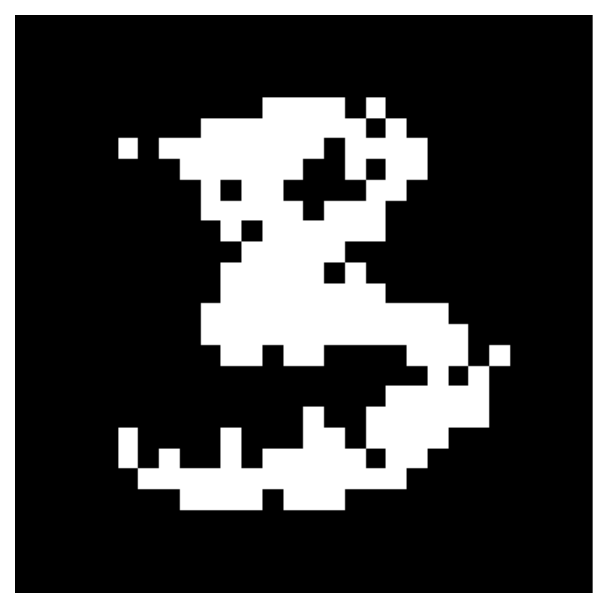}
        \includegraphics[width=\sizeNumbers]{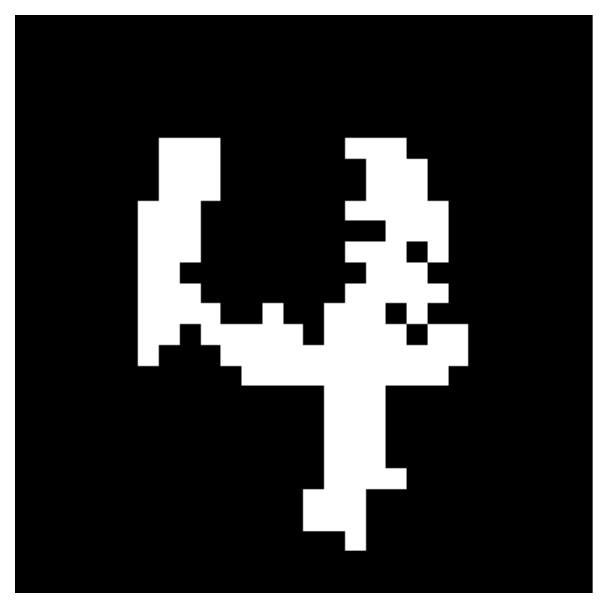}
        \includegraphics[width=\sizeNumbers]{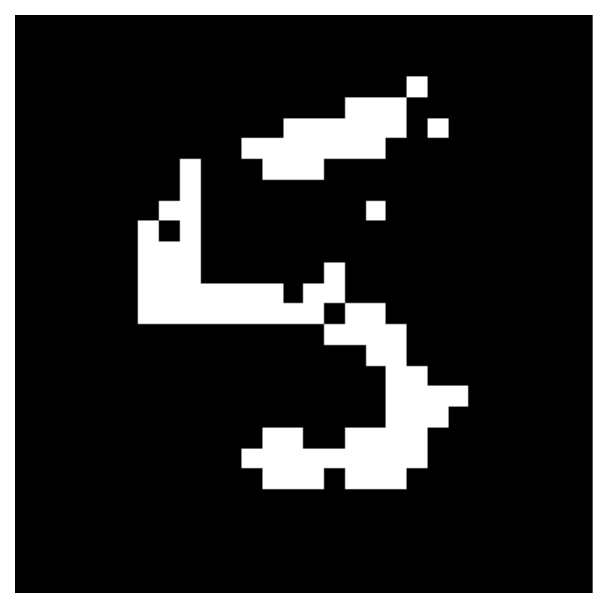}
        \includegraphics[width=\sizeNumbers]{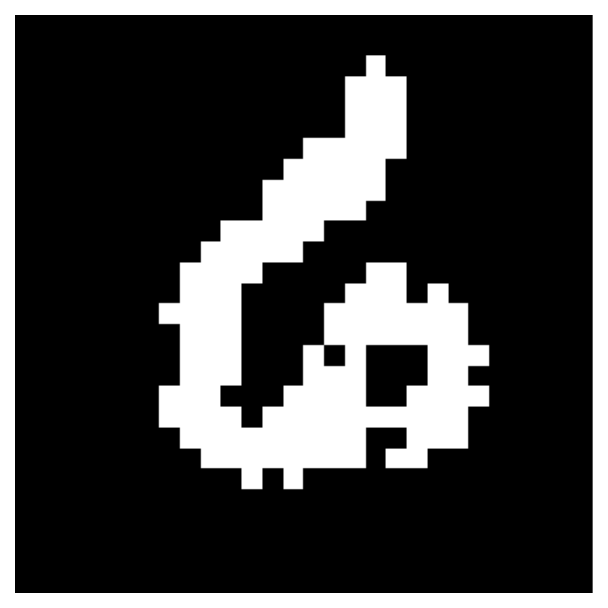}
        \includegraphics[width=\sizeNumbers]{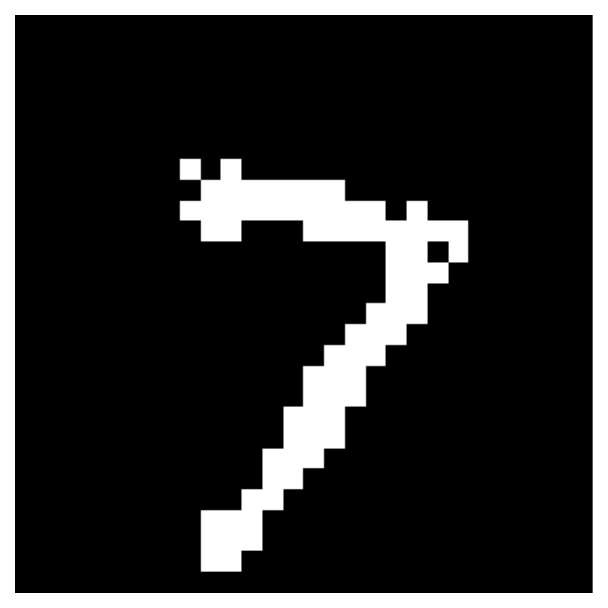}
        \includegraphics[width=\sizeNumbers]{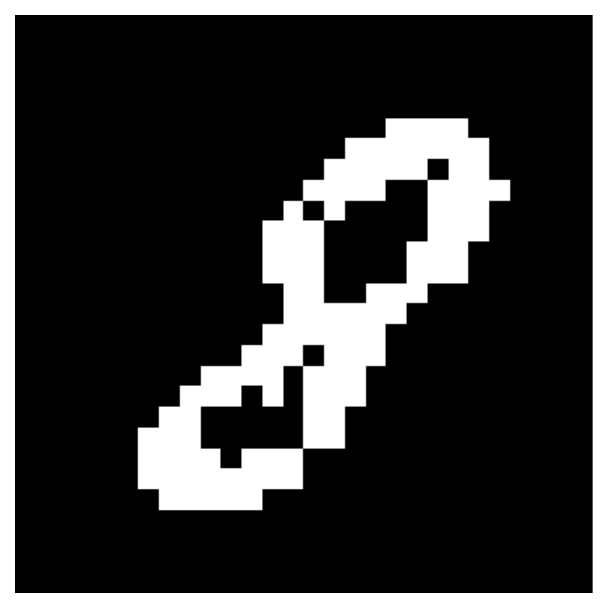}
        \includegraphics[width=\sizeNumbers]{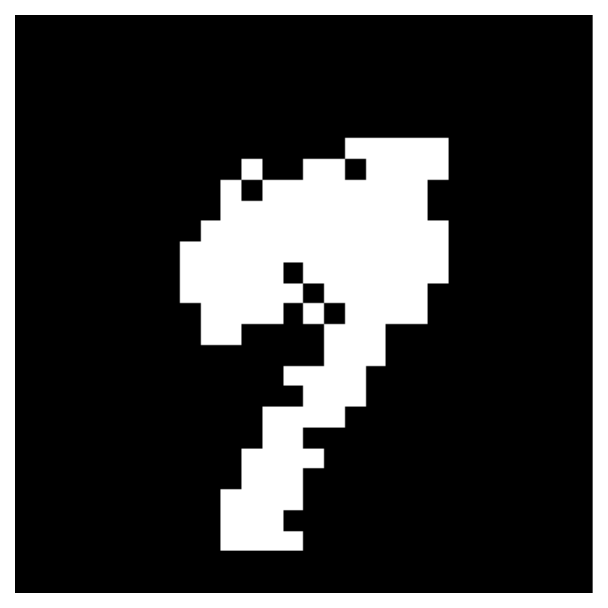}
    }
    \caption{Examples of samples generated by the trained RBMs. One sample was chosen for each existing digit. Figure (a) shows samples generated by a traditional RBM, while (b) shows equivalent sampled generated by the RBM trained with NCG, which was initialized with $p = 0.5$. Both RBMs were trained for 200 epochs with learning rate equivalent to $0.01$.}
    \label{fig:genSamples}
\end{figure}

Figure \ref{fig:genSamples} shows resulting samples generated by RBMs. Both models used were trained for 200 epochs, utilizing CD-10, $H=500$ hidden units, learning rate $\alpha = 0.01$, and a batch of 50 samples. 
The samples were acquired by appling Gibbs Sampling \cite{gibbsSampling} to RBMs initialized with valued acquired from the test set of the MNIST data set. 
Figure \ref{fig:genSamples:dense} has samples generated by a traditional RBM, which is compared to Figure \ref{fig:genSamples:ncg}, which has equivalent samples generated by an RBM trained with the NCG method. 

The qualitative evaluation of the images does not provide conclusive conclusions regarding which method has better performance. Although generally speaking most of the samples from the NCG method seem slightly more recognizable, that is a highly subjective conclusion and one can argue in favor of both estimators from the look of these samples, which is the reason that the authors sought to utilize other performance metrics throughout this work.

\end{document}